\documentclass[10pt,journal,compsoc]{IEEEtran}

\usepackage{epsfig}
\usepackage{graphicx}
\usepackage{amsmath,amsfonts,bm}

\def\eqref#1{equation~\ref{#1}}
\def\1{\bm{1}}

\def\vb{{\bm{b}}}
\def\vc{{\bm{c}}}

\def\vf{{\bm{f}}}
\def\vg{{\bm{g}}}
\def\vh{{\bm{h}}}
\def\vi{{\bm{i}}}

\def\vk{{\bm{k}}}

\def\vo{{\bm{o}}}

\def\vr{{\bm{r}}}

\def\vw{{\bm{w}}}
\def\vx{{\bm{x}}}

\def\vz{{\bm{z}}}

\def\mA{{\bm{A}}}

\def\mD{{\bm{D}}}

\def\mI{{\bm{I}}}

\def\mW{{\bm{W}}}

\DeclareMathAlphabet{\mathsfit}{\encodingdefault}{\sfdefault}{m}{sl}
\SetMathAlphabet{\mathsfit}{bold}{\encodingdefault}{\sfdefault}{bx}{n}
\newcommand{\tens}[1]{\bm{\mathsfit{#1}}}

\def\tB{{\tens{B}}}

\def\tH{{\tens{H}}}

\def\tK{{\tens{K}}}

\def\tW{{\tens{W}}}

\def\tZ{{\tens{Z}}}

\newcommand{\E}{\mathbb{E}}

\usepackage{url}
\usepackage{caption}
\usepackage{subcaption}
\usepackage{booktabs}
\usepackage{dblfloatfix}
\usepackage{authblk}

\ifCLASSOPTIONcompsoc
  \usepackage[nocompress]{cite}
\else
  \usepackage{cite}
\fi

\usepackage[pagebackref=false,breaklinks=true,colorlinks,bookmarks=false]{hyperref}

\begin{document}

%%%%%%%%% TITLE
\title{Gating Revisited: Deep Multi-layer RNNs That Can Be Trained}

%\author{Mehmet Ozgur Turkoglu,\ 
%Stefano D'Aronco,\ 
%Jan Dirk Wegner,\ 
%Konrad Schindler\\
%EcoVision Lab,\ ETH Zurich\\
%{\tt\small { \{ozgur.turkoglu, stefano.daronco, jan.wegner, schindler}\}@geod.baug.ethz.ch}
%}

\author[1]{Mehmet Ozgur Turkoglu}
\author[1]{Stefano D'Aronco}
\author[1,2]{Jan Dirk Wegner}
\author[1]{Konrad Schindler}
\affil[1]{EcoVision Lab - Photogrammetry and Remote Sensing, ETH Zurich, Switzerland}
\affil[2]{Institute for Computational Science, University of Zurich, Switzerland}

%\thispagestyle{empty}

%%%%%%%%% ABSTRACT
\IEEEtitleabstractindextext{%
\begin{abstract}
%Recurrent Neural Networks (RNNs) are widely used models for sequence data. Just like for feedforward networks, it has become common to build "deep" RNNs, i.e., stack multiple recurrent layers to obtain higher-level abstractions of the data. 
We propose a new STAckable Recurrent cell (STAR) for recurrent neural networks (RNNs), which has fewer parameters than widely used LSTM~\cite{lstm} and GRU~\cite{gruX} while being more robust against vanishing or exploding gradients. Stacking recurrent units into deep architectures suffers from two major limitations: \emph{(i)} many recurrent cells (e.g., LSTMs) are costly in terms of parameters and computation resources; and \emph{(ii)} deep RNNs are prone to vanishing or exploding gradients during training. We investigate the training of multi-layer RNNs and examine the magnitude of the gradients as they propagate through the network in the "vertical" direction. We show that, depending on the structure of the basic recurrent unit, the gradients are systematically attenuated or amplified. Based on our analysis we design a new type of gated cell that better preserves gradient magnitude. We validate our design on a large number of sequence modelling tasks and demonstrate that the proposed STAR cell allows to build and train deeper recurrent architectures, ultimately leading to improved performance while being computationally more efficient. %We have made the code available at \href{https://github.com/0zgur0/STAR_Network}{https://github.com/0zgur0/STAR\_Network}.

%, so that with an increasing depth they tend to vanish, respectively explode. 

%, and at the same time significantly reduces the number of parameters compared to other widely used recurrent units. 
%We validate our design on a large number of sequence modelling tasks and demonstrate that the proposed STAR cell allows to build and train deeper recurrent architectures, ultimately leading to improved performance while being computationally efficient.% compared to widely used LSTM~\cite{lstm} and GRU~\cite{gruX}.
\end{abstract}


}

\maketitle

\IEEEdisplaynontitleabstractindextext

% \fi
\IEEEpeerreviewmaketitle

%%%%%%%%% BODY TEXT
\section{Introduction}
Recurrent Neural Networks (RNN) have established themselves as a powerful tool for modelling sequential data. They have led to significant progress for a variety of applications, notably language processing and speech recognition~\cite{sutskever2014sequence, graves2013speech, vinyals2015neural}.

%RNN are basically composed by a cell composed by an input, an internal state, and an output. The input data corresponds to a sequence of length $T$. the cell processes sequentially all the elements in the input sequence updating the hidden state and the output state consequently. Depending on the objective of the sequence modeling the parameters of the 

The basic building block of an RNN is a computational \emph{unit} (or \emph{cell}) that combines two inputs: the data of the current time step in the sequence and the unit's own output from the previous time step. While RNNs can in principle handle sequences of arbitrary and varying length, they are (in their basic form) challenged by long-term dependencies, since learning those would require the propagation of gradients over many time steps.
To alleviate this limitation, \emph{gated} architectures have been proposed, most prominently Long Short-Term Memory (LSTM) cells~\cite{lstm} and Gated Recurrent Units (GRU)~\cite{gruX}. They use gating mechanisms to store and propagate information over longer time intervals, thus mitigating the vanishing gradient problem.

In general, abstract features are often represented better by deeper architectures~\cite{bengio2009learning}. In the same way that multiple hidden layers can be stacked in traditional feed-forward networks, multiple recurrent cells can also be stacked on top of each other, i.e., the output (or the hidden state) of the lower cell is connected to the input of the next-higher cell, allowing for different dynamics. For instance one might expect low-level cues to vary more with lighting, whereas high-level representations might exhibit object-specific variations over time. Several works~\cite{gatedFeedback,rhn,ntu} have shown the ability of deeper recurrent architectures to extract more complex features from the input and make better predictions. However, such architectures are usually composed of just two or three layers because training deeper recurrent architectures still presents an open problem.
%
%Another problematic aspect of many RNN architectures, notably those based on the LSTM design, is the large number of trainable parameters per cell.The training of such architectures is slow even by deep learning standards, and needs a large amount of memory.
More specifically, deep RNNs suffer from two main shortcomings: \emph{(i)} they are difficult to train because of gradient instability, i.e., the gradient  either explodes or vanishes during training; and \emph{(ii)} the large number of parameters contained in each single cell makes deep architectures extremely resource-intensive. Both issues restrict the practical use of deep RNNs and particularly their usage for image-like input data, which generally requires multiple convolutional layers to extract discriminative, abstract representations.
%Since recurrent architectures are trained by propagating gradients across time, it is convenient to "unwrap" them into a lattice with two axes for depth (abstraction level) and time, see Fig.~\ref{fig:RNN}. This view makes it apparent that gradients flow in two directions, namely backwards in time and downwards from deeper to shallower layers.
%
Our work aims to address these weaknesses by designing a recurrent cell that, on the one hand, requires fewer parameters and, on the other hand, allows for stable gradient back-propagation during training; thus allowing for deeper architectures. 
%
%We show that stacking several layers of common RNN cells, by their construction, leads to instabilities (e.g., for deep LSTMs the gradients tend to vanish; for deep vanilla RNNs they tend to explode).

\textbf{Contributions}
We present a detailed, theoretical analysis of how the gradient magnitude changes as it propagates through a cell in a deep RNN lattice. %Depending on the inner architecture of the RNN cell gradients tend to be either amplified or attenuated. %As depth increases, repeated amplification (resp., attenuation) increases the risk of exploding (resp., vanishing) gradients. 
Our analysis offers a different perspective compared to existing literature about RNN gradients, as it focuses on the gradient flow across layers in depth direction, rather than the recurrent flow across time. We show that the two dimensions behave differently, i.e., the ability to preserve gradients in time direction does not necessarily mean that they are preserved across layers, too. We believe that the analysis in this paper contributes a further, complementary step towards a full understanding of gradient propagation in deep RNNs.

We leverage our analysis to design a new, lightweight gated cell, termed the STAckale Recurrent (STAR) unit. The STAR cell better preserves the gradient magnitude in the deep RNN lattice, while at the same time using fewer parameters than existing gated cells like LSTM~\cite{lstm} and GRU~\cite{gruX}.
%leading ultimately to overall better performance.

We compare deep recurrent architectures built from different cells in an extensive set of experiments with several popular datasets. The results confirm our analysis: training very deep recurrent nets fails with most conventional units, whereas the proposed STAR unit allows for significantly deeper architectures. Moreover, our experiments show that the proposed cell outperforms alternative designs on several different tasks and datasets.

%Moreover it appears that, at a given parameter budget, deeper architectures composed of simple cells with few parameters outperform shallower architectures with more complex cells. Our work adds further evidence to the prevailing view that "depth beats width".
%
%In several cases, the ability to go deeper also leads to improved performance.
%\textcolor{red}{To be confirmed. I think if on two tasks (MNIST + one other) we beat the baselines, then we can claim that.}

\section{Related Work}
\label{related_work}

%\subsection{Sequence modelling with RNN}
%RNN and problem with RNN vanishing grad / long term dependencies
%Gating mechanism
Vanishing or exploding gradients during training are a long-standing problem of recurrent (and other) neural networks~\cite{vanishing_munich, vanishing_bengio}. 
Perhaps the most effective measure to address them so far has been to introduce gating mechanisms in the RNN structure, as first proposed by~\cite{lstm} in the form of the LSTM (long short-term memory), and  later by other architectures such as gated recurrent units~\cite{gruX}.

%architecture. LSTM, and its recent descendants like gated recurrent units~\cite[GRU, ][]{gruX} make it possible to employ back-propagation for tasks that require long-range dependencies over many time steps. To deal with exploding gradients, they often use gradient clipping \cite{gradient_clip} or weight norm penalties.

%Initialization method - orthogonal 
%Orthogonal initialization , orthogonal evolution RNNs
%we dont stack them showed in our experiments. 
Importantly, RNN training needs  proper initialisation. In~\cite{identitiy_init,orthagonal_init} it has been shown that initialising the weight matrices with identity and orthogonal matrices can be useful to stabilise the training.
%We initialize the networks with orthogonal weight matrices in our experiments as well. 
This idea is further develop in~\cite{uRNN,FCuRNN}, where authors impose the orthogonality throughout the entire training to keep the amplification factor of the weight matrices close to unity, leading to a more stable gradient flow. Unfortunately, it has been shown~\cite{softOrth} that such hard orthogonality constraints hurt the representation power of the model and in some cases even \emph{de}stabilise the optimisation.
%In this paper, we show that directly stacking RNN layers with orthogonality constraint would not work. #TODO do we show that?

%External memory networks

%Other line of works, FGRNN, the ones with extra connections
%skip connection
Another line of work has studied ways to mitigate the vanishing gradient problem by introducing additional (skip) connections across time and/or layers. Authors in~\cite{skipRNN} have shown that skipping state updates in RNNs  shrinks the effective computation graph and thereby helps to learn longer-range dependencies. Other works such as~\cite{residual_LSTM,resRNN} introduce a residual connection between LSTM layers; however, the performance improvements are limited. In~\cite{gatedFeedback} the authors propose a gated feedback RNN that extends the stacked RNN architecture with extra connections. An obvious disadvantage of such an architecture are the extra computations and memory costs of the additional connections. Moreover, the authors only report results for rather shallow networks up to 3 layers.
%In the discussion section, we discuss \cite{gatedFeedback}'s results which are in align with our findings. 

Many of the aforementioned works propose new RNN architectures by leveraging a gradient propagation analysis. However all of these studies, as well as other studies which specifically aim at modelling accurately gradient propagation in RNNs~\cite{uRNN,mhammedi2017efficient,mean_field_1}, overlook the propagation of the gradient along the "vertical" depth dimension. In this work we will employ similar gradient analysis techniques, but focus on the depth dimension of the network.

Despite the described efforts, it remains challenging to train deep RNNs. In~\cite{rhn} authors propose to combine LSTMs and highway networks~\cite{highway} to form Recurrent Highway Networks (RHN) and train deeper architectures. RHN are popular and perform well on language modelling tasks, but they are still prone to exploding gradients, as illustrated in our experiments. Another solution to alleviate gradient instability in deep RNNs was recently proposed in~\cite{indRNN}. The work suggests the use of a restricted RNN called IndRNN where all interactions are removed between neurons in the hidden state of a layer. This idea combined with the usage of batch normalization appears to greatly stabilize the gradient propagation through layers at the cost of a much lower representation power per layer. 
Such feature hinders IndRNN ability to achieve high performance for complex problems such as satellite image sequence classification or other computer vision tasks. In these tasks it is very important to merge information from neighboring pixels to increase the receptive field of the network so that the model has the ability to represent long-range spatial dependencies. Since IndRNN has no interaction between neurons it is difficult to achieve good spatio-temporal modeling effectively.

%Another solution to alleviate gradient instability has bee proposed in~\cite{indRNN}, this works suggest the use of a restricted RNN where all interactions are removed between neurons in the hidden state of a layer. This appears to greatly reduce the exploding gradient problem, at the cost of a much lower representation power per layer.

%\textcolor{red}{ideally we should find some other shortcoming if indRNN, one of our take-home messages is also better deeper than wider}% # citation needed?

%\subsection{Spatio-temporal modelling with convRNN}
%convLSTM paper
To process image sequence data, computer vision systems often rely on
Convolutional LSTMs~\cite{convLSTM}. But while very deep CNNs are very effective and now standard~\cite{alexnet,vgg}, stacks made of more than a few convLSTMs do not train well. Moreover, the computational cost increase rather quickly due to the large numbers of parameters in each LSTM cell. In practice, shallow versions are preferred, for instance~\cite{videolstm} use a single layer for action recognition, and~\cite{gesture_convLSTM} use two layers to recognise hand gestures (combined with a deeper feature extractor without recursion).

\section{Background and Problem Statement}\label{sec:background}

In this section we revisit the mathematics of RNNs with particular emphasis on the gradient propagation. We will then leverage this analysis to design a more stable recurrent cell, which is described in Sec.~\ref{sec:star}.
A RNN cell is a non-linear transformation that maps the input signal $\vx_t$ at time $t$ and the hidden state of the previous time step $t-1$ to the current hidden state $\vh_t$:
\begin{equation}
    \vh_{t} = f (\vx_{t},\vh_{t-1},\mW)\,,
\end{equation}
with $\mW$ the trainable parameters of the cell. The input sequences have an overall length of $T$, which can vary. It depends on the task whether the final state $\vh_T$, the complete sequence of states $\{\vh_{t}\}$, or a single sequence label (typically defined as the average $\frac{1}{T}\sum_t\vh_t$) are the desired target prediction for which loss $\mathcal{L}$ is computed. Learning amounts to fitting $\mW$ to minimise the loss, usually with stochastic gradient descent. 
 
\begin{figure}%[t]
    \centering
    \begin{subfigure}[b]{0.85\columnwidth}
        \includegraphics[width=\textwidth]{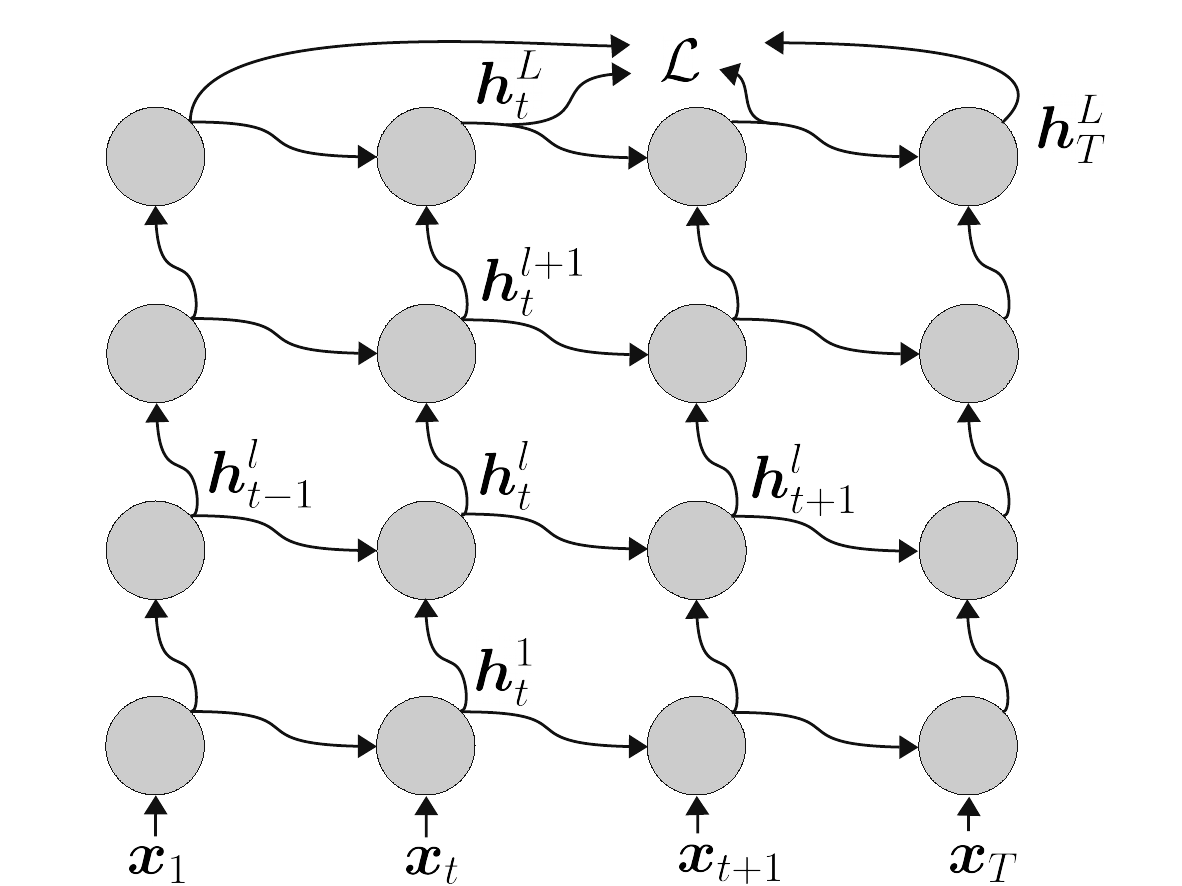}
        \caption{}
        \label{fig:grads_1}
    \end{subfigure}
    \begin{subfigure}[b]{0.85\columnwidth}
        \includegraphics[width=\textwidth]{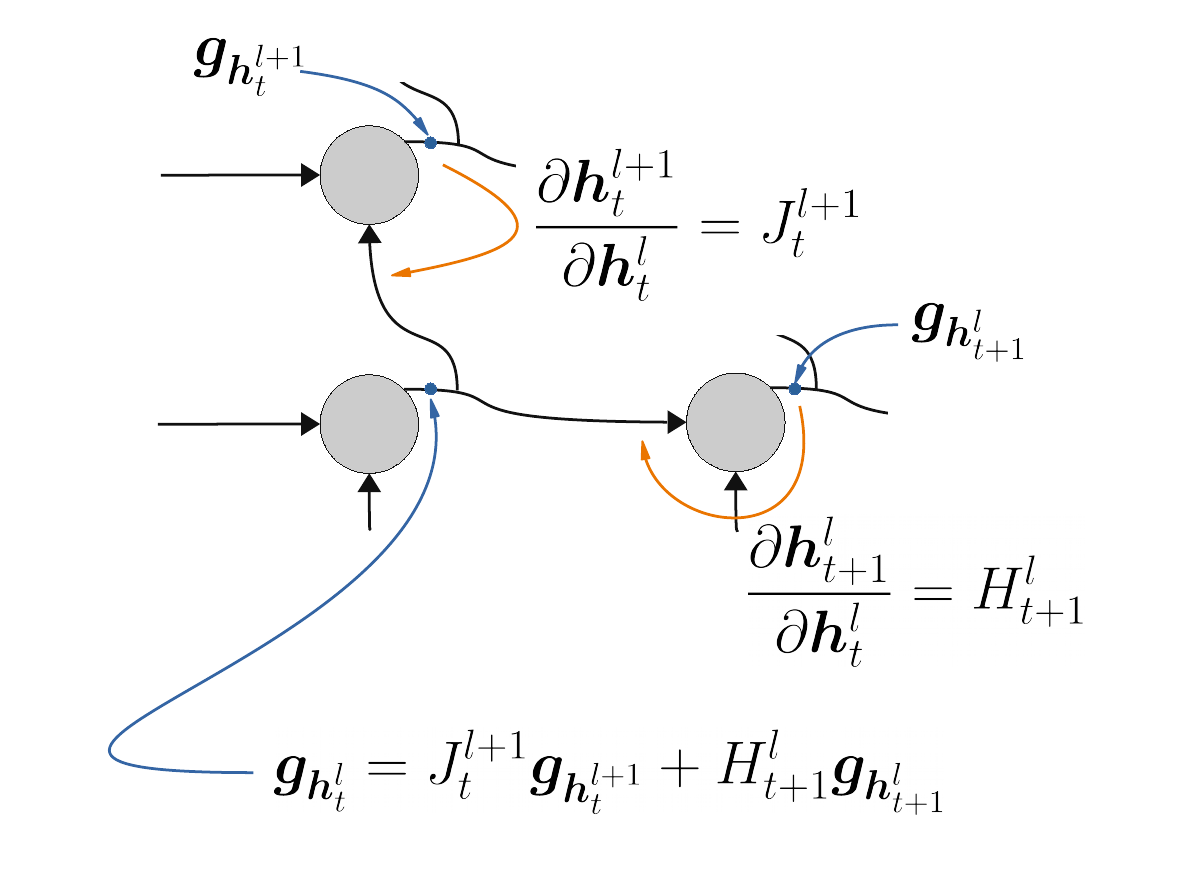}
        \caption{}
        \label{fig:grads_2}
    \end{subfigure}
    \caption{ (a) General structure of an unfolded deep RNN (b) Detail of the gradient backpropagation in the two dimensional lattice.}\label{fig:RNN}
\end{figure}

When stacking multiple RNN cells on top of each other, the hidden state of the lower level $l-1$ is passed on as input to the next-higher level $l$ (Fig.~\ref{fig:RNN}). In mathematical terms this corresponds to the recurrence relation
\begin{equation}
    \vh^l_{t} = f (\vh^{l-1}_{t},\vh^{l}_{t-1},\vw)\,.
\end{equation}
Temporal unfolding leads to a two-dimensional lattice with depth $L$ and length $T$ (Fig.~\ref{fig:RNN}), the forward pass runs from left to right and from bottom to top. Gradients flow in opposite direction: at each cell the gradient w.r.t.\ the loss arrives at the output gate and is used to compute the gradient w.r.t.\ \emph{(i)} the weights, \emph{(ii)} the input, and \emph{(iii)} the previous hidden state. The latter two gradients are then propagated through the respective gates to the preceding cells in time and depth. In the following, we investigate how the magnitude of these gradients changes across the lattice. The analysis, backed up by numerical simulations, shows that common RNN cells are biased towards attenuating or amplifying the gradients and thus prone to destabilising the training of deep recurrent networks.

\subsection{Gradient Magnitudes}

The gradient w.r.t. the trainable weights at a single cell in the lattice is
\begin{equation}
    %\nabla_\vw \mathcal{L} = \frac{\partial \vh_t^l}{\partial \vw} \nabla_{\vh_t^l} \mathcal{L},
    \vg_\vw = \frac{\partial \vh_t^l}{\partial \vw} \vg_{\vh_t^l},
\end{equation}
where $\frac{\partial \vh_t^l}{\partial \vw}$ denotes the Jacobian matrix and $\vg_{\vh_t^l}$ is a column vector containing the partial derivatives of the loss w.r.t.\ the cell's output (hidden) state.
From the equation, it becomes apparent that the Jacobian acts as a "gain matrix" on the gradients, and should on average preserve their magnitude to prevent them from vanishing or exploding. We obtain the recurrence for propagation by expanding the gradient $\vg_{\vh_t^l}$
\begin{equation}
    \vg_{\vh_t^l} = \tfrac{\partial \vh_t^{l+1}}{\partial \vh_t^{l}} \vg_{\vh_t^{l+1}}  + 
                                   \tfrac{\partial \vh_{t+1}^{l}}{\partial \vh_t^{l}} \vg_{\vh_{t+1}^{l}}
                            \!= J^{l+1}_t \vg_{\vh_t^{l+1}}  + H^{l}_{t+1} \vg_{\vh_{t+1}^{l}}, 
\label{eq:grad_propagation}
\end{equation}
with $J^{l}_t$ the Jacobian w.r.t.\ the input and $H^{l}_t$ the Jacobian w.r.t.\ the hidden state. Ideally we would like the gradient magnitude $\| \vg_{\vh_t^l} \|_2$ to remain stable for arbitrary $l$ and $t$.
Characterising that magnitude completely is difficult because correlations may exist between $\vg_{\vh_t^{l+1}}$ and $\vg_{\vh_{t+1}^{l}}$ for instance, due to weight sharing. Nonetheless, it is evident that the two Jacobians $J^{l+1}_t$ and $H^l_{t+1}$ play a fundamental role: if their singular values are small, they will attenuate the gradients and cause them to vanish sooner or later. If their singular values are large, they will amplify the gradients and make them explode.%
\footnote{A subtle point is that sometimes large gradients are the precursor of vanishing gradients, if the associated large parameter updates cause the non-linearities to saturate.}

In the following, we analyse the behaviour of the two matrices for two widely used RNN cells. We first consider the most simple RNN cell, hereinafter called Vanilla RNN (vRNN). Its recurrence equation reads
\begin{equation}
\vh_t^l = \tanh (\mW_{x}\vh_t^{l-1}+ \mW_{h}\vh_{t-1}^l+ \vb)\,
\end{equation}
from which we get the two Jacobians
\begin{align}
J^{l}_t = \mD_{\tanh(\mW_{x}\vh_t^{l-1}+ \mW_{h}\vh_{t-1}^l + \vb)'}\mW_{x}\\
H^{l}_t = \mD_{\tanh(\mW_{x}\vh_t^{l-1}+ \mW_{h}\vh_{t-1}^l + \vb)'}\mW_{h}\,
\label{eq:weightmat}
\end{align}
where $\mD_\vx$ denotes a diagonal matrix with the elements of vector $\vx$ as diagonal entries.
Ideally, we would like to know the expected values of the two matrices' singular values. Unfortunately, there is no easy way to derive a closed-form analytical expressions for them, but we can compute them for a fixed, representative point. The most natural and illustrative choice is to set $\vh_t^{l-1}=\vh_{t-1}^l=0$ because \emph{(i)} in practice, RNNs' initial hidden states are set to $h_{0}^l=0$ (like in our experiments), and \emph{(ii)} it is a stable and attracting fixed point so if the hidden state is perturbed around this point, it tends to return to its initial point. 
Note that this does not mean that the hidden state of the network is always equal to zero. The goal of the assumption is only simply to fix the value of the hidden state, in order to analyse the gradient propagation.
We further choose weight matrices $\mW_{h}$ and $\mW_{x}$ with average singular value equal to one and $\vb=0$ (different popular initialisation strategies, such as orthogonal and identity matrices, are aligned with this assumption).
Moreover, according to~\cite{lee2019wide,jacot2018neural}, in the limit of a very wide network the parameters tend to stay close to their initial values, as a result the assumptions made are still legitimate during training  %\textcolor{blue}{(see Fig. \ref{fig:weight_norms} and \ref{fig:mean_h} in the Appendix for empirical evidence)}.
(see the Appendix for empirical evidence).
%
%and pre-activations tend to be distributed around their initial values which is discussed in .   
%
Since the derivative $\tanh'(0)=1$, the average singular values of all matrices in Eq.~(\ref{eq:weightmat}) are equal to $1$ in this configuration.

\begin{table*}
\begin{align}
&J^{l}_t = \mD_{\tanh(\vc_t^l)}\mD_{(\vo_{t}^{l})'}\mW_{xo} +  \mD_{\tanh(\vc_t^l)'}\mD_{\vo_t^l}(\mD_{\vc_{t-1}^l}\mD_{(\vf_t^l)'}\mW_{xf} + \mD_{\vz_t^l}\mD_{(\vi_t^l)'}\mW_{xi} + \mD_{\vi_t^l}\mD_{(\vz_t^l)'}\mW_{xz} )
\label{eq:J_lstm}\\
&H^{l}_t = \mD_{\tanh(\vc_t^l)}\mD_{(\vo_{t}^{l})'}\mW_{ho} +  \mD_{\tanh(\vc_t^l)'}\mD_{\vo_t^l}(\mD_{\vc_{t-1}^l}\mD_{(\vf_t^l)'}\mW_{hf} + \mD_{\vz_t^l}\mD_{(\vi_t^l)'}\mW_{xi} + \mD_{\vi_t^l}\mD_{(\vz_t^l)'}\mW_{hz} )
\label{eq:H_lstm}
\end{align}
%\protect\caption{Long equation}
%\label{tab:long-eq}
\end{table*}

\begin{figure}
    \centering
    \renewcommand\tabcolsep{0pt}
    \small
    \begin{tabular}{ccc}
%\rotatebox{0}{$\qquad$a)} &
\includegraphics[width=0.33\columnwidth]{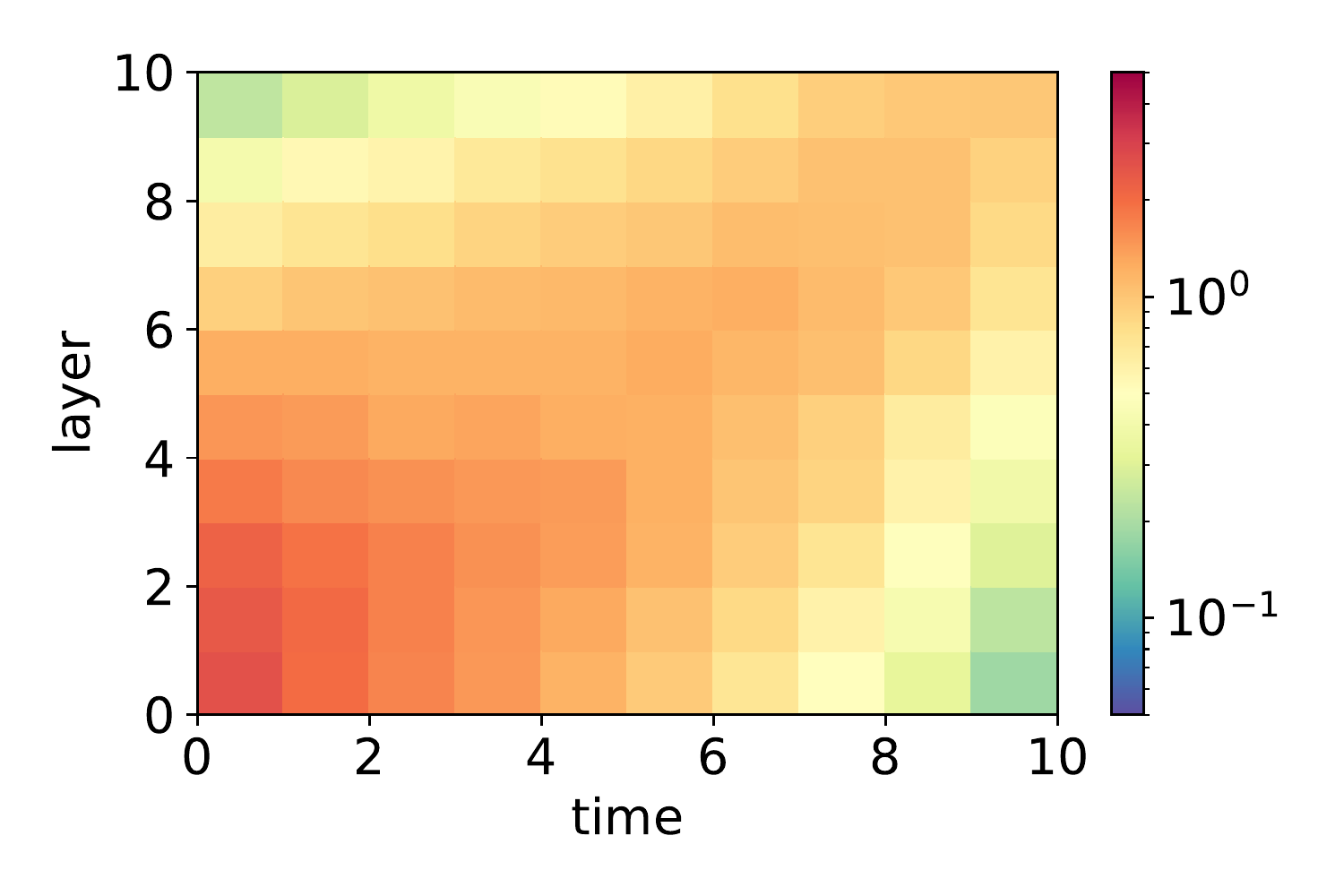} &
\includegraphics[width=0.33\columnwidth]{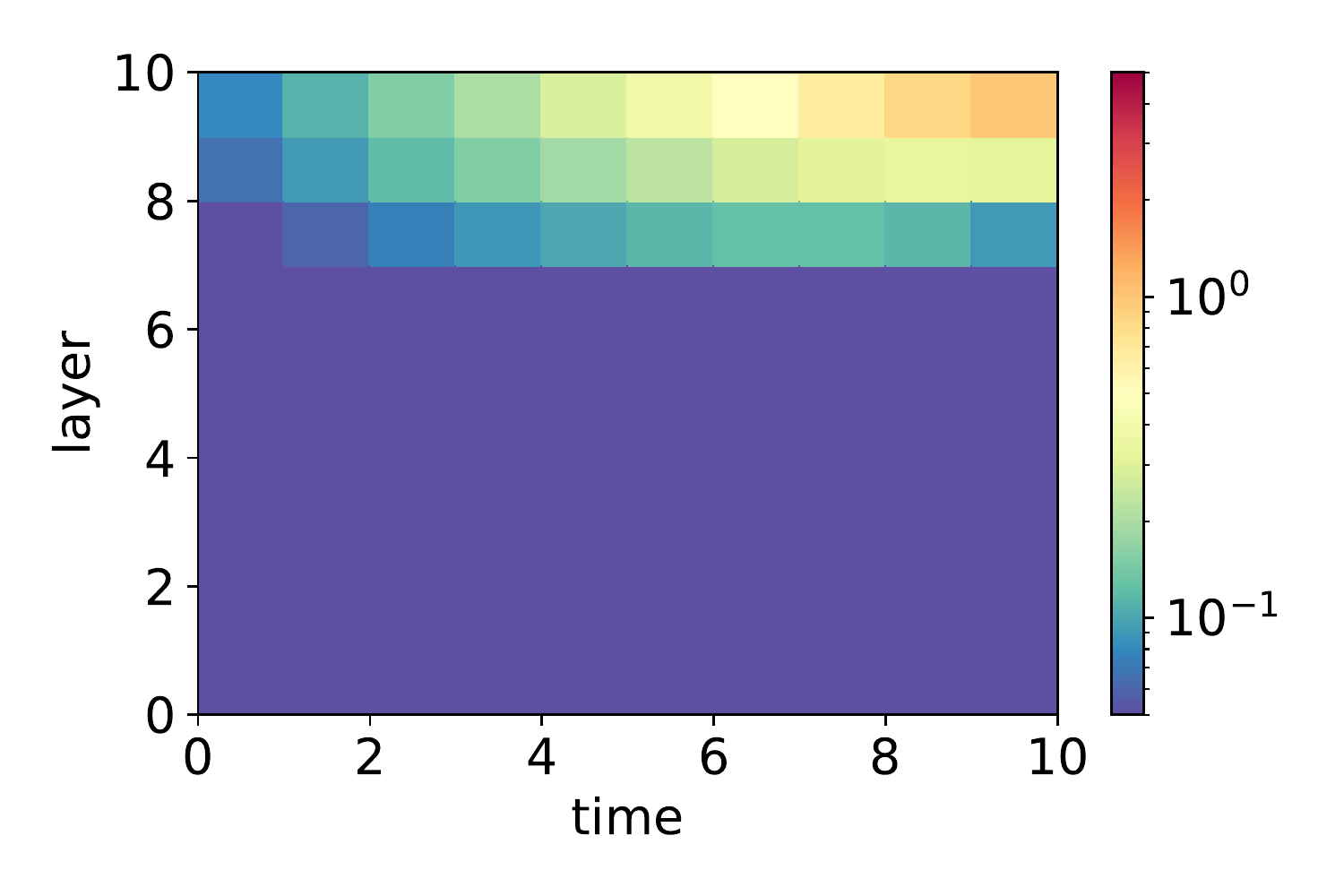} &
\includegraphics[width=0.33\columnwidth]{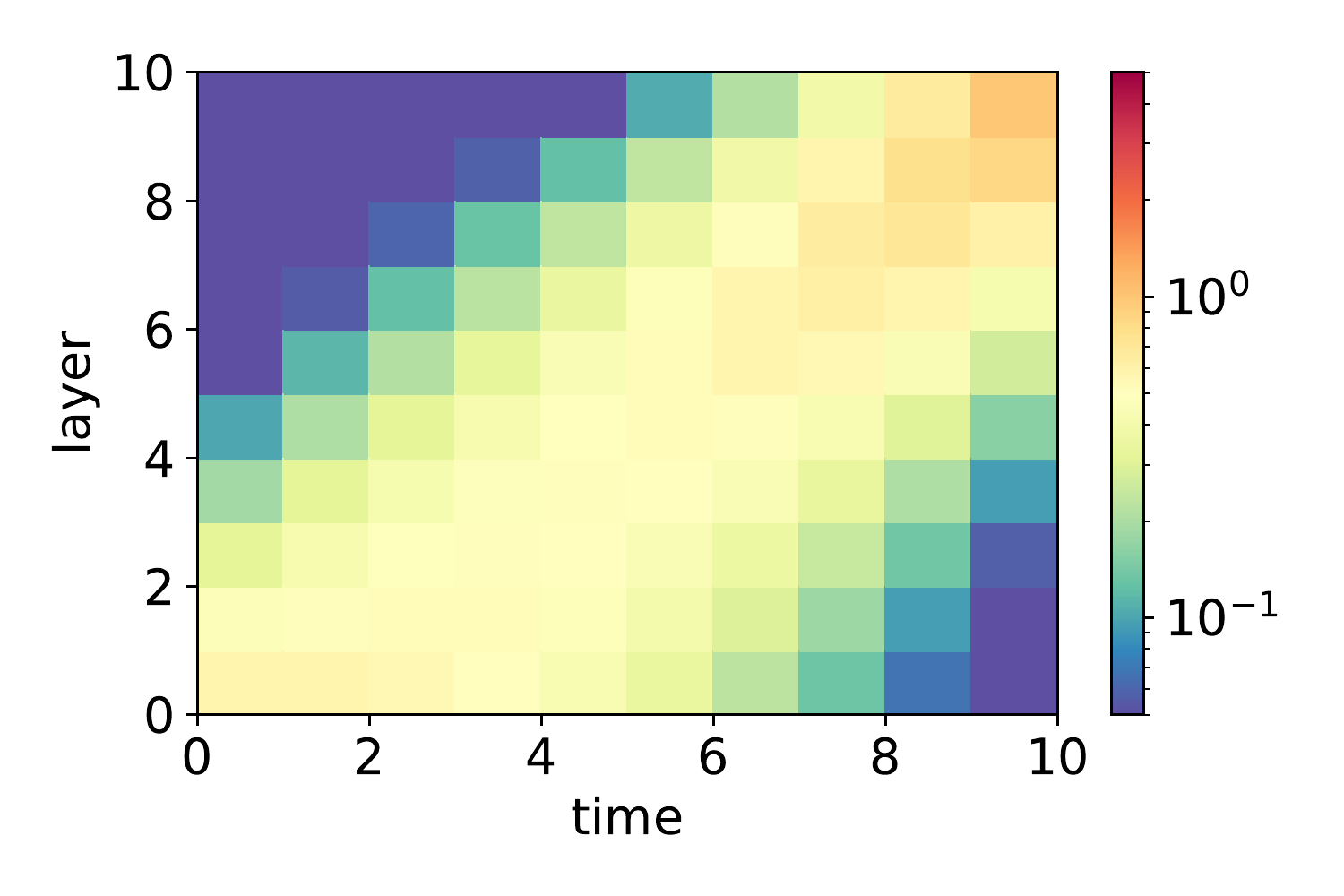} \\
%\rotatebox{0}{$\qquad\,\,$b)} &
\includegraphics[width=0.33\columnwidth]{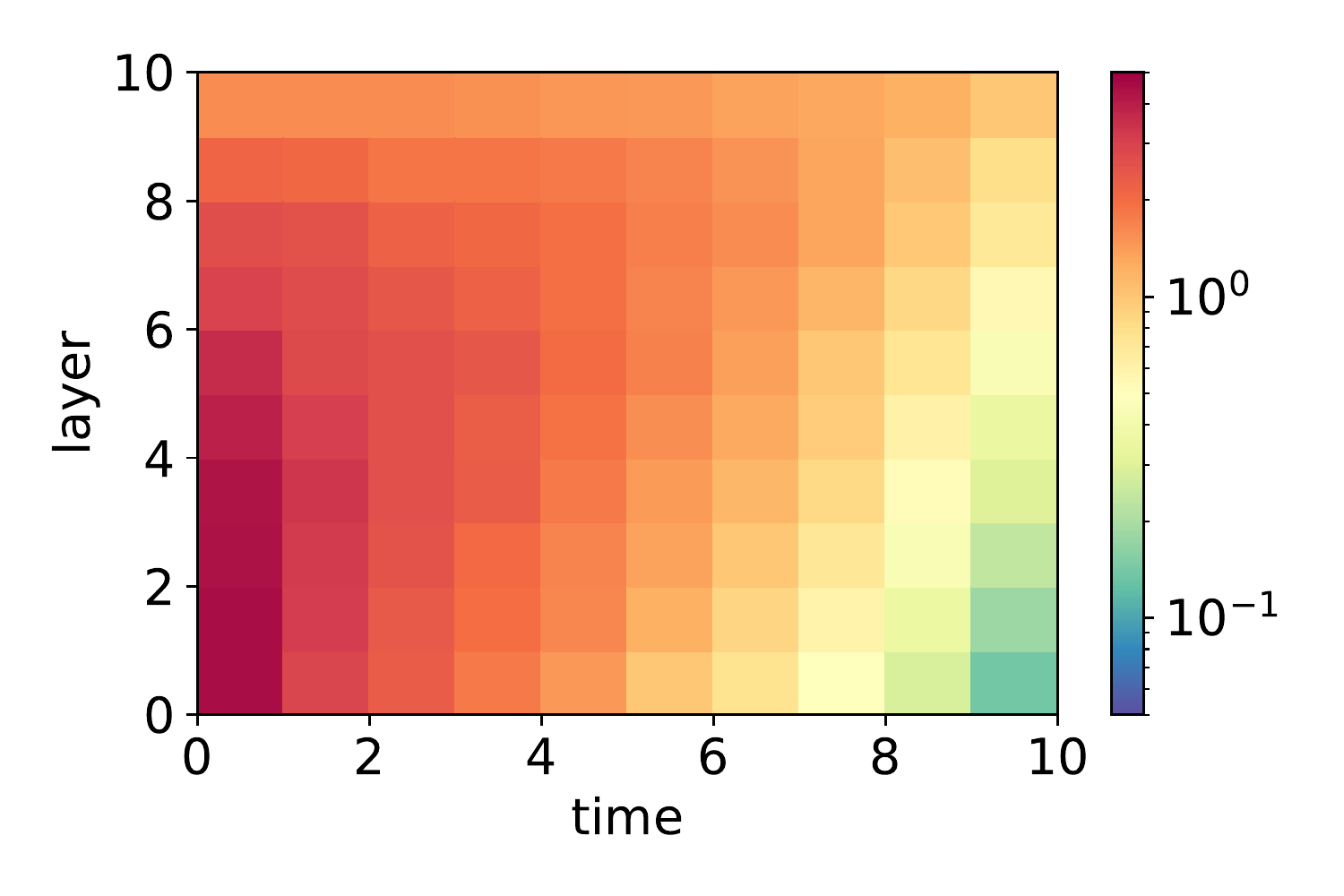} &
\includegraphics[width=0.33\columnwidth]{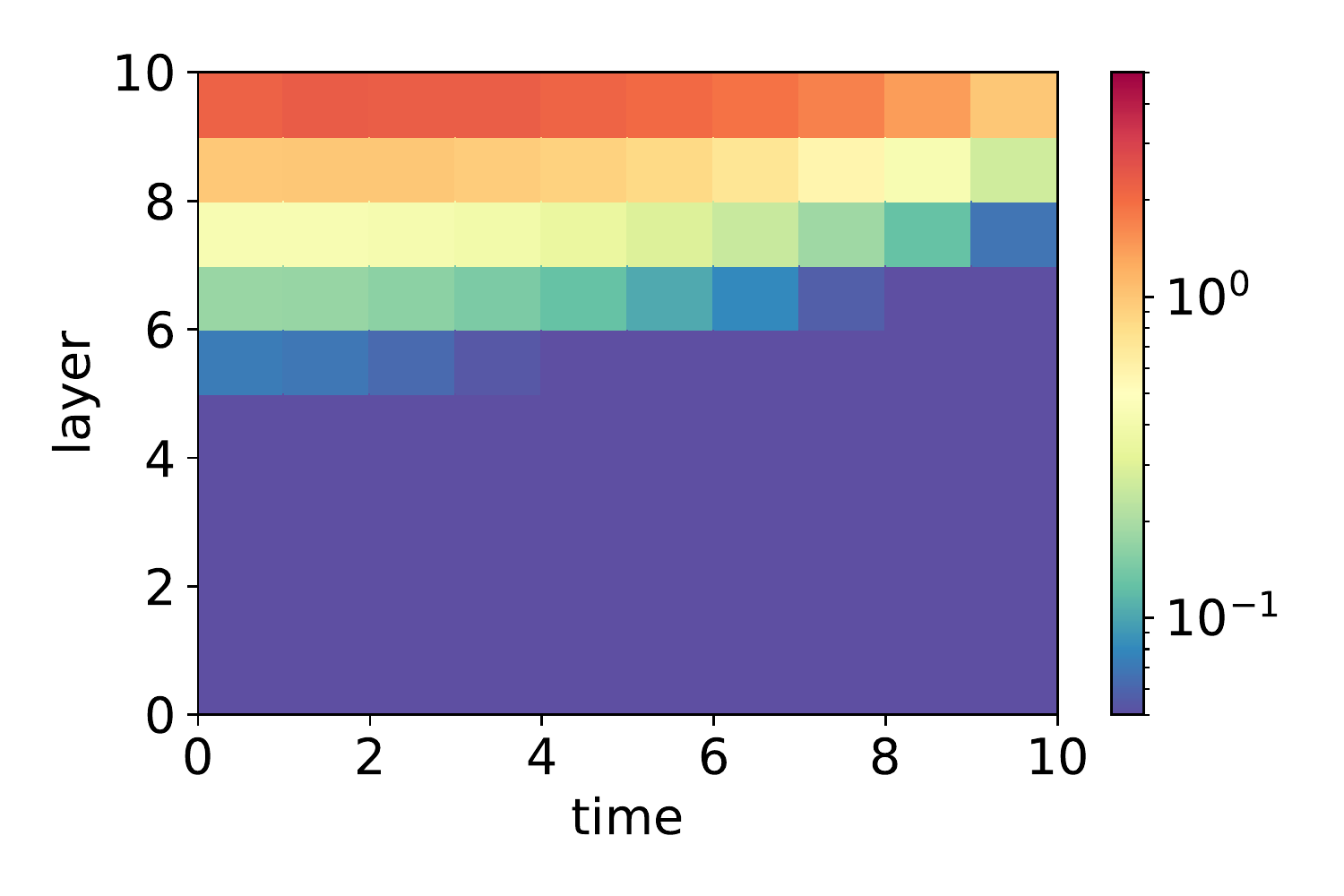} &
\includegraphics[width=0.33\columnwidth]{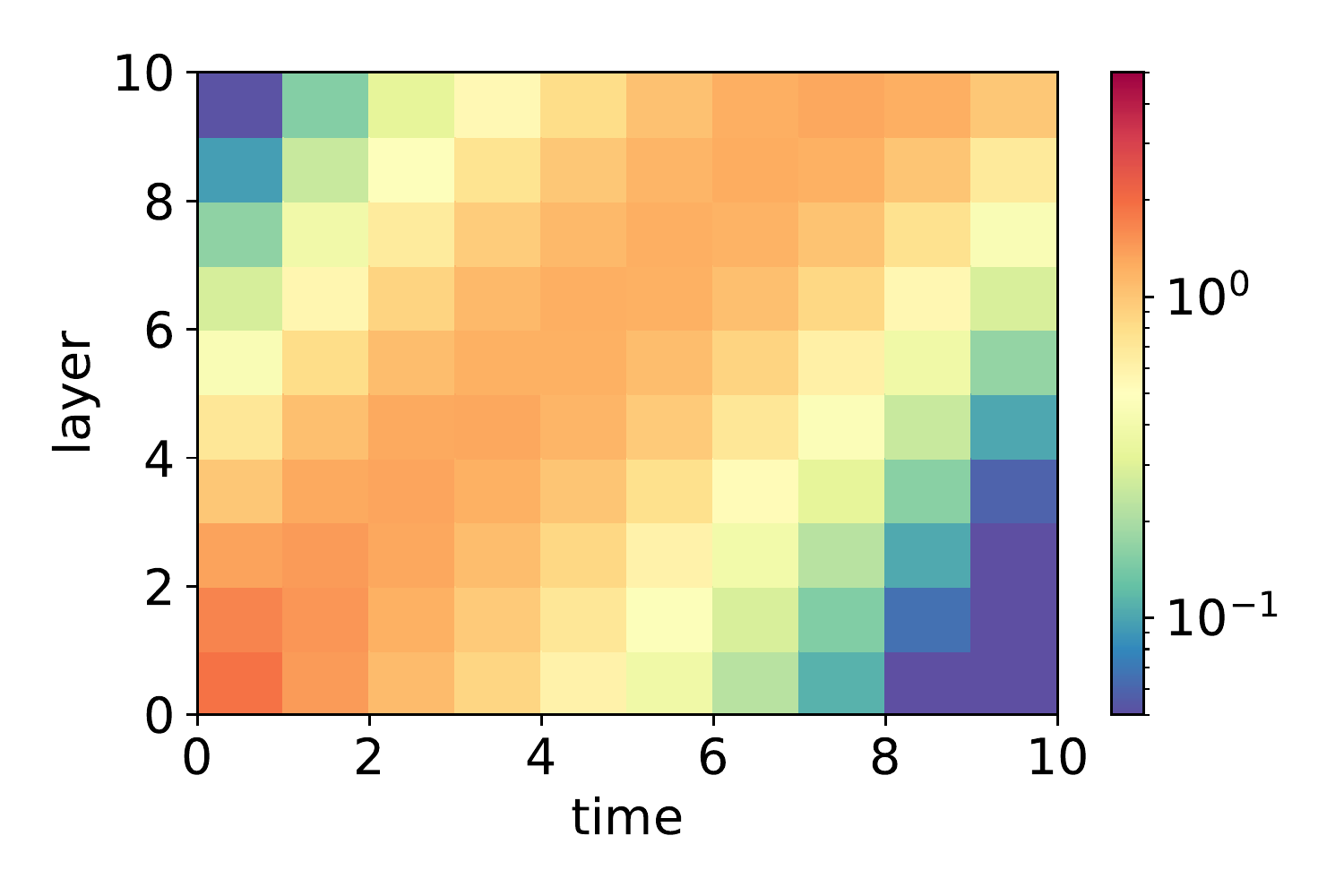} \\
 vRNN & LSTM & STAR \\
\end{tabular}
    \caption{Mean value of gradient magnitude with respect to the parameters for different RNN units. \emph{top row:} loss $\mathcal{L}(h^L_T)$ only on final prediction. \emph{bottom row:} loss $\mathcal{L}(h^L_1\hdots h^L_T)$ over all time steps. As the gradients flow back through time and layers, for a network of vanilla RNN units they get amplified; for LSTM units they get attenuated; whereas the proposed STAR unit approximately preserves their magnitude. See the Appendix for the results with the real data.}
    %\textcolor{red}{Move text below with details of the simulation to text body. Caption is too long and unwieldy.}
    %Hidden size of the networks is set to $4$. Hidden states are initialized with zero vectors and the bias terms are set to zero. Orthogonal matrices are used for all weight matrices. Input sequences are generated with random process ($\vx_t = \alpha \vx_{t-1} + (1-\alpha)\vz$ where $\vz \sim \mathcal{N}(0,1)$). Here $\alpha=0.5$ is a correlation factor; see appendix for results with different $\alpha$ values. Every simulation is run for 1K iterations and with 1K different initialization. So values are averaged over samples (iterations) and different network states.}
    \label{fig:simulation}
\end{figure}

We expect to obtain a gradient $\vg_{\vh_t^l}$ with a larger magnitude by combining the contributions of $\vg_{\vh_t^{l+1}}$ and $\vg_{\vh_{t+1}^l}$. To obtain a more precise estimate of the resulting gradient we should take into account the correlation between the two terms. However, if we examine two extreme cases \emph{(i)} there is no or very small correlation between two gradient contributions, and \emph{(ii)} they are highly (positively) correlated. The scaling factors of vRNN for the gradient are $1.414$ and $2$. respectively. Therefore, regardless of the correlation between the two terms, the gradient of vRNN is systematically growing while it propagates back in time and through layers. A deep network made of vRNN cells with orthogonal or identity initialisation can thus be expected to suffer, especially in the initial training phase, from exploding gradients as we move towards shallower layers and further back in time. To validate this assumption, we set up a toy example of a deep vRNN and compute the average gradient magnitude w.r.t.\ the network parameters for each cell in the unfolded network.
For the numerical simulation we initialise all the hidden states and biases to $0$, and chose random orthogonal matrices for the weights. Input sequences are generated with the random process $\vx_t = \alpha \vx_{t-1} + (1-\alpha)\vz$, where $\vz \sim \mathcal{N}(0,1)$ and the correlation factor $\alpha=0.5$ (the choice of the correlation factor does not seem to qualitatively affect the results).
%(see appendix for results with different $\alpha$ values).
%
Figure~\ref{fig:simulation} depicts average gradient magnitudes over 10K runs with different weight initialisations and input sequences. As expected, the magnitude grows rapidly towards the earlier and shallower part of the network.

We perform a similar analysis for the classical LSTM cell~\cite{lstm}. The recurrent equations of the LSTM cell are the following:
\begin{align}
&\vi_t^l = \sigma(\mW_{xi}\vh_t^{l-1}+ \mW_{hi}\vh_{t-1}^l+ \vb_i)\\
&\vf_t^l = \sigma(\mW_{xf}\vh_t^{l-1}+ \mW_{hf}\vh_{t-1}^l+ \vb_f)\\
&\vo_t^l = \sigma(\mW_{xo}\vh_t^{l-1}+ \mW_{ho}\vh_{t-1}^l+ \vb_o)\\
&\vz_t^l = \tanh (\mW_{xz}\vh_t^{l-1}+ \mW_{hz}\vh_{t-1}^l+ \vb_z)\\
&\vc_t^l = \vf_t^l \circ \vc_{t-1}^l + \vi_t^l \circ \vz_t^l\\ 
&\vh_t^l = \vo_t^l \circ \tanh(\vc_t^l),
\end{align}
where $\vi$, $\vf$, $\vo$ are the input, forget, and output gate activations, respectively, $\vc$ is the cell state.
The expressions of the Jacobians are reported in Eqs.~(\ref{eq:J_lstm},\ref{eq:H_lstm}) where $\mD_\vx$ again denotes a diagonal matrix with the elements of vector $\vx$ as diagonal entries. The equations are slightly more complicated, but are still amenable to the same type of analysis. We again choose the same exemplary conditions as for the vRNN above, i.e., hidden states and biases equal to zero and orthogonal weight matrices. 
%
%his time we additionally need the value of the sigmoid activation at the considered point, $\sigma(0)=0.5$. 
By substituting the numerical values in the aforementioned equations, we can see that the sigmoid function causes the expected singular value of the two Jacobians to drop to $0.25$. Contrary to the vRNN cell, we expect that even the two Jacobians combined will produce an attenuation factor well below 1 (considering the same two extreme cases, i.e., uncorrelated and highly correlated, the value is $0.354$ and $0.5$, respectively) such that the gradient magnitude will decline and eventually vanish. We point out that LSTM cells have a second hidden state, the so-called "cell state". The cell state only propagates along the time dimension and not across layers, which makes the overall effect of the corresponding gradients more difficult to analyse. However, for the same reason one would, in a first approximation, expect that the cell state mainly influences the gradients in the time direction, but cannot help the flow through the layers.
Again the numerical simulation results support our hypothesis as can be seen in Fig.~\ref{fig:simulation}. The LSTM gradients propagate relatively well backward through time, but vanish quickly towards shallower layers.
%
%\textcolor{blue}{We refer to the appendix for further numerical analysis, e.g., LSTMs with only a forget gate, and GRUs.}\textcolor{red}{do we still plan to put this in the supplementary?}

In summary, the gradient propagation behaves differently in time and depth directions. When considering the latter we need to take into consideration the gradient of the output w.r.t.\ the input state, too, and not exclusively consider the gradient w.r.t.\ the previous hidden state. Moreover, we need to take into account that the output of each cell is connected to \emph{two} cells rather then one adjacent cell. Note that this analysis is valid both when the loss is computed only using the final state $T$, and when all states are used (Fig.~\ref{fig:simulation}). In the latter case, we simply need to sum the contribution of all the separate losses. 
Usually, parameters are shared among different times $t$ in RNNs, but not among different layers. If parameters are shared among different time steps, gradients accumulate row-wise (Fig.~\ref{fig:simulation}) increasing the gradient magnitude w.r.t.\ the parameters. This, however, is not true in the vertical direction as weights are not shared. As a consequence, it is particularly important to ensure that the gradient magnitude is preserved between adjacent layers.

%\textcolor{blue}{
%Here, we briefly draw some connections between our analysis and the empirical results of~\cite{gatedFeedback}, who propose a gated feedback RNN (GFRNN) that extends the stacked RNN architecture with extra connections between adjacent layers.
%
%According to the empirical results shown in~\cite{gatedFeedback}, GFRNN improves a 3-layer LSTM, but degrades the vanilla RNN performance. We conjecture that this might be due to the extra connections strengthening the gradient propagation. According to our findings, the additional gradient flow would benefit the LSTM, by bolstering the dwindling gradients; whereas for the vRNN, where the initial gradients are already too high, the added flow might be counterproductive.} \textcolor{red}{move to supplementary}

\begin{figure*}%[t]
    \centering
        \includegraphics[width=1.95\columnwidth]{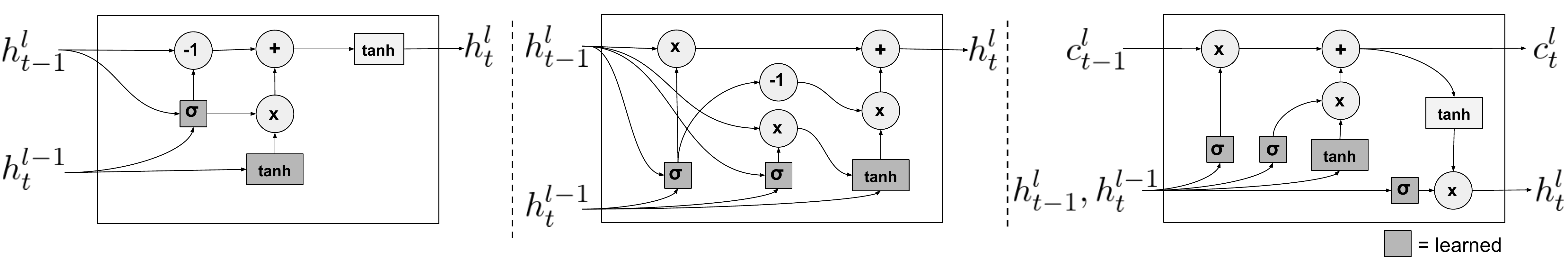}
    \caption{RNN cell structures: STAR, GRU and LSTM, respectively.}
    \label{fig:rnns}
\end{figure*}

\section{The STAR Unit}\label{sec:star}
%Start with LSTM equations
%Compute Jacobian
%Then simplification and justification - some intuitive explanation

%As described in the previous section we believe that the %Jacobian matrices $H^{l}_t$ and $J^{l}_t$ play a %fundamental role in the problem of vanishing and exploding %gradient. In this section we introduce a novel RNN cell %specifically designed to alleviate this problem.

%$\frac{\partial \vh_t^{l}}{\partial \vh_t^{l-1}}$ is the main factor for gradient flowing from upper to lower layers. In the analysis, 
 
%we then re-design the LSTM cell to enforce a largest Jacobian norm.
%$\frac{\partial \vh_t^{l}}{\partial \vh_t^{l-1}}$ for the LSTM is following.
%\begin{equation}
%\frac{\partial \vh_t^{l}}{\partial \vh_t^{l-1}} = \mD_{\tanh(\vc_t^l)}\mD_{(\vo_{t}^{l})'}\mW_{xo} +  \mD_{\tanh(\vc_t^l)'}\mD_{\vo_t^l}(\mD_{\vc_{t-1}^l}\mD_{(\vf_t^l)'}\mW_{xf} + \mD_{\vz_t^l}\mD_{(\vi_t^l)'}\mW_{xi} + \mD_{\vi_t^l}\mD_{(\vz_t^l)'}\mW_{xz} )
%\end{equation}
Building upon the previous analysis, we introduce a novel RNN cell designed to avoid vanishing or exploding gradients while reducing the number of parameters. We start from the Jacobian matrix of the LSTM cell and investigate what design features are responsible for such low singular values.
We see in Eq.~(\ref{eq:H_lstm}) that every multiplication with $\tanh$ non-linearity ($\mD_{\tanh(.)}$), gating functions ($\mD_{\sigma(.)}$), and with their derivatives can only ever decrease the singular values of $\mW$, since all those terms are always $<$1.
% $||\mD_{\tanh(.)}||\leq 1$, $||\mD_{\tanh(.)'}||\leq 1$, $||\mD_{\sigma(.)}||\leq 1$ and 
The effect is particularly pronounced for the sigmoid and its derivative,  $|\sigma'(\cdot)| \leq 0.25$ and $\mathbb{E} [|\sigma(x)|] = 0.5$ for zero-mean, symmetric distribution of $x$. In particular, the output gate $\vo_t^l$ is a sigmoid and plays a major role in shrinking the overall gradients, as it multiplicatively affects all parts of both Jacobians.
As a first measure, we thus propose to remove the output gate, which leads to $\vh^{l}_{t}$ and $\vc^{l}_{t}$ carrying the same information (the hidden state becomes an element-wise non-linear transformation of the cell state). To avoid this duplication and further simplify the design, we transfer the $\tanh$ non-linearity to the hidden state and remove the cell state altogether.

%\textcolor{red}{
As a final modification, we also remove the input gate $\vi^{l}_{t}$ from the architecture and couple it with the forget gate. We observed in detailed simulations that the input gate harms performance of deeper networks. This finding is consistent with the theory: for an LSTM cell with only the output gate removed, the Jacobians $H_t^l,J^t_l$ will on average have singular values $1$, respectively $0.5$ (under the same conditions of Sec.~\ref{sec:background}). This suggests exploding gradients, which we indeed observe in numerical simulations. Moreover, signal propagation is less stable: state values can easily saturate if the two gates that control flow into the memory go out of sync. The gate structure of RHN~\cite{rhn} is similar to that configuration, and does empirically suffer from exploding, then vanishing, gradient (Fig.~\ref{fig:grads_2}).

More formally, our proposed STAR cell in the $l$-th layer takes the input $\vh_{t}^{l-1}$ (in the first layer, $\vx_{t}$) at time $t$ and non-linearly projects it to the space where the hidden vector $\vh^{l}$ lives,~\eqref{eq:star_z}. Furthermore, the previous hidden state and the new input are combined into the gating variable $\vk^{l}_{t}$ (\eqref{eq:star_k}). $\vk^{l}_{t}$ is our analogue of the forget gate and controls how information from the previous hidden state and the new input are combined into a new hidden state.
%
%One could also intuitively interpret $\vk^{l}_{t}$ as a sort of "Kalman gain": if it is large, the new observation is deemed reliable and dominates; otherwise the previous hidden state is conserved.
%Finally the $\vk^{l}_{t}$ and $\vz_t^l$ are combined to create the new state.
The complete dynamics of the STAR unit is given by the expressions
\begin{align}
\vz_t^l &=  \tanh(\mW_z \vh_{t}^{l-1} + \vb_z) \label{eq:star_z}\\ %\psi(\vh_{t}^{l-1}) =
%\end{equation}
%\begin{equation}
\vk_{t}^l &=  \sigma( \mW_x\vh_{t}^{l-1} + \mW_h\vh_{t-1}^l + \vb_k )  \label{eq:star_k}\\ % \phi(\vh_{t}^{l-1},\vh_{t-1}^l ) =
%\end{equation}
%\begin{equation}
\vh_{t}^l &= \tanh \big( (1 - \vk_{t}^l)\circ \vh_{t-1}^l + \vk_{t}^l\circ \vz_t^l \big). \label{eq:star_h}
\end{align}
%
%
%These equations lead to the following Jacobian matrices:
%\begin{align}
%J_t^l &= \mD_{\tanh(\vh_{t-1}^l + \vk_{t}^l\circ(\vz_t^l-\vh_{t-1}^l))'}(\mD_{\vz_t^l-\vh_{t-1}^l}\mD_{(\vk_t^l)'}\mW_{x}  + \mD_{\vk_t^l}\mD_{(\vz_t^l)'}\mW_{z} )\\
%H_t^l &= \mD_{\tanh(\vh_{t-1}^l + \vk_{t}^l\circ(\vz_t^l-\vh_{t-1}^l))'} (\mI + \mD_{\vz_t^l-\vh_{t-1}^l} \mD_{(\vk_t^l)'}\mW_h - \mD_{\vk_t^l} ). % \mD_{\vz_t^l-\vh_{t-1}^l}\mD_{(\vk_t^l)'}\mW_{x}  + \mD_{\vk_t^l}\mD_{(\vz_t^l)'}\mW_{z} )
%\end{align}
%
The Jacobian matrices for the STAR cell can be computed similarly to how it is done for the vRNN and LSTM (see the Appendix).
In this case each of the two Jacobians has average singular values equal to $0.5$. In the same two extreme cases previously considered, the scaling factor for the gradient becomes $0.707$ and $1$, respectively. Even if the gradient decays in the first case (worst case scenario, no correlation between two gradient contributions), it does so more slowly compared to LSTM. In the second case, the gradient can propagate without decaying or amplifying which is the ideal scenario. 
Empirically we have observed that, for an arbitrary STAR cell in the grid, these two terms are highly positively correlated, leading, ultimately, to a gradient scaling factor close to one. We repeat the same numerical simulations as above for the STAR cell, and find that it indeed maintains healthy gradient magnitudes throughout most of the deep RNN (Fig.~\ref{fig:simulation}). 
%
%This puts them between the vRNN cell and the LSTM cell, and, when added together, approximately preserve the gradient magnitude. We repeat the same numerical simulations as above for the STAR cell, and find that it indeed maintains healthy gradient magnitudes throughout most of the deep RNN (Fig.~\ref{fig:simulation}). 
%
%We have observed empirically that these two terms are highly correlated. If we take an arbitrary STAR cell in the grid, compute both terms separately and calculate the average cosine similarity between two gradient vectors ($\cos{\theta}= a^{T}b/(|a||b|)$), we obtain  $\displaystyle  \E [\cos{\theta}] \approx 0.98$. This finding is a strong indicator that both vectors are highly correlated, ultimately leading to a gradient scaling factor close to one. Moreover, it is (empirical) evidence that our STAR cell can retain the gradient magnitude nearly perfectly, which is supported by simulations with synthetic and real (MNIST) data. In conclusion, STAR has better theoretical bounds for gradient scaling factors compared to vRNN and LSTM. In practice, the gradient can propagate more stably back in time and through the different layers. 
%
Finally, we point out that our proposed STAR architecture requires significantly less memory than most alternative designs. For the same input and hidden state size, STAR has a  $50\%$, respectively and $60\%$ smaller memory footprint than GRU and LSTM. %$50\%$ larger compared to vRNN but also 
In the next section, we experimentally validate on real datasets that deep RNNs built from STAR units can be trained to a significantly greater depth while performing on par or better than state-of-the-art despite having fewer parameters.

\begin{figure}
    \centering
    \begin{subfigure}[b]{0.5\textwidth}
        \includegraphics[width=\textwidth]{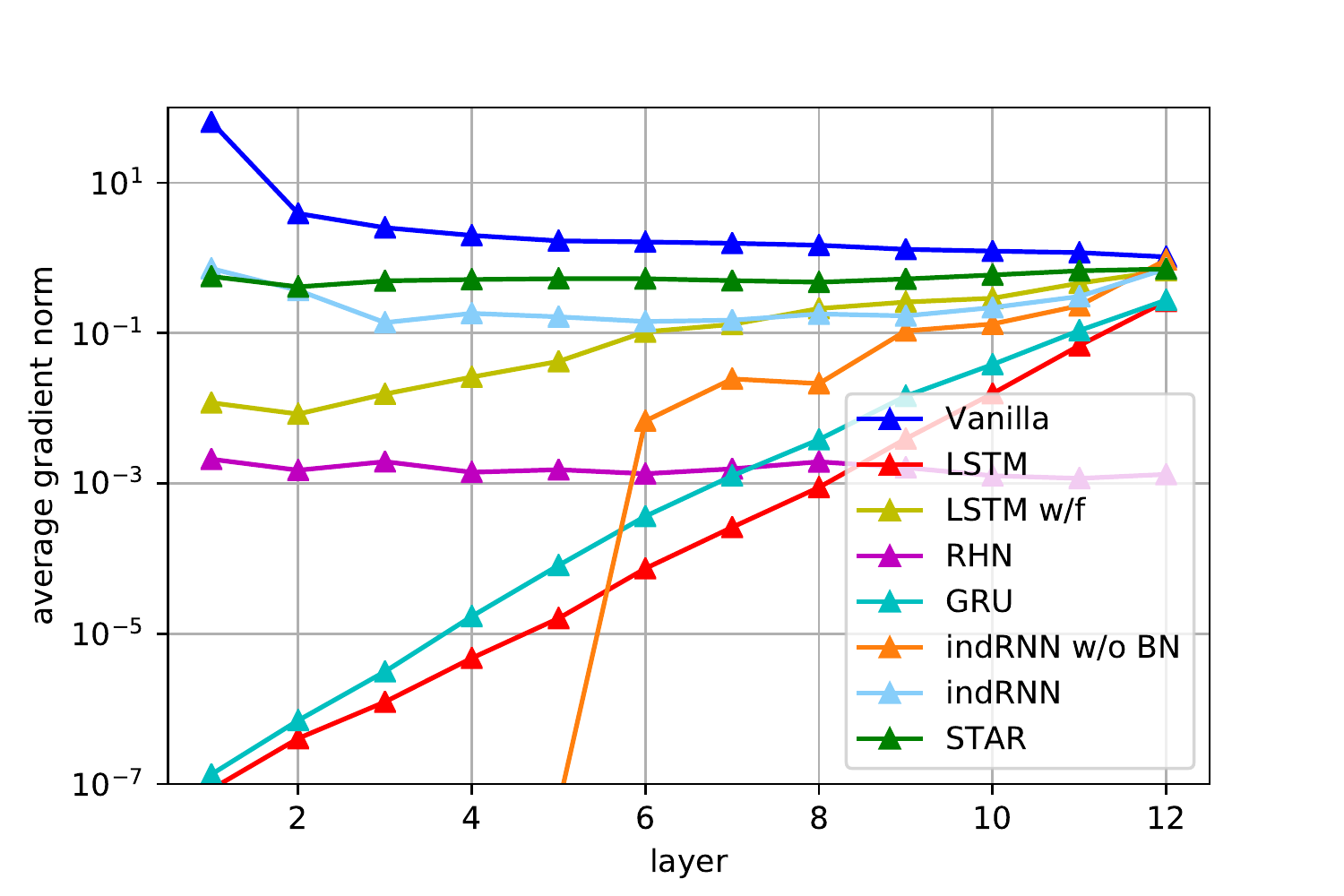}
        \caption{gradient norm versus layer}
        \label{fig:grads_1}
    \end{subfigure}  
      \begin{subfigure}[b]{0.5\textwidth}
        \includegraphics[width=\textwidth]{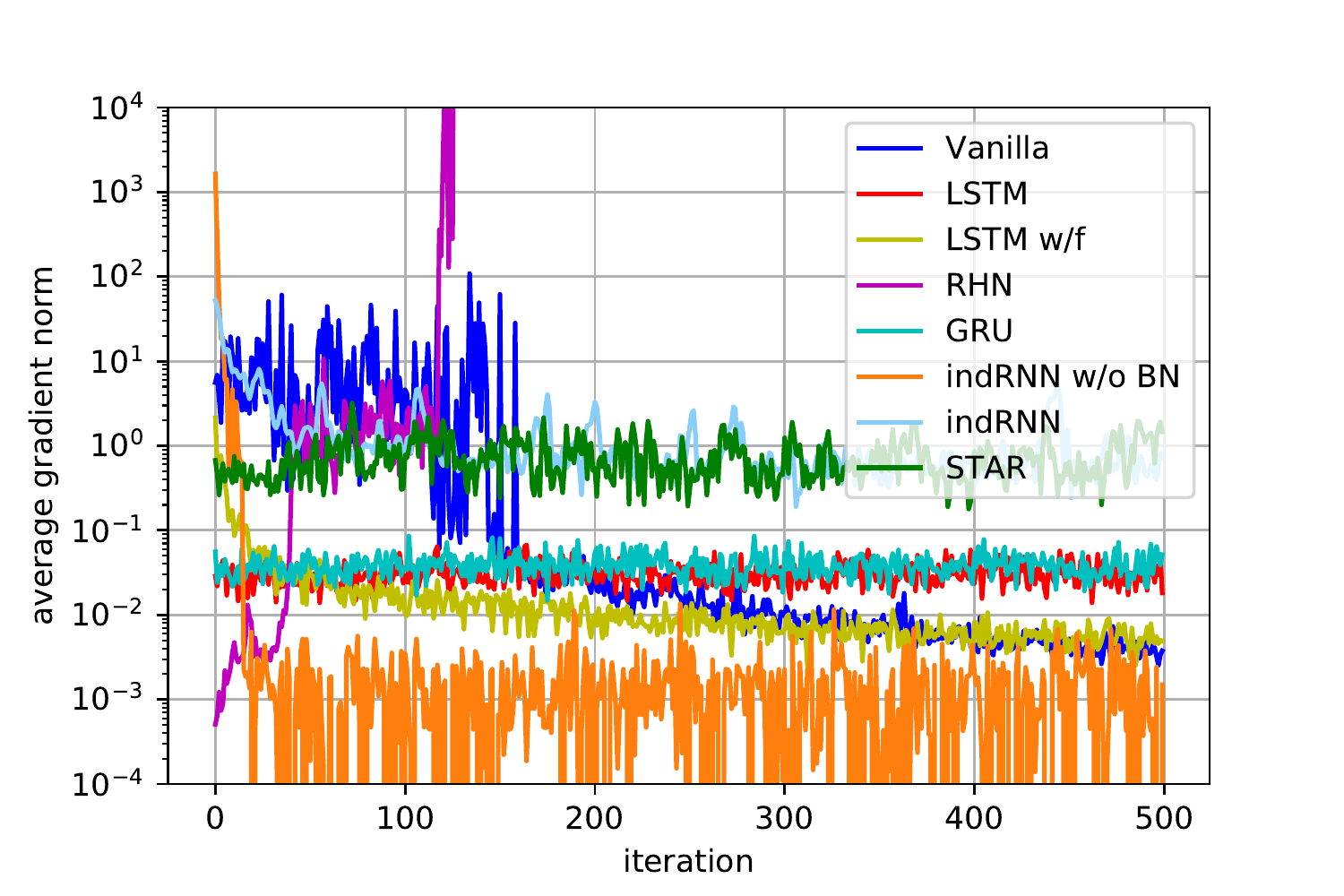}
        \caption{gradient norm versus iteration, 1$^\text{st}$ epoch}
        \label{fig:grads_2}
    \end{subfigure}
    \begin{subfigure}[b]{0.5\textwidth}
        \includegraphics[width=\textwidth]{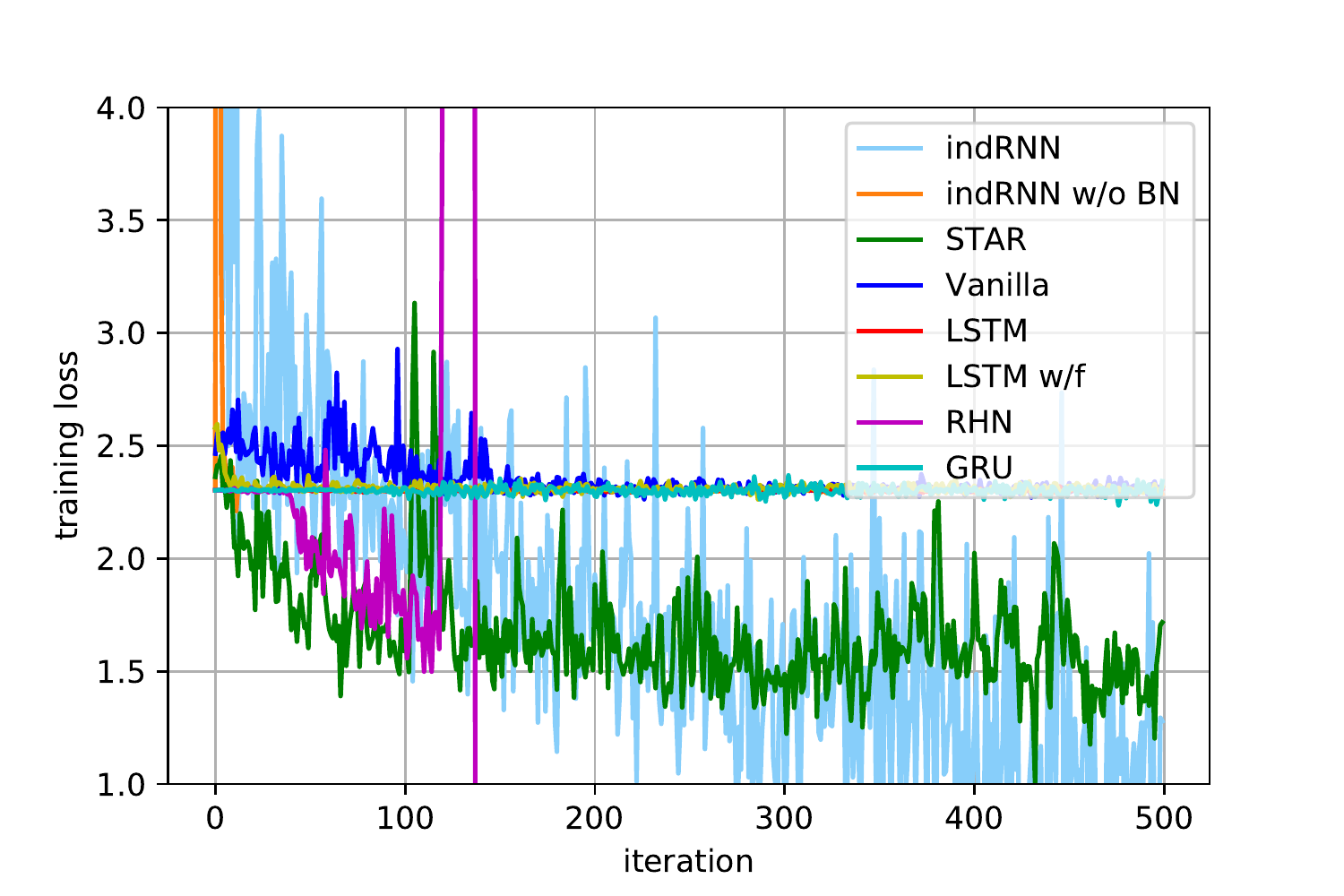}
        \caption{training loss versus iteration, 1$^\text{st}$ epoch}
        \label{fig:grads_3}
    \end{subfigure}  
    \caption{Gradient magnitudes of pix-by-pix MNIST. (a) Mean gradient norm per layer at the start of training. (b) Evolution of gradient norm during 1$^\text{st}$ training epoch. (c) Loss during 1$^\text{st}$ epoch.}\label{fig:grads}
\end{figure}
\section{Experiments}\label{experiments}
We evaluate the performance of several well-known RNN cells as well as that of the proposed STAR cell on different sequence modelling tasks with ten different datasets: sequential versions of \emph{MNIST}~\cite{mnist}, the adding~\cite{lstm}, and copy memory~\cite{vanishing_bengio} problems, music modeling~\cite{jsb_dataset,piano_dataset}, character-level language modeling~\cite{ptb}, which are a common testbeds for recurrent networks; three different remote sensing datasets, where time series of intensities observed in satellite images shall be classified into different agricultural crops~\cite{fieldrnn,tum_image,breizhcrops}; and Jester~\cite{jester} for hand gesture recognition. We use convolutional layers for gesture recognition and pixel-wise crop classification, whereas we employ conventional fully connected layers for the other tasks. 
The recurrent units we compare include the vRNN, the LSTM, the LSTM with only a forget gate~\cite{UnEffForGate}, the GRU, the RHN \cite{rhn}, IndRNN~\cite{indRNN}, temporal convolution network (TCN)~\cite{tcn}, transformer~\cite{transformer}, and the proposed STAR. The experimental protocol is similar for all tasks: For each model variant, we train multiple versions with different depth (number of layers) and the best performing one is picked.
%For each variant and each depth, we report the performance of the model with the lowest validation loss.
Classification performance is measured by the rate of correct predictions (top-1 accuracy) for classification tasks, bits per character for character-level language modeling task and negative log-likelihood (NLL) for the rest of the tasks. Throughout the different experiments, we use orthogonal initialisation for weight matrices of RNNs. Training and network details for each experiment can be found in the Appendix.\footnote{ Code and trained models (in Tensorflow), as well as code for the simulations (in PyTorch), are available online: \href{https://github.com/0zgur0/STAR_Network}{https://github.com/0zgur0/STAR\_Network}.}

%We have made the code available at \href{https://github.com/0zgur0/STAR_Network}{https://github.com/0zgur0/STAR\_Network}.

\subsection{Pixel-by-pixel MNIST}
We flatten all 28$\times$28 grey-scale images of handwritten digits of the MNIST dataset~\cite{mnist} into 784$\times$1 vectors, and the 784 values are sequentially presented to the RNN. The models' task is to predict the digit after having seen all pixels. The second task, pMNIST, is more challenging. Before flattening the images pixels are shuffled with a fixed random permutation, turning correlations between spatially close pixels into non-local long-range dependencies. As a consequence, the model needs to remember dependencies between distant parts of the sequence to classify the digit correctly.
Fig.~\ref{fig:grads_1} shows the average gradient norms per layer at the start of training for 12-layer networks built from different RNN cells. Propagation through the network increases the gradients for the vRNN and shrinks them for the LSTM. As the optimisation proceeds, we find that STAR and IndRNN remain stable, whereas all other units see a rapid decline of the gradients already within the first epoch, except for RHN, where the gradients explode, see Fig.~\ref{fig:grads_2}. Consequently, STAR and IndRNN are the only units for which a 12-layer model can be trained, as also confirmed by the evolution of the training loss, Fig.~\ref{fig:grads_3}. 

IndRNN's gradient propagation through layers is also stable even though not as good as STAR's. However, IndRNN strongly relies on Batch Normalization (BN)~\cite{batch_norm} for stable gradient propagation through layers while STAR does not require BN. If we remove the BNs between consecutive layers in IndRNN (denoted IndRNN w/o BN), its gradient propagation through layers and iterations becomes very unstable (see Fig. \ref{fig:grads_1} and \ref{fig:grads_2}). Indeed, IndRNN cannot be trained in those cases. It does not only fail for deeper, 12-layer setups applied to sequential MNIST, but also for shallower designs. Apart from increasing the computation overhead, general drawbacks of IndRNN's dependency on BNs are: (i) slow convergence during training and (ii) poor performance during inference if batch size is small (see the Appendix for further quantitative analysis).

Fig.~\ref{fig:pmnist_bn_added} confirms that stacking into deeper architectures does benefit RNNs (except for vRNN); but it increases the risk of a catastrophic training failure. STAR is significantly more robust in that respect and can be trained up to $>$20 layers. On the comparatively easy and saturated MNIST data, the performance is comparable to a successfully trained LSTM (at depth 2-8 layers, LSTM training sometimes catastrophically fails; the displayed accuracies are averaged only over successful training runs).

In~\ref{table:SOTA} we show that our STAR cell mostly outperforms existing methods. As STAR is specifically designed to improve gradient propagation in the vertical direction, we conduct one additional experiment with a hybrid architecture: we use LSTM with a forget gate (which achieves good performance on the MNIST dataset in the one layer case) as first layer of the network and we stack seven layers of STAR cells on top. Such a design increases the capacity of the first layer without endangering gradient propagation. This further improves accuracy for both MNIST and pMNIST, leading to on par performance across both tasks with the best state-of-the-art methods BN-LSTM~\cite{bn_lstm} and IndRNN~\cite{indRNN}. Both methods employ Batch Normalization~\cite{batch_norm} inside the cells to improve the performance wrt to the simpler form of LSTM and IndRNN. We tested a version of the STAR cell which used BN and also in this case the modification lead to some performance improvements. This modification, however, is rather general and independent of the cell architecture, as it can be added to most of the other existing methods.

\begin{figure}
    \centering
    \begin{subfigure}[b]{0.49\textwidth}
        \includegraphics[width=\textwidth]{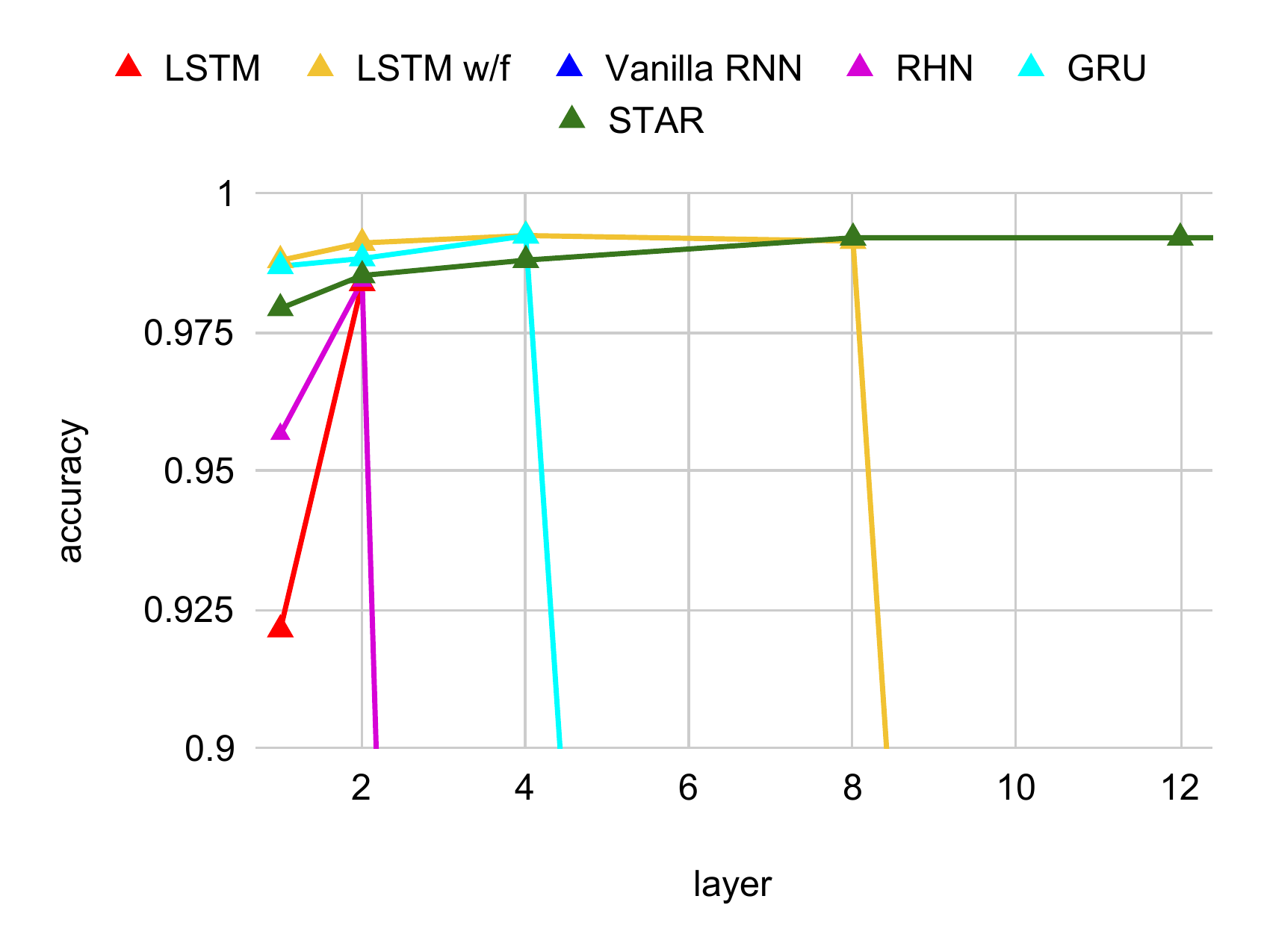}
        \caption{MNIST}
        \label{fig:mnist_1}
    \end{subfigure}  
      \begin{subfigure}[b]{0.49\textwidth}
        \includegraphics[width=\textwidth]{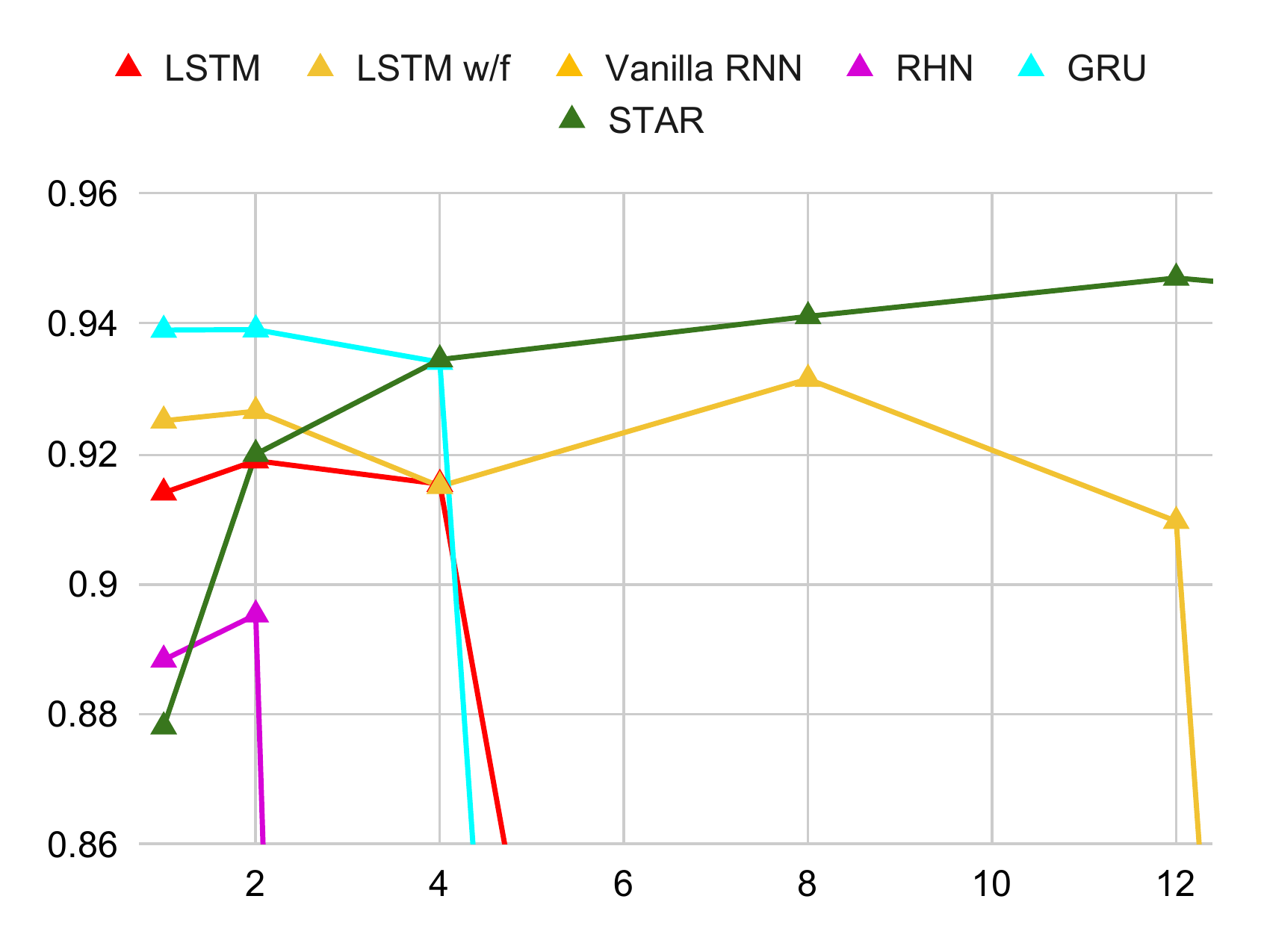}
       \caption{pMNIST}
       \label{fig:mnist_2}
    \end{subfigure}

    \caption{Accuracy results for pixel-by-pixel MNIST tasks. }\label{fig:pmnist_bn_added}
\end{figure}

%\begin{figure}%[t]
%    \centering
%        \includegraphics[width=\columnwidth]{new_images/pmnist_bn_star_added.pdf}
%    \caption{\textcolor{red}{Accuracy results for permuted pixel-by-pixel MNIST task.} }
%    \label{fig:pmnist_bn_added}
%\end{figure}

\begin{table}[t]

\begin{center}
\begin{tabular}{llll}
%\multicolumn{1}{c}{\bf Method}  &\multicolumn{1}{c}{\bf MNIST} &\multicolumn{1}{c}{\bf pMNIST} &\multicolumn{1}{c}{\bf \#units} &\multicolumn{1}{c}{\bf \#params}\\
\multicolumn{1}{c}{ Method}  &\multicolumn{1}{c}{ MNIST} &\multicolumn{1}{c}{ pMNIST} &\multicolumn{1}{c}{ units} \\
\midrule

\multicolumn{1}{l}{vRNN (1 layer)}     &\multicolumn{1}{c}{24.3\%} &\multicolumn{1}{c}{44.0\%} &\multicolumn{1}{c}{128} \\

%\multicolumn{1}{l}{LSTM (1 layers)}     &\multicolumn{1}{c}{92.1\%} &\multicolumn{1}{c}{91.4\%} &\multicolumn{1}{c}{128}  \\

\multicolumn{1}{l}{LSTM (2 layers)}     &\multicolumn{1}{c}{98.4\%} &\multicolumn{1}{c}{91.9\%} &\multicolumn{1}{c}{128}  \\

%\multicolumn{1}{l}{GRU (1 layers)}     &\multicolumn{1}{c}{98.7\%} &\multicolumn{1}{c}{93.9\%} &\multicolumn{1}{c}{128} \\

\multicolumn{1}{l}{GRU (2 layers)}     &\multicolumn{1}{c}{98.8\%} &\multicolumn{1}{c}{93.9\%} &\multicolumn{1}{c}{128} \\

\multicolumn{1}{l}{RHN (2 layers)}     &\multicolumn{1}{c}{98.4\%} &\multicolumn{1}{c}{89.5\%} &\multicolumn{1}{c}{128} \\

\multicolumn{1}{l}{iRNN \cite{identitiy_init}}     &\multicolumn{1}{c}{97.0\%} &\multicolumn{1}{c}{82.0\%} &\multicolumn{1}{c}{100} \\

\multicolumn{1}{l}{uRNN \cite{uRNN}}     &\multicolumn{1}{c}{95.1\%} &\multicolumn{1}{c}{91.4\%} &\multicolumn{1}{c}{512}  \\

\multicolumn{1}{l}{FC uRNN \cite{FCuRNN}}     &\multicolumn{1}{c}{96.9\%} &\multicolumn{1}{c}{94.1\%} &\multicolumn{1}{c}{512}  \\

\multicolumn{1}{l}{Soft ortho \cite{softOrth}}     &\multicolumn{1}{c}{94.1\%} &\multicolumn{1}{c}{91.4\%} &\multicolumn{1}{c}{128} \\

\multicolumn{1}{l}{AntisymRNN \cite{antisymRNN}}     &\multicolumn{1}{c}{98.8\%} &\multicolumn{1}{c}{93.1\%} &\multicolumn{1}{c}{128}\\ 

\multicolumn{1}{l}{IndRNN \cite{indRNN}}     &\multicolumn{1}{c}{\textbf{99.0\%}} &\multicolumn{1}{c}{\textbf{96.0\%}} &\multicolumn{1}{c}{128}\\ 

\multicolumn{1}{l}{BN-LSTM \cite{bn_lstm}}     &\multicolumn{1}{c}{\textbf{99.0\%}} &\multicolumn{1}{c}{\textbf{95.4\%}} &\multicolumn{1}{c}{100}\\ 

\multicolumn{1}{l}{sTANH-RNN \cite{stanh_rnn}}     &\multicolumn{1}{c}{98.1\%} &\multicolumn{1}{c}{94.0\%} &\multicolumn{1}{c}{128} 
\\ \midrule
\multicolumn{1}{l}{STAR (8 layers)}     &\multicolumn{1}{c}{99.2\%} &\multicolumn{1}{c}{94.1\%} &\multicolumn{1}{c}{128} \\
\multicolumn{1}{l}{STAR (12 layers)}     &\multicolumn{1}{c}{99.2\%} &\multicolumn{1}{c}{94.7\%} &\multicolumn{1}{c}{128}\\

\multicolumn{1}{l}{\begin{tabular}[c]{@{}l@{}}LSTM w/f STAR \\(8 layers)\end{tabular}}     &\multicolumn{1}{c}{\textbf{\underline{99.4\%}}} &\multicolumn{1}{c}{\textbf{\underline{95.4\%}}} &\multicolumn{1}{c}{128} 

%\multicolumn{1}{l}{\begin{tabular}[c]{@{}l@{}}BN-STAR \\(12 layers)\end{tabular}}     &\multicolumn{1}{c}{-} &\multicolumn{1}{c}{\textbf{\underline{95.7\%}}} &\multicolumn{1}{c}{128}  

\\ \hline 

\end{tabular}
\end{center}
\caption{Performance comparison for pixel-by-pixel MNIST tasks. Our best performing configuration \textbf{\underline{bold underlined}}, top performers state-of-the-art \textbf{bold}.}
\label{table:SOTA}
\end{table}

\subsection{Adding Problem / Copy Memory}
\begin{figure*}
    \centering
    \begin{subfigure}[b]{0.45\textwidth}
        \includegraphics[width=\textwidth]{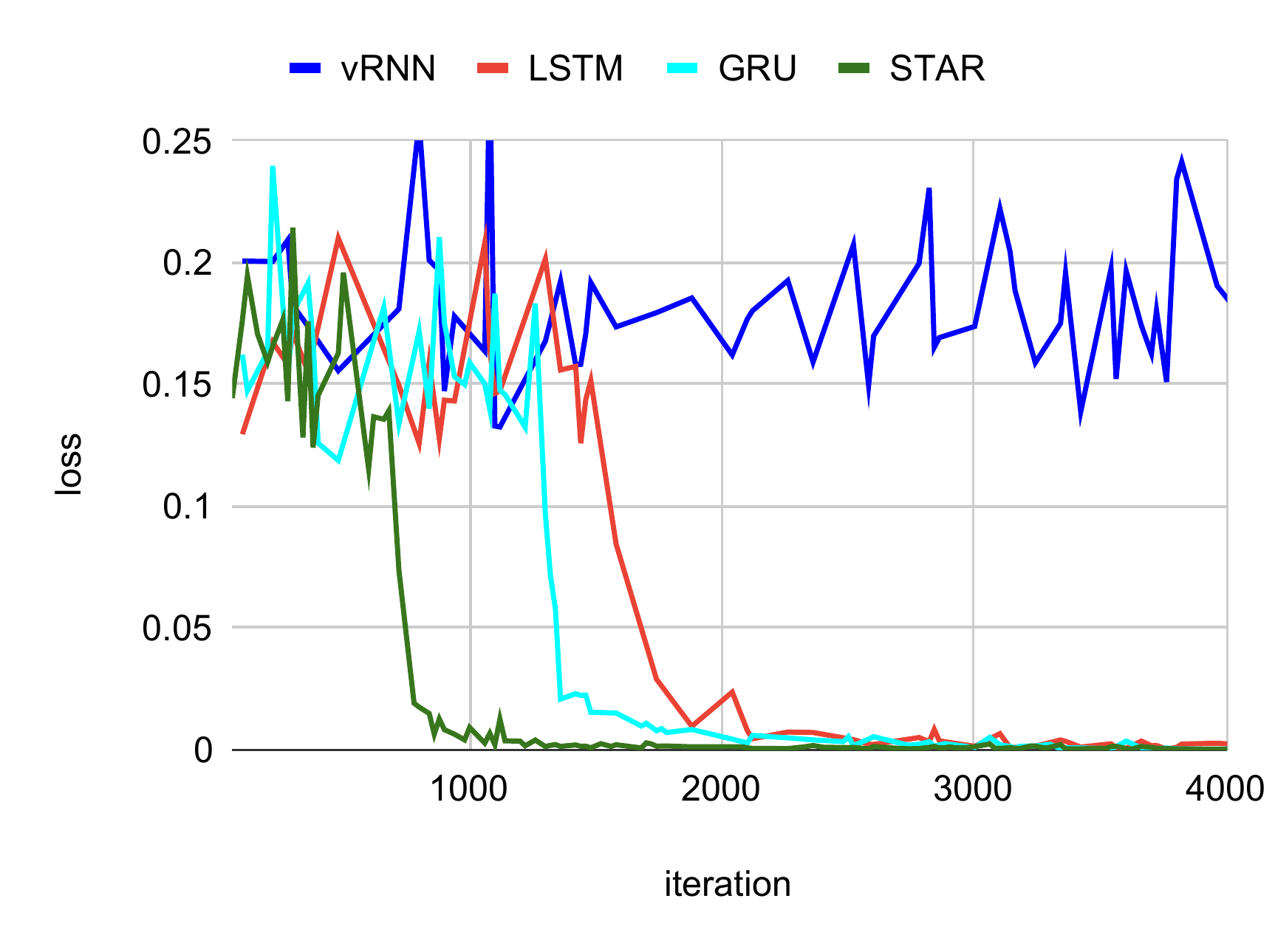}
        \caption{T=200}
        \label{fig:add200}
    \end{subfigure}  
      \begin{subfigure}[b]{0.45\textwidth}
        \includegraphics[width=\textwidth]{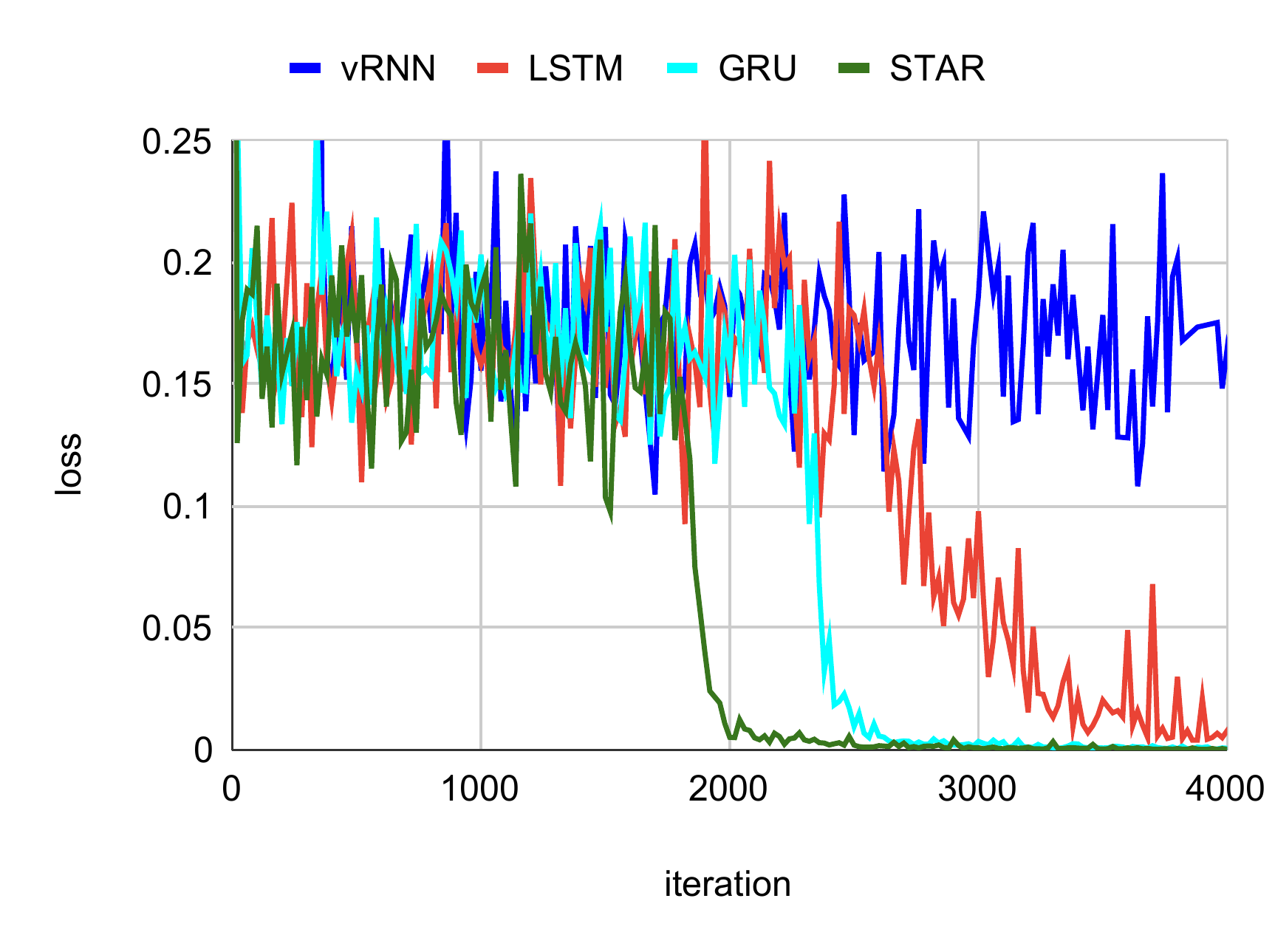}
        \caption{T=1000}
        \label{fig:add1000}
    \end{subfigure}
    \caption{Performance comparison for adding problem.}
    \label{fig:add}
\end{figure*}

\begin{figure*}
    \centering
    \begin{subfigure}[b]{0.45\textwidth}
        \includegraphics[width=\textwidth]{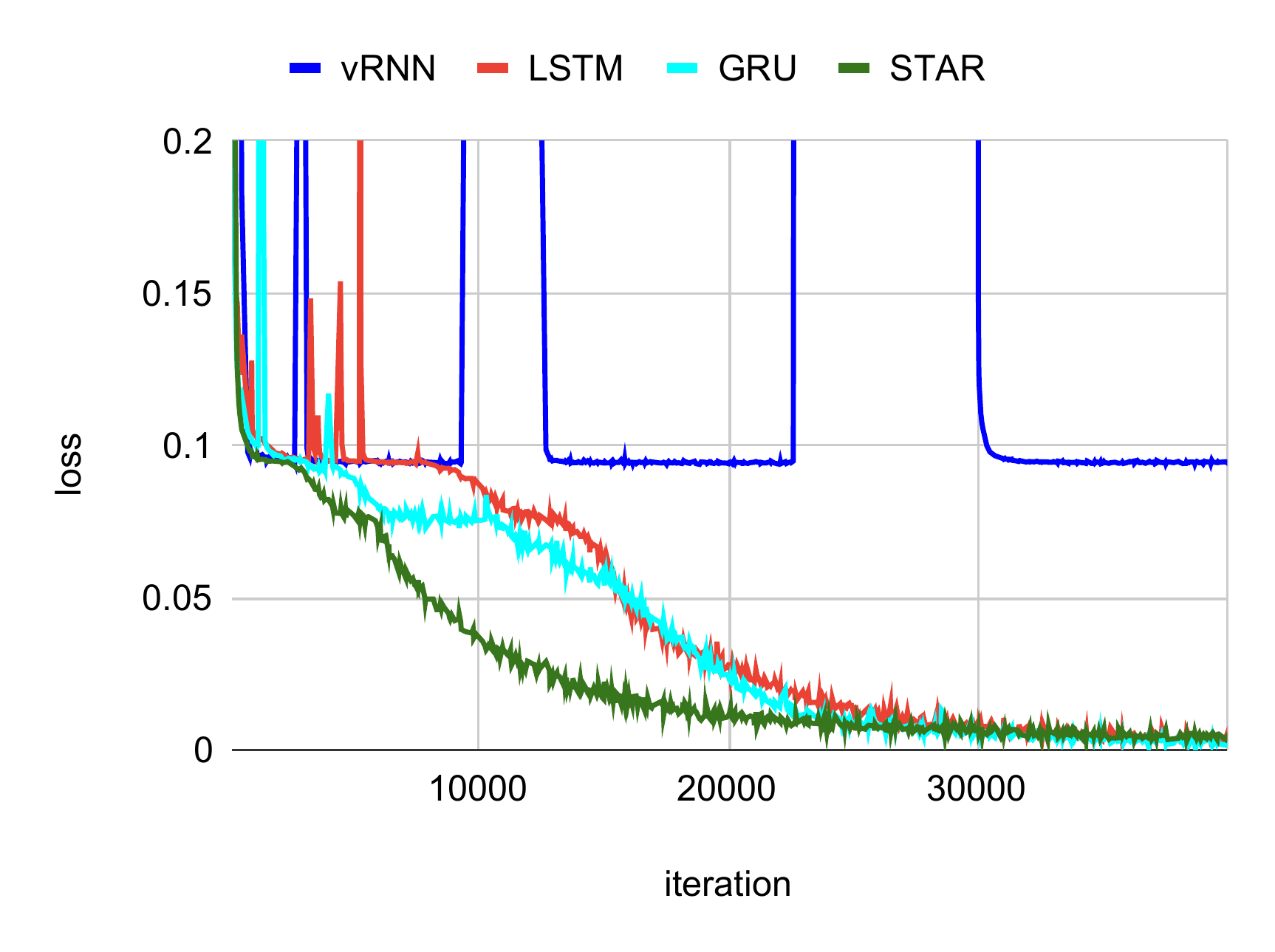}
        \caption{T=200}
        \label{fig:copy200}
    \end{subfigure}  
      \begin{subfigure}[b]{0.45\textwidth}
        \includegraphics[width=\textwidth]{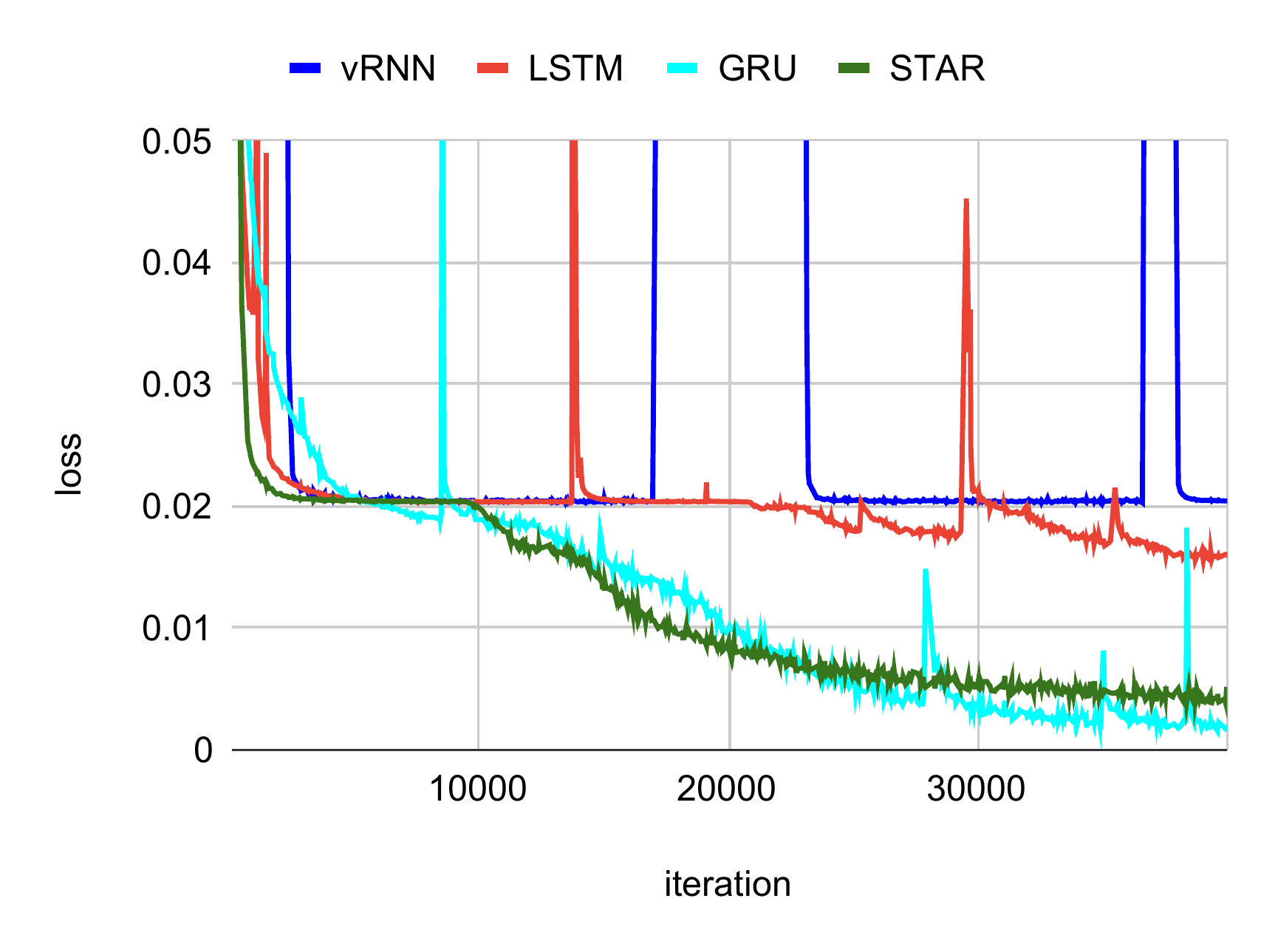}
        \caption{T=1000}
        \label{fig:copy1000}
    \end{subfigure}
    \caption{Performance comparison for copy memory task.}
    \label{fig:copy}
\end{figure*}

The adding problem~\cite{lstm} and the copy memory~\cite{vanishing_bengio} are common benchmarks to evaluate whether a network is able to learn long-term memory. In the adding problem, two sequences of length $T$ are taken as input: the first sequence consists of independent samples in range $(0, 1)$, while the second sequence is a binary vector with two entries set to $1$ and the rest $0$. The goal is to sum the two entries of the first sequence indicated by the two entries of $1$ in the second sequence.
In copy memory task, the input sequence is of length $T + 20$. The first $10$ values in the sequence are chosen randomly among the digits $\{1, . . . , 8\}$, the sequence is then followed by $T$ zeros, the last $11$ entries are filled with the digit $9$ (the first $9$ is a delimiter). The goal is to generate an output of the same length that is zero everywhere except for the last $10$ values after the delimiter, where the model is expected to repeat the $10$ values encountered at the beginning of the input sequence.
We perform experiments with two different sequence lengths, $T=200$ and $T=1000$, using different RNNs with the same number of parameters ($70K$). The results are shown in Fig.~\ref{fig:add}, \ref{fig:copy}. The vRNN is unable to perform long-term memorization, whereas LSTM has issue with longer sequences ($T=1000$). In contrast, both STAR and GRU, can learn long-term memory even when the sequences are very long. An advantage of STAR in this case is its faster convergence.

\subsection{TUM \& BreizhCrops Time Series Classification}\label{TUM_section}
\begin{figure}
    \centering
    \begin{subfigure}[b]{0.49\textwidth}
        \includegraphics[width=\textwidth]{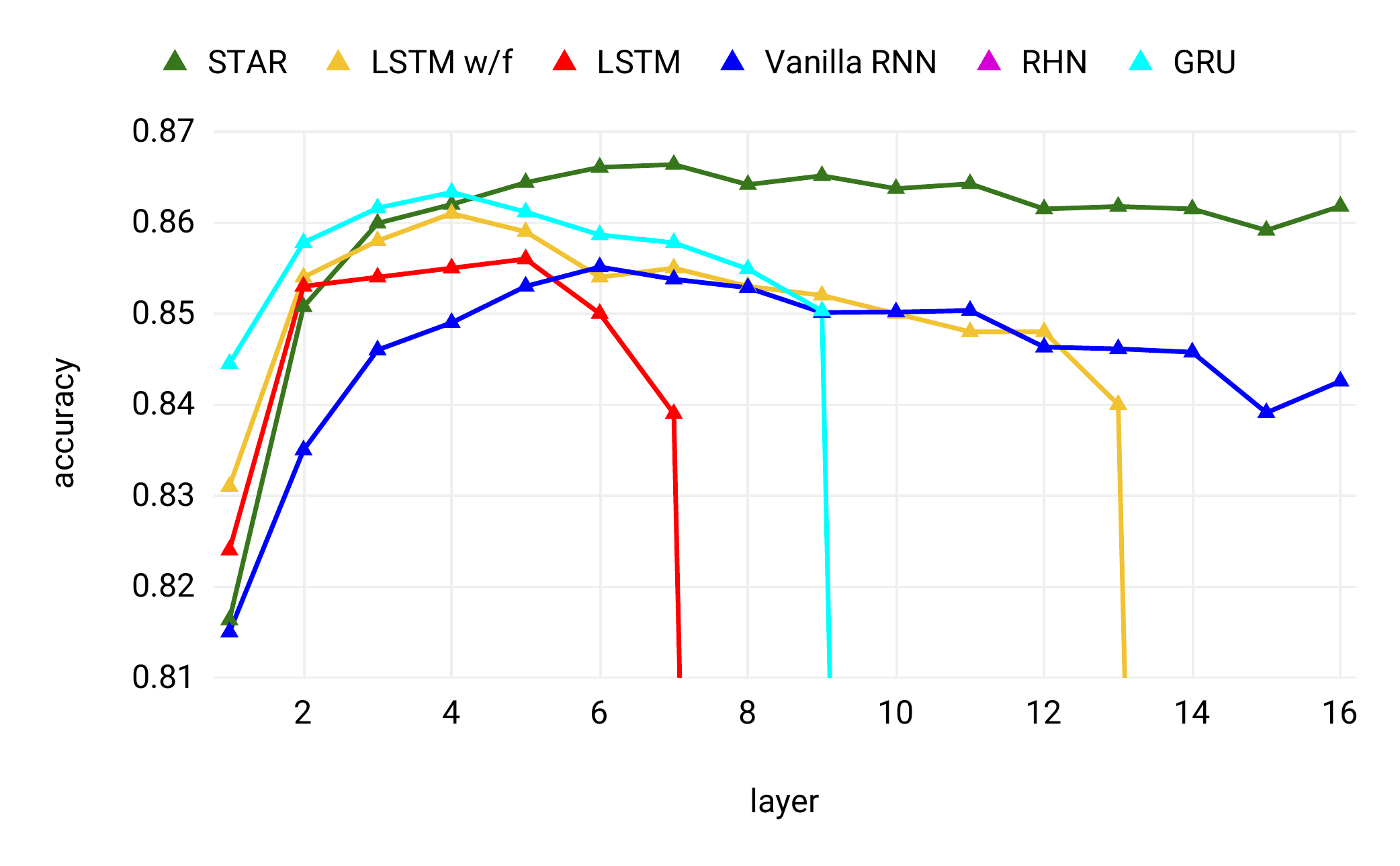}
        \caption{TUM, per time-step labels}
        \label{fig:tum_1}
    \end{subfigure}  
    %  \begin{subfigure}[b]{0.48\textwidth}
    %    \includegraphics[width=\textwidth]{new_images/TUM2.pdf}
    %    \caption{TUM, single label / sequence}
    %    \label{fig:tum_2}
    %\end{subfigure}
    \begin{subfigure}[b]{0.49\textwidth}
        \includegraphics[width=\textwidth]{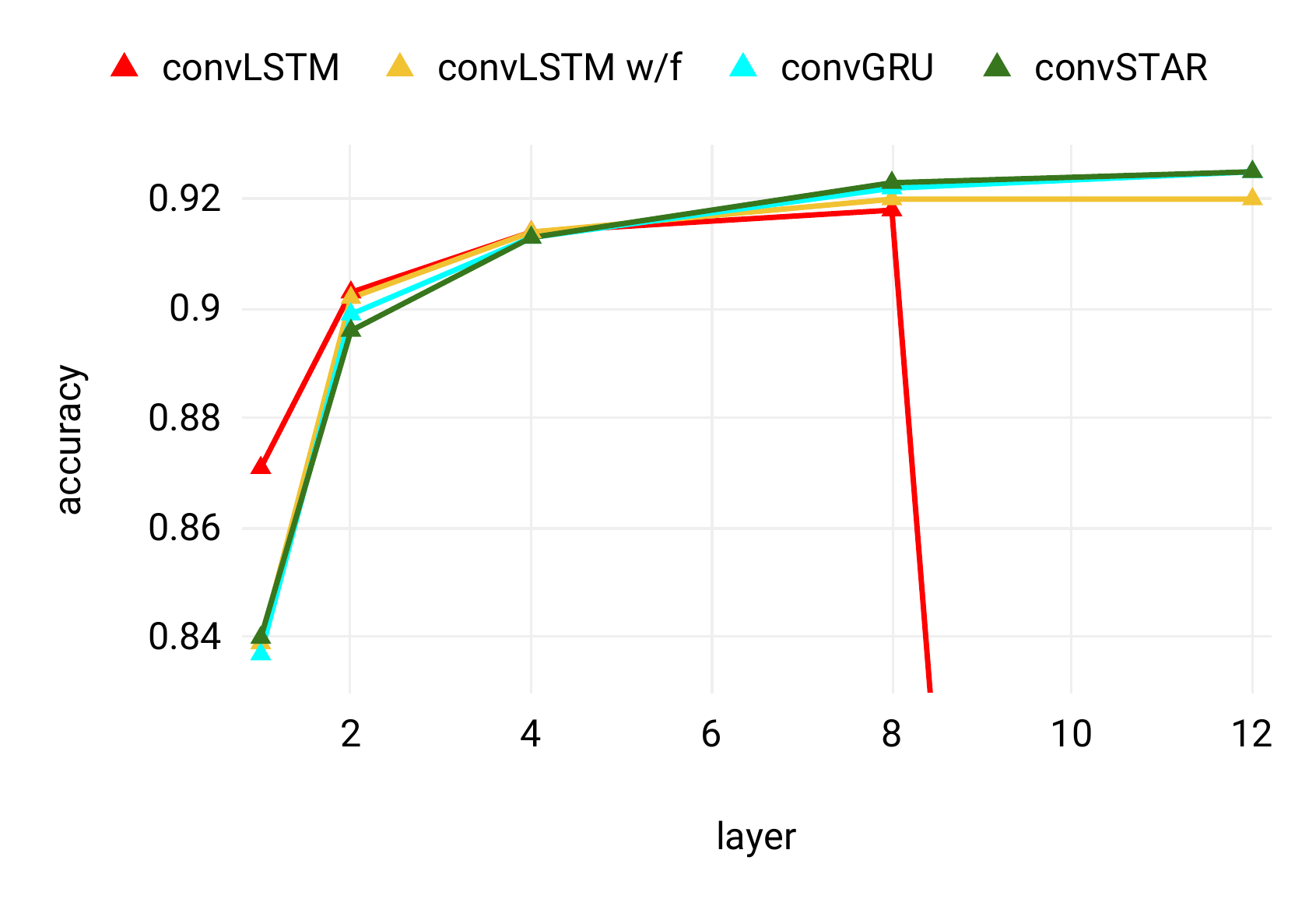}
        \caption{Jester, single label / sequence}
        \label{fig:jester}
    \end{subfigure}
    \caption{Time series classification. \emph{(a)} Crop classes. \emph{(b)} Hand gestures (convolutional RNNs).}\label{fig:tum}
\end{figure}

We evaluate model performance on a more realistic sequence modelling problem, where the aim is to classify agricultural crop types using sequences of satellite images. In this case, time-series modelling captures phenological evidence, i.e. different crops have different growing patterns over the season.
For the TUM dataset, the input is a time series of 26 multi-spectral Sentinel-2A satellite images with a ground resolution of 10$\,$m collected over a 102 km x 42 km area north of Munich, Germany between December 2015 and August 2016~\cite{fieldrnn}.
%The data is atmospherically corrected using the standard Sen2cor software package. 
%
We use patches of 3$\times$3 pixels recorded in 6 spectral channels and flattened into 54$\times$1 vectors as input.
For the BreizhCrop dataset, the input is a time series of 45 multi-spectral Sentinel-2A satellite images with a ground resolution of 10$\,$m collected from 580k field parcels in the Region of Brittany, France of the season 2017.
The input is 4 spectral channels (R, G, B, NIR)~\cite{breizhcrops}.
In the first task, only TUM dataset is used. The vectors are sequentially presented to the RNN model, which outputs a prediction at every time step (note that for this task the correct answer can sometimes be "cloud", "snow", "cloud shadow" or "water", which are easier to recognise than many crops). STAR outperforms all baselines, and it is again more suitable for stacking into deep architectures (Fig.~\ref{fig:tum}). 
In the second task, both datasets are used. The goal is a single-step prediction i.e., the model predicts a crop type after the entire sequence is presented. STAR significantly outperforms all the baselines including TCN and the recently proposed method, IndRNN~\cite{indRNN} (Tab.~\ref{table:crop}). Note that IndRNN also aims to build deep multi-layer RNNs. The performance gain is stronger in the BreizhCrop datasets. This is probably because the sequence is longer and the depth of the network helps to capture more complex dependencies in the data.

%We also run an experiment for single step prediction (the model predicts a crop type after the entire sequence is presented.). STAR achieves $71.9\%$ test accuracy against $70.5\%$, $70.4\%$, $70.2\%$, $70.1\%$ and $68.8\%$ accuracies obtained by GRU, RHN, LSTM w/f, LSTM and vRNN, respectively.

%In the second task, the model makes only one crop prediction for the complete sequence, via an additional layer that averages across time.
%  
% For the single-output task also the STAR network fails at 14 layers. We have not yet been able to identify the reason for this, possibly it is due to cloud cover that completely blanks out the signal over extended time windows and degrades the propagation.

%\begin{figure}[t]
%\centering
%\includegraphics[width=10.0cm]{imgs/jester2.pdf}
%\caption{Results for hand-gesture recognition task. Convolutional RNNs are compared in this task.}
%\label{fig:jester}
%\end{figure}

% Crop task table----------------------------------------
\begin{table}[t]
\begin{center}
\begin{tabular}{lllll}
\multicolumn{1}{c}{} & \multicolumn{2}{c}{\bf TUM} &\multicolumn{2}{c}{\bf BreizhCrops}\\
\cmidrule(lr){2-5}
\multicolumn{1}{c}{\bf Method}  &\multicolumn{1}{c}{\bf Acc}  &\multicolumn{1}{c}{\bf \#params} &\multicolumn{1}{c}{\bf Acc} &\multicolumn{1}{c}{\bf \#params}\\
\midrule
\multicolumn{1}{c}{ \begin{tabular}[c]{@{}c@{}}vRNN (2 layers)\end{tabular}  }     &\multicolumn{1}{c}{84.4\%} &\multicolumn{1}{c}{45k} &\multicolumn{1}{c}{38.1\%} &\multicolumn{1}{c}{55k} \\
\multicolumn{1}{c}{ \begin{tabular}[c]{@{}c@{}}LSTM (2 layers)\end{tabular}}     &\multicolumn{1}{c}{85.9\%}  &\multicolumn{1}{c}{170k}  &\multicolumn{1}{c}{60.1\%} &\multicolumn{1}{c}{210k}\\
\multicolumn{1}{c}{ \begin{tabular}[c]{@{}c@{}}LSTM w/f (2 layers)\end{tabular}}     &\multicolumn{1}{c}{85.7\%}  &\multicolumn{1}{c}{90k}  &\multicolumn{1}{c}{58.3\%} &\multicolumn{1}{c}{105k}\\
\multicolumn{1}{c}{ \begin{tabular}[c]{@{}c@{}}GRU (2/4 layers)\end{tabular} }     &\multicolumn{1}{c}{85.7\%} &\multicolumn{1}{c}{130k}  &\multicolumn{1}{c}{66.4\%} &\multicolumn{1}{c}{350k}\\
\multicolumn{1}{c}{ \begin{tabular}[c]{@{}c@{}}RHN (2 layers)\end{tabular} }     &\multicolumn{1}{c}{85.6\%} &\multicolumn{1}{c}{100k}  &\multicolumn{1}{c}{-} &\multicolumn{1}{c}{-}\\
\multicolumn{1}{c}{ \begin{tabular}[c]{@{}c@{}}IndRNN (4 layers)\\ \end{tabular}}     &\multicolumn{1}{c}{85.5\%}  &\multicolumn{1}{c}{90k}  &\multicolumn{1}{c}{56.8\%} &\multicolumn{1}{c}{105k}\\
\multicolumn{1}{c}{ \begin{tabular}[c]{@{}c@{}}TCN (2 layers)\\ \end{tabular}}     &\multicolumn{1}{c}{84.9\%}  &\multicolumn{1}{c}{300k}  &\multicolumn{1}{c}{61.5\%} &\multicolumn{1}{c}{360k}
\\ \midrule
\multicolumn{1}{c}{ \begin{tabular}[c]{@{}c@{}}STAR (4 layers)\end{tabular}}     &\multicolumn{1}{c}{\textbf{87.7\%}}  &\multicolumn{1}{c}{130k} &\multicolumn{1}{c}{68.2\%} &\multicolumn{1}{c}{170k}\\
\multicolumn{1}{c}{ \begin{tabular}[c]{@{}c@{}}STAR (6 layers)\end{tabular}}     &\multicolumn{1}{c}{87.6\%}  &\multicolumn{1}{c}{210k} &\multicolumn{1}{c}{\textbf{69.6\%}} &\multicolumn{1}{c}{270k}
\\ \hline 

\end{tabular}
\end{center}
\caption{Performance comparison for time series crop classification. }
\label{table:crop}
\end{table}
% Crop task table----------------------------------------

\subsection{Music Modeling}
JSB Chorales~\cite{jsb_dataset} is a polyphonic music dataset consisting of the entire corpus of 382 four-part harmonized chorales by J.~S.~Bach. Each input is a sequence of chord elements. Each element is an 88-bit binary code that corresponds to the 88 keys of a piano, with 1 indicating a key pressed at a given time.  Piano-Midi~\cite{piano_dataset} is a classical piano MIDI archive that consists of 130 pieces by various composers. These datasets have been used in several previous works to investigate the ability of RNNs to represent music~\cite{gruX,diagRNN}. The performance on both tasks is measured in terms of per-frame negative log-likelihood (NLL) on a test set. We follow the exact same experimental setup described in~\cite{diagRNN}. STAR works better than all tested RNN baselines, and performs on par with TCN (see Tab.~\ref{table:music}).

\begin{table}[t]
\begin{center}
\begin{tabular}{lllll}
\multicolumn{1}{c}{} & \multicolumn{2}{c}{\bf JSB Chorales} &\multicolumn{2}{c}{\bf Piano-Midi}\\
\cmidrule(lr){2-5}
\multicolumn{1}{c}{\bf Method}  &\multicolumn{1}{c}{\bf NLL}  &\multicolumn{1}{c}{\bf \#params} &\multicolumn{1}{c}{\bf NLL} &\multicolumn{1}{c}{\bf \#params}\\
\midrule
\multicolumn{1}{c}{ \begin{tabular}[c]{@{}c@{}}vRNN \cite{diagRNN}\end{tabular}  }     &\multicolumn{1}{c}{8.72} &\multicolumn{1}{c}{40k} &\multicolumn{1}{c}{7.65} &\multicolumn{1}{c}{140k} \\
\multicolumn{1}{c}{ \begin{tabular}[c]{@{}c@{}}LSTM \cite{diagRNN}\end{tabular}}     &\multicolumn{1}{c}{8.51}  &\multicolumn{1}{c}{650k}  &\multicolumn{1}{c}{7.84} &\multicolumn{1}{c}{480k}\\
\multicolumn{1}{c}{ \begin{tabular}[c]{@{}c@{}}GRU \cite{diagRNN}\end{tabular} }     &\multicolumn{1}{c}{8.53} &\multicolumn{1}{c}{640k}  &\multicolumn{1}{c}{7.62} &\multicolumn{1}{c}{690k}\\
\multicolumn{1}{c}{ \begin{tabular}[c]{@{}c@{}}diagRNN \cite{diagRNN}\\ \end{tabular}}     &\multicolumn{1}{c}{8.14}  &\multicolumn{1}{c}{420k}  &\multicolumn{1}{c}{7.48} &\multicolumn{1}{c}{360k}\\
\multicolumn{1}{c}{ \begin{tabular}[c]{@{}c@{}}TCN (2 layers) \cite{tcn}\\ \end{tabular}}     &\multicolumn{1}{c}{8.10}  &\multicolumn{1}{c}{300k}  &\multicolumn{1}{c}{-} &\multicolumn{1}{c}{-}
\\ \midrule
\multicolumn{1}{c}{ \begin{tabular}[c]{@{}c@{}}STAR (2 layers)\end{tabular}}     &\multicolumn{1}{c}{8.13}  &\multicolumn{1}{c}{360k} &\multicolumn{1}{c}{\textbf{7.40}} &\multicolumn{1}{c}{480k}\\
\multicolumn{1}{c}{ \begin{tabular}[c]{@{}c@{}}STAR (4 layers)\end{tabular}}     &\multicolumn{1}{c}{\textbf{8.09}}  &\multicolumn{1}{c}{830k} &\multicolumn{1}{c}{-} &\multicolumn{1}{c}{-}
\\ \hline 

\end{tabular}
\end{center}
\caption{Performance comparison for music task. The performance is measured in terms of negative log-likelihood (NLL).}
\label{table:music}
\end{table}

\subsection{Character-level Language Modeling}
For this task we used the PennTreebank (PTB)~\cite{ptb}. When used as a character-level language corpus, PTB contains 5,059K characters for training, 396K for validation, and 446K for testing, with an alphabet size of 50. The goal is to predict the next character given the preceding context. We follow the exact same experimental setup as~\cite{tcn}. The performance is measured in terms of bits per character (BPC, i.e. average cross entropy over the alphabet) on the test set. On this task STAR outperforms all baselines, including Transformer and TCN (see Tab.~\ref{table:PTB}).

\begin{table}[t]
\begin{center}
\begin{tabular}{lll}
\multicolumn{1}{c}{\bf Method}  &\multicolumn{1}{c}{\bf BPC}  &\multicolumn{1}{c}{\bf \#params} \\
\midrule

\multicolumn{1}{c}{ \begin{tabular}[c]{@{}c@{}}vRNN \cite{tcn}\end{tabular}  }     &\multicolumn{1}{c}{1.48} &\multicolumn{1}{c}{3M}  \\
\multicolumn{1}{c}{ \begin{tabular}[c]{@{}c@{}} LSTM (2 layers) \cite{tcn} \end{tabular}}     &\multicolumn{1}{c}{1.36}  &\multicolumn{1}{c}{3M} \\
\multicolumn{1}{c}{ \begin{tabular}[c]{@{}c@{}}GRU \cite{tcn} \end{tabular}}     &\multicolumn{1}{c}{1.37}  &\multicolumn{1}{c}{3M}\\
\multicolumn{1}{c}{ \begin{tabular}[c]{@{}c@{}} IndRNN (6 layers)*\end{tabular}}     &\multicolumn{1}{c}{1.42}  &\multicolumn{1}{c}{3M}\\
\multicolumn{1}{c}{ \begin{tabular}[c]{@{}c@{}}TCN (3 layers) \cite{tcn}\end{tabular}}     &\multicolumn{1}{c}{1.31}  &\multicolumn{1}{c}{3M}\\
\multicolumn{1}{c}{ \begin{tabular}[c]{@{}c@{}} Transformer (3 layers) \cite{r_transformer}\end{tabular}}     &\multicolumn{1}{c}{1.45}  &\multicolumn{1}{c}{-}\\
\midrule
\multicolumn{1}{c}{ \begin{tabular}[c]{@{}c@{}} STAR (6 layers)\end{tabular}}     &\multicolumn{1}{c}{\textbf{1.30}}  &\multicolumn{1}{c}{3M} 

\\ \hline 

\end{tabular}
\end{center}
\caption{Performance comparison for PennTreebank character-level language modeling. The performance is measured in terms of bits per character (BPC). *We run this experiment as designed in \cite{tcn}'s experimental setup with a limited number of parameters to allow for a fair comparison. Note that \cite{indRNN} reports a better result, but uses many more model parameters.} 
\label{table:PTB}
\end{table}

\subsection{Hand-gesture recognition from video} 
\begin{figure*}[t]
    \centering
        \includegraphics[width=0.9\textwidth]{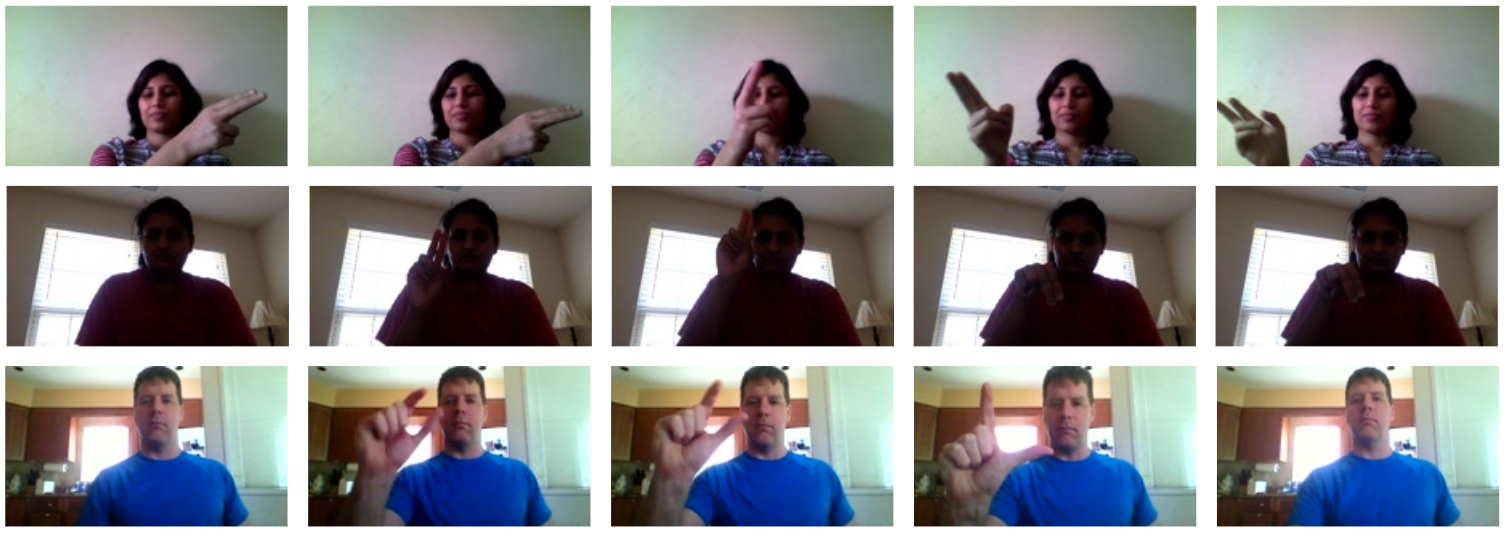}
    \caption{Example frames of the Jester dataset. Columns show $1^{st}$, $8^{th}$, $16^{th}$, $24^{th}$, $32^{nd}$ frames, respectively. First row: Sliding two fingers right. Second row: Sliding two fingers down. Third row: Zooming in with two fingers.}
    \label{fig:jester_samples}
\end{figure*}

We also evaluate STAR on sequences of images, using \emph{convolutional} layers. We analyse performance of STAR versus state-of-the-art on gesture recognition from video and pixel-wise crop classification.
The 20BN-Jester dataset V1 \cite{jester} is a large collection of densely-labelled short video clips, where each clip contains a predefined hand gesture performed by a worker in front of a laptop camera or webcam. In total, the dataset includes 148'094 RGB video files of 27 types of gestures (see Fig.~\ref{fig:jester_samples}).
The task is to classify which gesture is seen in a video. 32 consecutive frames of size 112$\times$112 pixels are sequentially presented to the convolutional RNN. At the end, the model again predicts a gesture class via an averaging layer over all time steps.
The outcome for convolutional RNNs is coherent with the previous results, see Fig.~\ref{fig:jester}, Tab.~\ref{table:jester}. Going deeper improves the performance of all four tested convRNNs. The improvement is strongest for convolutional STAR, and the best performance is reached with a deep model (12 layers). %, where training the baselines mostly fails.
In summary, the results confirm both our intuitions that depth is particularly useful for convolutional RNNs, and that STAR is more suitable for deeper architectures, where it achieves higher performance with better memory efficiency.
We note that in the shallow 1-2 layer setting the conventional LSTM performs a slightly better than the three others, likely due to its larger capacity. Lastly, we conduct the same additional experiment with the hybrid architecture as we do for MNIST tasks. We stack seven layers of STAR on top of one layer of LSTM. This further improves the results and achieves $92.7\%$ accuracy (compared this to eight LSTM layers, which achieve only $91.8\%$ accuracy with about twice as many parameters).

% Jester---------------------------------------------------------------
\begin{table}[t]
\begin{center}
\begin{tabular}{lll}
\multicolumn{1}{c}{\bf Method}  &\multicolumn{1}{c}{\bf Accuracy}  &\multicolumn{1}{c}{\bf \#params} \\
\midrule

\multicolumn{1}{c}{ \begin{tabular}[c]{@{}c@{}}convLSTM (8 layers) \end{tabular}  }     &\multicolumn{1}{c}{91.8 \%} &\multicolumn{1}{c}{2.2M}  \\
\multicolumn{1}{c}{ \begin{tabular}[c]{@{}c@{}} convLSTM w/f (8 layers) \end{tabular}}     &\multicolumn{1}{c}{92.0 \%}  &\multicolumn{1}{c}{1.1M} \\
\multicolumn{1}{c}{ \begin{tabular}[c]{@{}c@{}}convGRU (12 layers) \end{tabular}}     &\multicolumn{1}{c}{92.5 \%}  &\multicolumn{1}{c}{2.5M}\\
\midrule
\multicolumn{1}{c}{ \begin{tabular}[c]{@{}c@{}} convSTAR (8 layers)\end{tabular}}     &\multicolumn{1}{c}{92.3 \%}  &\multicolumn{1}{c}{0.8M}\\ 
\multicolumn{1}{c}{ \begin{tabular}[c]{@{}c@{}} convSTAR (12 layers)\end{tabular}}     &\multicolumn{1}{c}{92.5 \%}  &\multicolumn{1}{c}{1.2M}\\ 
\multicolumn{1}{c}{ \begin{tabular}[c]{@{}c@{}} convLSTM convSTAR (8 layers)\end{tabular}}     &\multicolumn{1}{c}{\textbf{92.7 \%}}  &\multicolumn{1}{c}{0.9M}
\\ \hline 

\end{tabular}
\end{center}
\caption{Performance comparison for the gesture recognition task (Jester).}
\label{table:jester}
\end{table}
% Jester---------------------------------------------------------------

%\subsection{Relation to prior work}
%Discuss the Gated Feedback RNN results. \cite{gatedFeedback}
%We would like to point out experimental results in \citep{gatedFeedback}. \citeauthor{gatedFeedback} propose a gated feedback RNN (GFRNN) that extends the stacked RNN architecture with introducing extra connections between consecutive layers. Their method improves the stacked RNN performance in language modelling task and they suggest that the improvement arises because the GFRNN can adaptively assign different layers to different timescales. However, while GFRNN improves the 3-layers LSTM and 3-layers GRU performance, it worsens the vanilla RNN performance. So the improvement might be because simply extra connections between layers strengthen the gradient propagation in LSTM and GRU cases but over-strengthen it in the vanilla RNN case and cause unstable gradient propagation as we observed in this paper.

\subsection{TUM image series pixel-wise classification}
\begin{figure*}[t]
    \centering
        \includegraphics[width=0.9\textwidth]{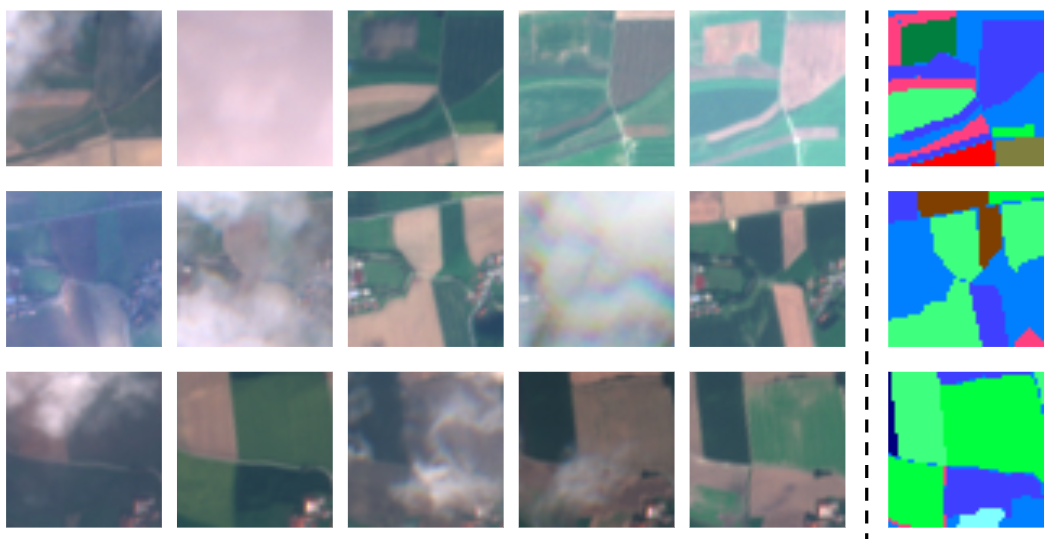}
    \caption{Example satellite images of the TUM dataset. Each row shows randomly sampled images (in order, only R, G and B channels) from a satellite image time-series. The last column shows the ground-truth where different colors correspond to different crop types.}
    \label{fig:tum_samples}
\end{figure*}

%\paragraph{TUM image series pixel-wise classification} 
In another experiment with convolutional RNNs, we classify crops pixel-wise (and thus use convolutional layers) using a dataset~\cite{tum_image} (TUM) containing Sentinel-2A optical satellite image sequences (RGB and NIR at $10~m$ ground sampling distance) accompanied by ground-truth land cover maps. Each satellite image sequence contains 30 images of size $48\times48~px$  collected in 2016 within a $102~km\times42~km$ region north of Munich, Germany (see Fig.~\ref{fig:tum_samples}).
We compare pixel-wise classification accuracy for a network with a fixed depth of four layers and for four different basic recurrent cells LSTM, LSTM with only a forget gate, GRU, and the proposed STAR cell (Tab.~\ref{table:SOTA_RS}). Moreover we include the performance obtained in~\cite{tum_image} using a bidirectional convolutional GRU with a single layer. Our STAR cell outperforms all other methods (Tab.~\ref{table:SOTA_RS}) while requiring less memory and being computationally less costly.

%MOVE TO SUPPLEMENTARY
%\begin{figure}%[t]
%    \centering
%        \includegraphics[width=\columnwidth]{new_images/learning_curves_grd.pdf}
%    \caption{TUM image sequences - learning curves - \textcolor{red}{should go to appendix I guess}}
%    \label{fig:learning}
%\end{figure}

\begin{table}[t]
\begin{center}
\begin{tabular}{llll}
\multicolumn{1}{c}{\bf Method}  &\multicolumn{1}{c}{\bf Acc}  &\multicolumn{1}{c}{\bf \#params} &\multicolumn{1}{c}{\bf \#compute}\\
\midrule

\multicolumn{1}{c}{ \begin{tabular}[c]{@{}c@{}}bi-convGRU\\(1 layer) \cite{tum_image}\end{tabular}  }     &\multicolumn{1}{c}{89.7\%} &\multicolumn{1}{c}{6.2M} &\multicolumn{1}{c}{46bn} \\
%\midrule
\multicolumn{1}{c}{ \begin{tabular}[c]{@{}c@{}}convLSTM\\ (4 layers) \end{tabular}}     &\multicolumn{1}{c}{90.6\%}  &\multicolumn{1}{c}{292k}  &\multicolumn{1}{c}{2.7bn} \\

\multicolumn{1}{c}{ \begin{tabular}[c]{@{}c@{}}convLSTM w/f\\ (4 layers)\end{tabular} }     &\multicolumn{1}{c}{89.6\%} &\multicolumn{1}{c}{161k}  &\multicolumn{1}{c}{1.5bn} \\

\multicolumn{1}{c}{ \begin{tabular}[c]{@{}c@{}}convGRU\\ (4 layers)\end{tabular}}     &\multicolumn{1}{c}{90.1\%}  &\multicolumn{1}{c}{227k}  &\multicolumn{1}{c}{2.1bn} 

\\ \midrule
\multicolumn{1}{c}{ \begin{tabular}[c]{@{}c@{}}convSTAR\\ (4 layers)\end{tabular}}     &\multicolumn{1}{c}{\textbf{91.9\%}}  &\multicolumn{1}{c}{124k} &\multicolumn{1}{c}{1.1bn} 

\\ \hline 

\end{tabular}
\end{center}
\caption{Performance comparison for TUM pixel-wise image classification task.}
\label{table:SOTA_RS}
\end{table}

\subsection{Computational Resources and Training Time}

\begin{figure}%[t]
    \centering
        \includegraphics[width=\columnwidth]{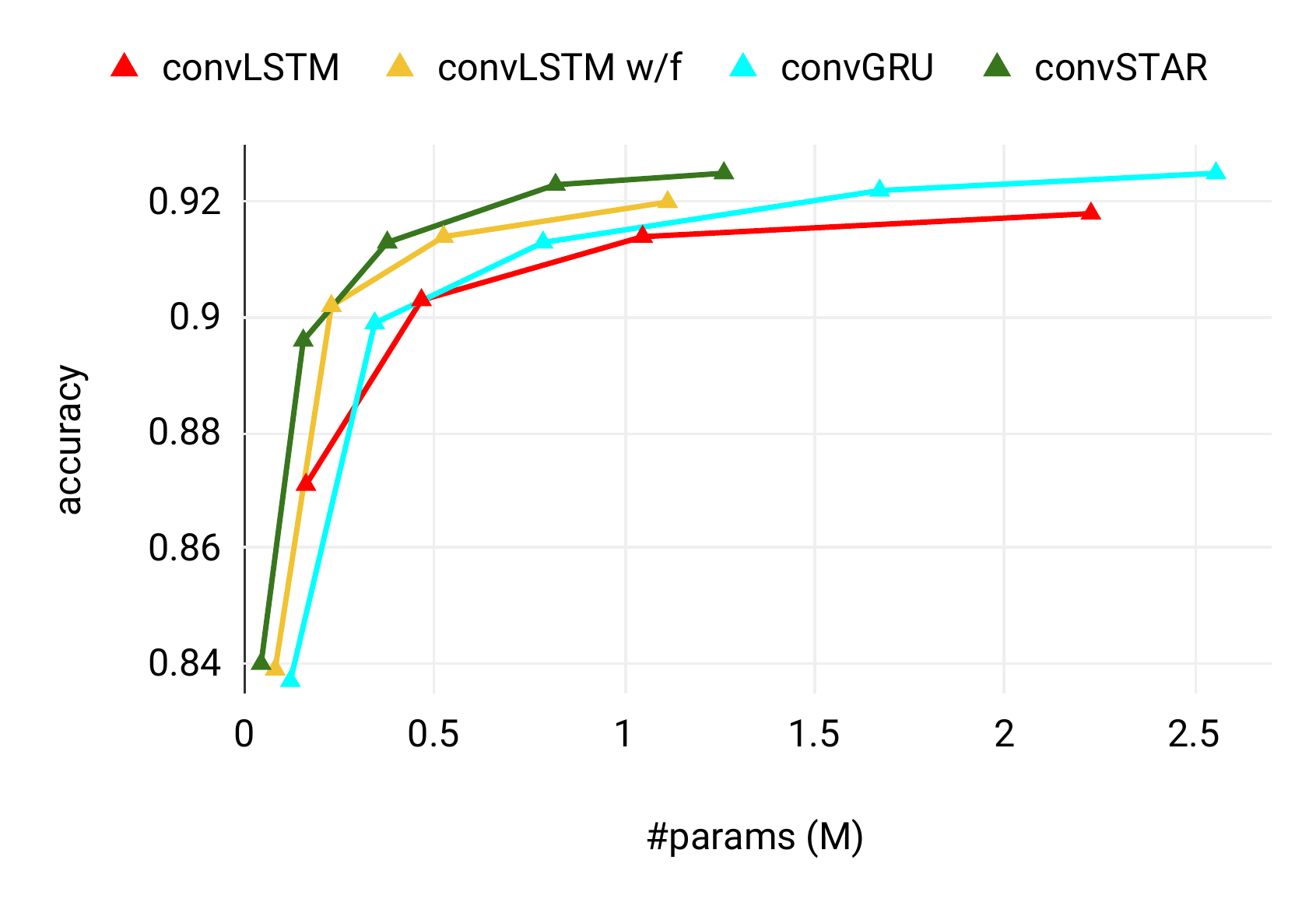}
    \caption{Accuracy versus number of model parameters for the gesture recognition task (Jester).}
    \label{fig:params}
\end{figure}

\begin{figure}%[t]
    \centering
        \includegraphics[width=\columnwidth]{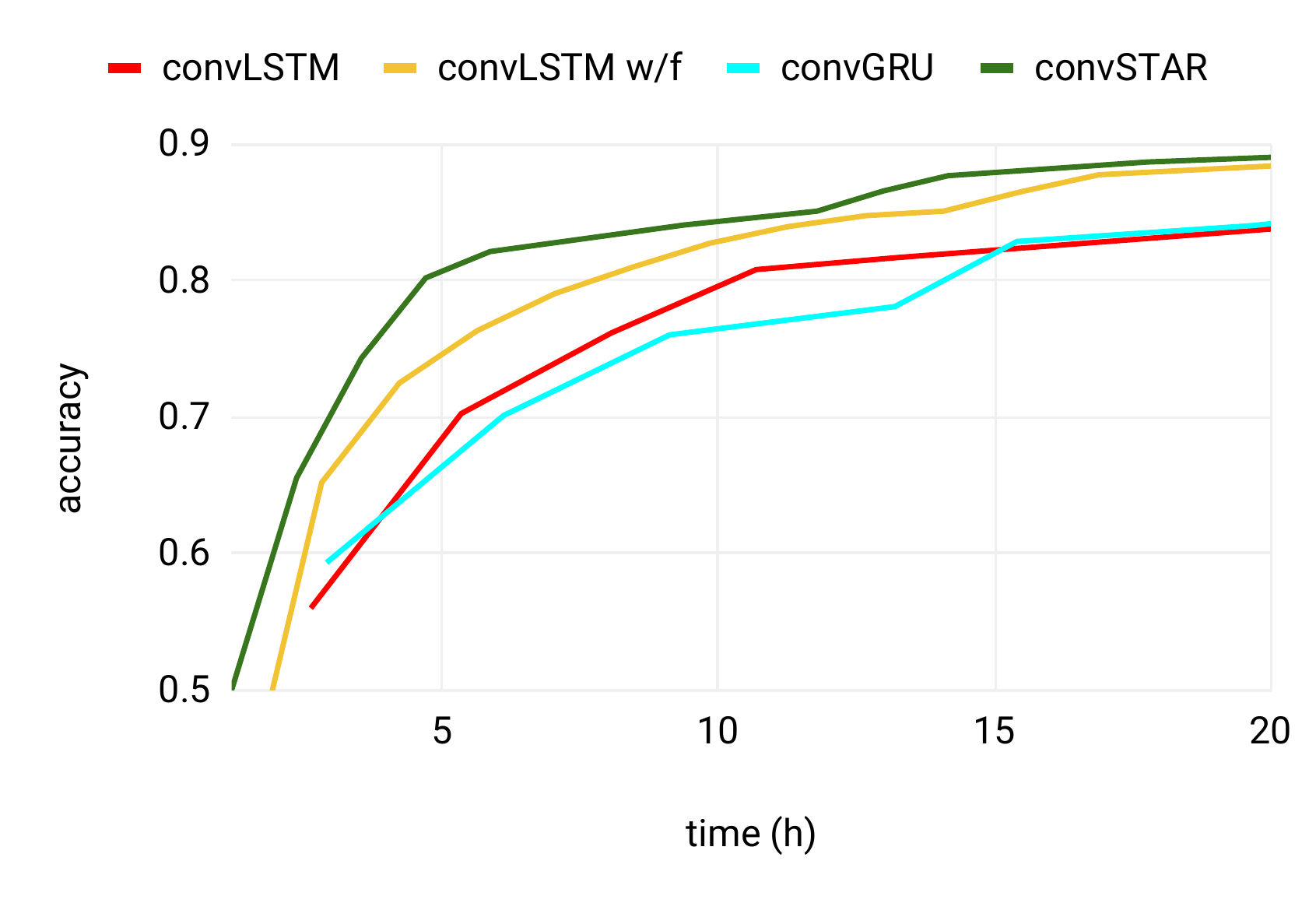}
    \caption{Test accuracy versus training time for the gesture recognition task (Jester), 4 layers networks.}
    \label{fig:one_day}
\end{figure}

%When designing the STAR cell we target a low number of parameters in the attempt of having a more lightweight recurrent cell. 
Last, we compare to widely used recurrent units LSTM and GRU in terms of parameter efficiency and training time for the convolutional version used in gesture recognition. We plot performance versus number of parameters (Fig.~\ref{fig:params}) %is similar to the one in Fig.~\ref{fig:tum}(b) but in this case the x-axis denotes the number of parameters and not number of layers. 
STAR outperforms LSTM and performs on par with GRU, but requires only half the number of parameters. % We can clearly see in this case the advantage of the propose STAR cell. The best accuracy achieved by the STAR cell and the GRU are the same but the STAR cell requires almost half of the parameters, making it by far the most computationally efficient.
We plot accuracy on the validation dataset versus training time for different recurrent units for the gesture recognition task in Fig.~\ref{fig:one_day}. STAR does not only require significantly less parameters but can also be trained much faster: the validation accuracy on the dataset after 8 hours is comparable to the best validation achieved by the LSTM and the GRU after 20 hours of training.

%MOVE TO SUPPLEMENTARY: %In Fig.~\ref{fig:learning} we show the learning curves for different recurrent units for TUM pixel-wise image classification. As can be seen the STAR cell is not only more efficient in terms of parameters but is also able to be trained much faster: the loss on the validation dataset after 5 epochs is comparable to the best validation achieved by the other methods at the end of the training.

\section{Conclusion}

We have proposed STAR, a novel stackable recurrent cell type that it is specifically designed to be employed in deep recurrent architectures. 
%
%We first conduct an analysis on the gradient propagation in deep recurrent architectures. 
A theoretical analysis and associated numerical simulations indicated that widely used standard RNN cells like LSTM and GRU do not preserve gradient magnitudes in the "vertical" direction during backpropagation. As the depth of the network grows, the risk of either exploding or vanishing gradients increases. We leveraged this analysis to design a novel cell that better preserves the gradient magnitude between two adjacent layers, is better suited for deep architectures, and requires fewer parameters than other widely used recurrent units.
%In a second step, we have proposed a new RNN cell, termed the STAckable Recurrent unit, which better preserves gradients through deep architectures and facilitates their training. 
An extensive experimental evaluation on several publicly available datasets confirms that STAR units can be stacked into deeper architectures and in many cases performs better than state-of-the-art architectures. 

We see two main directions for future work. On the one hand, it would be worthwhile to develop a more formal and thorough mathematical analysis of the gradient flow, and perhaps even derive rigorous bounds for specific cell types, that could, in turn, inform the network design.
On the other hand, it appears promising to investigate whether the analysis of the gradient flows could serve as a basis for better initialisation schemes to compensate the systematic influences of the cells structure, e.g., gating functions, in the training of deep RNNs.

\section*{Acknowledgments}
We thank the Swiss Federal Office for Agriculture (FOAG) for partially funding this Research project through the DeepField Project.
%\vfill

{\small
\bibliographystyle{ieee_fullname}
\bibliography{main}

\begin{thebibliography}{10}\itemsep=-1pt

\bibitem{jester}
The 20bn-jester dataset v1.
\newblock \url{https://20bn.com/datasets/jester}.

\bibitem{jsb_dataset}
Moray Allan and Christopher Williams.
\newblock Harmonising chorales by probabilistic inference.
\newblock In {\em Advances in neural information processing systems}, 2005.

\bibitem{uRNN}
Martin Arjovsky, Amar Shah, and Yoshua Bengio.
\newblock Unitary evolution recurrent neural networks.
\newblock In {\em ICML}, 2016.

\bibitem{tcn}
Shaojie Bai, J~Zico Kolter, and Vladlen Koltun.
\newblock An empirical evaluation of generic convolutional and recurrent
  networks for sequence modeling.
\newblock {\em arXiv preprint arXiv:1803.01271}, 2018.

\bibitem{bengio2009learning}
Yoshua Bengio.
\newblock Learning deep architectures for {AI}.
\newblock {\em Foundations and trends{\textregistered} in Machine Learning},
  2(1):1--127, 2009.

\bibitem{vanishing_bengio}
Yoshua Bengio, Patrice Simard, Paolo Frasconi, et~al.
\newblock Learning long-term dependencies with gradient descent is difficult.
\newblock {\em IEEE TNN}, 5(2):157--166, 1994.

\bibitem{skipRNN}
V{\'\i}ctor Campos, Brendan Jou, Xavier Gir{\'o}-i Nieto, Jordi Torres, and
  Shih-Fu Chang.
\newblock Skip {RNN}: Learning to skip state updates in recurrent neural
  networks.
\newblock In {\em ICLR}, 2018.

\bibitem{antisymRNN}
Bo Chang, Minmin Chen, Eldad Haber, and Ed~H Chi.
\newblock {AntisymmetricRNN}: A dynamical system view on recurrent neural
  networks.
\newblock In {\em ICLR}, 2019.

\bibitem{mean_field_1}
Minmin Chen, Jeffrey Pennington, and Samuel~S Schoenholz.
\newblock Dynamical isometry and a mean field theory of rnns: Gating enables
  signal propagation in recurrent neural networks.
\newblock In {\em ICML}, 2018.

\bibitem{gruX}
Junyoung Chung, Caglar Gulcehre, KyungHyun Cho, and Yoshua Bengio.
\newblock Empirical evaluation of gated recurrent neural networks on sequence
  modeling.
\newblock In {\em NIPS Workshop}, 2014.

\bibitem{gatedFeedback}
Junyoung Chung, Caglar Gulcehre, Kyunghyun Cho, and Yoshua Bengio.
\newblock Gated feedback recurrent neural networks.
\newblock In {\em ICML}, 2015.

\bibitem{bn_lstm}
T. Cooijmans, N. Ballas, C. Laurent, C. G{\"u}lcehre, and A. Courville.
\newblock Recurrent batch normalization.
\newblock In {\em ICLR}, 2017.

\bibitem{graves2013speech}
Alex Graves, Abdel-rahman Mohamed, and Geoffrey Hinton.
\newblock Speech recognition with deep recurrent neural networks.
\newblock In {\em ICASSP}, 2013.

\bibitem{orthagonal_init}
Mikael Henaff, Arthur Szlam, and Yann LeCun.
\newblock Recurrent orthogonal networks and long-memory tasks.
\newblock In {\em ICML}, 2016.

\bibitem{vanishing_munich}
Sepp Hochreiter.
\newblock Untersuchungen zu {Dynamischen} {Neuronalen} {Netzen}.
\newblock {\em Diploma Thesis, Technische Universit{\"a}t M{\"u}nchen}, 91(1),
  1991.

\bibitem{lstm}
Sepp Hochreiter and J{\"u}rgen Schmidhuber.
\newblock Long short-term memory.
\newblock {\em Neural Computation}, 9(8):1735--1780, 1997.

\bibitem{batch_norm}
Sergey Ioffe and Christian Szegedy.
\newblock Batch normalization: Accelerating deep network training by reducing
  internal covariate shift.
\newblock In {\em ICML}, 2015.

\bibitem{jacot2018neural}
Arthur Jacot, Franck Gabriel, and Cl{\'e}ment Hongler.
\newblock Neural tangent kernel: Convergence and generalization in neural
  networks.
\newblock In {\em Advances in neural information processing systems}, 2018.

\bibitem{residual_LSTM}
Jaeyoung Kim, Mostafa El-Khamy, and Jungwon Lee.
\newblock Residual {LSTM}: Design of a deep recurrent architecture for distant
  speech recognition.
\newblock In {\em Interspeech}, 2017.

\bibitem{adam}
Diederik~P Kingma and Jimmy Ba.
\newblock Adam: A method for stochastic optimization.
\newblock In {\em ICLR}, 2014.

\bibitem{alexnet}
Alex Krizhevsky, Ilya Sutskever, and Geoffrey~E Hinton.
\newblock Imagenet classification with deep convolutional neural networks.
\newblock In {\em Advances in neural information processing systems}, 2012.

\bibitem{identitiy_init}
Quoc~V Le, Navdeep Jaitly, and Geoffrey~E Hinton.
\newblock A simple way to initialize recurrent networks of rectified linear
  units.
\newblock {\em arXiv preprint arXiv:1504.00941}, 2015.

\bibitem{mnist}
Yann LeCun, L{\'e}on Bottou, Yoshua Bengio, Patrick Haffner, et~al.
\newblock Gradient-based learning applied to document recognition.
\newblock {\em Proc.\ IEEE}, 86(11):2278--2324, 1998.

\bibitem{lee2019wide}
Jaehoon Lee, Lechao Xiao, Samuel Schoenholz, Yasaman Bahri, Roman Novak, Jascha
  Sohl-Dickstein, and Jeffrey Pennington.
\newblock Wide neural networks of any depth evolve as linear models under
  gradient descent.
\newblock In {\em Advances in neural information processing systems}, 2019.

\bibitem{indRNN}
Shuai Li, Wanqing Li, Chris Cook, Ce Zhu, and Yanbo Gao.
\newblock Independently recurrent neural network (indrnn): Building a longer
  and deeper rnn.
\newblock In {\em CVPR}, 2018.

\bibitem{videolstm}
Zhenyang Li, Kirill Gavrilyuk, Efstratios Gavves, Mihir Jain, and Cees~GM
  Snoek.
\newblock Video{LSTM} convolves, attends and flows for action recognition.
\newblock {\em CVIU}, 166:41--50, 2018.

\bibitem{ptb}
Mitchell Marcus, Beatrice Santorini, and Mary~Ann Marcinkiewicz.
\newblock Building a large annotated corpus of english: The penn treebank.
\newblock 1993.

\bibitem{mhammedi2017efficient}
Zakaria Mhammedi, Andrew Hellicar, Ashfaqur Rahman, and James Bailey.
\newblock Efficient orthogonal parametrisation of recurrent neural networks
  using householder reflections.
\newblock In {\em ICML}, 2017.

\bibitem{piano_dataset}
Graham~E Poliner and Daniel~PW Ellis.
\newblock A discriminative model for polyphonic piano transcription.
\newblock {\em EURASIP Journal on Advances in Signal Processing}, 2007:1--9,
  2006.

\bibitem{resRNN}
Sabeek Pradhan and Shayne Longpre.
\newblock Exploring the depths of recurrent neural networks with stochastic
  residual learning, 2016.

\bibitem{fieldrnn}
Marc Ru{\ss}wurm and Marco K{\"o}rner.
\newblock Temporal vegetation modelling using long short-term memory networks
  for crop identification from medium-resolution multi-spectral satellite
  images.
\newblock In {\em CVPR Workshops}, 2017.

\bibitem{tum_image}
Marc Ru{\ss}wurm and Marco K{\"o}rner.
\newblock Multi-temporal land cover classification with sequential recurrent
  encoders.
\newblock {\em ISPRS International Journal of Geo-Information}, 7(4):129, 2018.

\bibitem{breizhcrops}
Marc Ru{\ss}wurm, S{\'e}bastien Lef{\`e}vre, and Marco K{\"o}rner.
\newblock Breizhcrops: A satellite time series dataset for crop type
  identification.
\newblock In {\em ICML Workshop}, 2019.

\bibitem{ntu}
Amir Shahroudy, Jun Liu, Tian-Tsong Ng, and Gang Wang.
\newblock Ntu rgb+ d: A large scale dataset for 3d human activity analysis.
\newblock In {\em CVPR}, 2016.

\bibitem{vgg}
K. Simonyan and A. Zisserman.
\newblock Very deep convolutional networks for large-scale image recognition.
\newblock In {\em ICLR}, 2015.

\bibitem{highway}
Rupesh~Kumar Srivastava, Klaus Greff, and J{\"u}rgen Schmidhuber.
\newblock Highway networks.
\newblock In {\em ICML Workshop}, 2015.

\bibitem{diagRNN}
Y~Cem Subakan and Paris Smaragdis.
\newblock Diagonal rnns in symbolic music modeling.
\newblock In {\em 2017 IEEE Workshop on Applications of Signal Processing to
  Audio and Acoustics (WASPAA)}. IEEE, 2017.

\bibitem{sutskever2014sequence}
Ilya Sutskever, Oriol Vinyals, and Quoc~V Le.
\newblock Sequence to sequence learning with neural networks.
\newblock In {\em Advances in neural information processing systems}, 2014.

\bibitem{chrono_init}
Corentin Tallec and Yann Ollivier.
\newblock Can recurrent neural networks warp time?
\newblock In {\em ICLR}, 2018.

\bibitem{UnEffForGate}
Jos Van Der~Westhuizen and Joan Lasenby.
\newblock The unreasonable effectiveness of the forget gate.
\newblock {\em arXiv preprint arXiv:1804.04849}, 2018.

\bibitem{transformer}
Ashish Vaswani, Noam Shazeer, Niki Parmar, Jakob Uszkoreit, Llion Jones,
  Aidan~N Gomez, {\L}ukasz Kaiser, and Illia Polosukhin.
\newblock Attention is all you need.
\newblock In {\em Advances in neural information processing systems}, 2017.

\bibitem{vinyals2015neural}
Oriol Vinyals and Quoc Le.
\newblock A neural conversational model.
\newblock {\em arXiv preprint arXiv:1506.05869}, 2015.

\bibitem{softOrth}
Eugene Vorontsov, Chiheb Trabelsi, Samuel Kadoury, and Chris Pal.
\newblock On orthogonality and learning recurrent networks with long term
  dependencies.
\newblock In {\em ICML}, 2017.

\bibitem{r_transformer}
Zhiwei Wang, Yao Ma, Zitao Liu, and Jiliang Tang.
\newblock R-transformer: Recurrent neural network enhanced transformer.
\newblock {\em arXiv preprint arXiv:1907.05572}, 2019.

\bibitem{FCuRNN}
Scott Wisdom, Thomas Powers, John Hershey, Jonathan Le~Roux, and Les Atlas.
\newblock Full-capacity unitary recurrent neural networks.
\newblock In {\em Advances in neural information processing systems}, 2016.

\bibitem{convLSTM}
Shi Xingjian, Zhourong Chen, Hao Wang, Dit-Yan Yeung, Wai-Kin Wong, and
  Wang-chun Woo.
\newblock Convolutional {LSTM} network: A machine learning approach for
  precipitation nowcasting.
\newblock In {\em Advances in neural information processing systems}, 2015.

\bibitem{stanh_rnn}
et~al. Zhang, Saizheng.
\newblock Architectural complexity measures of recurrent neural networks.
\newblock In {\em Advances in neural information processing systems}, 2016.

\bibitem{gesture_convLSTM}
Liang Zhang, Guangming Zhu, Lin Mei, Peiyi Shen, Syed Afaq~Ali Shah, and
  Mohammed Bennamoun.
\newblock Attention in convolutional {LSTM} for gesture recognition.
\newblock In {\em Advances in neural information processing systems}, 2018.

\bibitem{rhn}
Julian~Georg Zilly, Rupesh~Kumar Srivastava, Jan Koutn{\'\i}k, and J{\"u}rgen
  Schmidhuber.
\newblock Recurrent highway networks.
\newblock In {\em ICML}, 2017.

\end{thebibliography}
}

\begin{IEEEbiography}[{\includegraphics[width=1in,height=1.25in,clip,keepaspectratio]{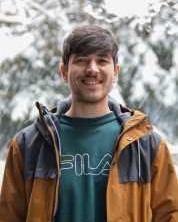}}]{Mehmet Ozgur Turkoglu}
received his BSc degrees in both electrical engineering and physics from Bogazici University in 2016. He studied a master's in electrical engineering with a specialization in computer vision at the University of Twente. He is a PhD candidate in the EcoVision group at ETH Z\"urich since 2018. His research interests include computer vision, deep learning and their applications to remote sensing data. He is particularly interested in deep sequence modeling of time-series data.
\end{IEEEbiography}

\begin{IEEEbiography}[{\includegraphics[width=1in,height=1.25in,clip,keepaspectratio]{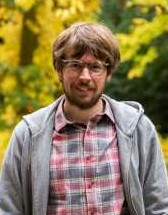}}]{Stefano D'Aronco}
received his BS and MS degrees in electronic engineering from the Università degli studi di Udine, in 2010 and 2013 respectively. He then joined the Signal Processing Laboratory (LTS4) in 2014 as a PhD student under the supervision of Prof. Pascal Frossard. He received his PhD in Electrical Engineering from École Polytechnique Fédérale de Lausanne in 2018. He is Postdoctoral researcher in the EcoVision group at ETH Z\"urich since 2018. His research interests include several machine learning topics, such as Bayesian inference method and deep learning, with particular emphasis on applications related to remote sensing an environmental monitoring.
\end{IEEEbiography}

\begin{IEEEbiography}[{\includegraphics[width=1in,height=1.25in,clip,keepaspectratio]{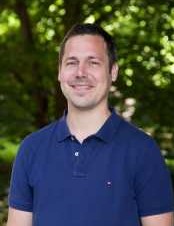}}]{Jan Dirk Wegner}
is associate professor at University of Zurich and head of the EcoVision Lab at ETH Zurich. Jan was PostDoc (2012-2016) and senior scientist (2017-2020) in the Photogrammetry and Remote Sensing group at ETH Zurich after completing his PhD (with distinction) at Leibniz Universität Hannover in 2011. He was granted multiple awards, among others an ETH Postdoctoral fellowship and the science award of the German Geodetic Commission. Jan was selected for the WEF Young Scientist Class 2020 as one of the 25 best researchers world-wide under the age of 40 committed to integrating scientific knowledge into society for the public good. Jan is chair of the ISPRS II/WG 6 "Large-scale machine learning for geospatial data analysis" and organizer of the CVPR EarthVision workshops.
\end{IEEEbiography}

\begin{IEEEbiography}[{\includegraphics[width=1in,height=1.25in,clip,keepaspectratio]{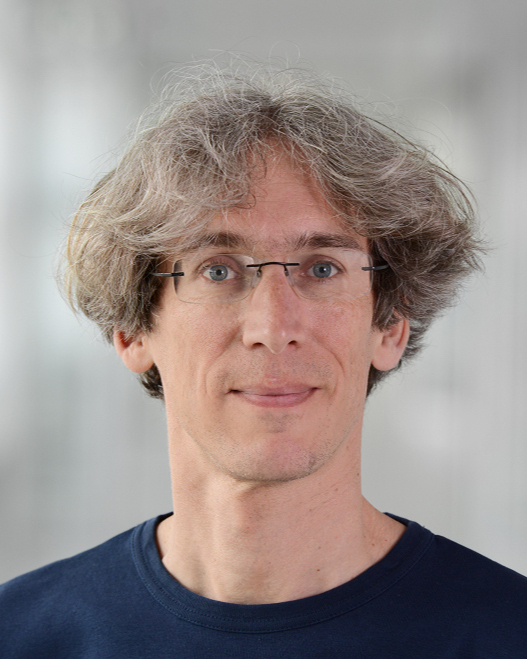}}]{Konrad Schindler}
(M'05–SM'12) received the Diplomingenieur (M.Tech.) degree from Vienna University of Technology, Vienna, Austria, in 1999, and the Ph.D.\ degree from Graz University of Technology, Graz, Austria, in 2003.
He was a Photogrammetric Engineer in the private industry and held researcher positions at Graz University of Technology, Monash University, Melbourne, VIC, Australia, and ETH Z\"urich, Z\"urich, Switzerland. He was an Assistant Professor of Image Understanding with TU Darmstadt, Darmstadt, Germany, in 2009. Since 2010, he has been a Tenured Professor of Photogrammetry and Remote Sensing with ETH Z\"urich. His research interests include computer vision, photogrammetry, and remote sensing.
\end{IEEEbiography}
\vfill

\clearpage 
\renewcommand{\thesection}{\Alph{section}}
\setcounter{section}{1}

\section*{Appendix}

\subsection{RNN Cells Dynamics}\label{app:dynamics}
In the following, we provide more detailed insights about the updating rules of the tested cell types.

\noindent
Vanilla RNN update rule:
\begin{equation}
\vh_t^l = \tanh (\mW_{x}\vh_t^{l-1}+ \mW_{h}\vh_{t-1}^l+ \vb) 
\end{equation}
%
%Jacobians:
%\begin{equation}
%J_t^l = \mD_{\tanh(\mW_{x}\vh_t^{l-1}+ \mW_{h}\vh_{t-1}^l+ \vb)'}\mW_{x} 
%\end{equation}
%\begin{equation}
%H_t^l = \mD_{\tanh(\mW_{x}\vh_t^{l-1}+ \mW_{h}\vh_{t-1}^l+ \vb)'}\mW_{h} 
%\end{equation}
%
LSTM update rule:
%LSTM update rule:
%------------LSTM dynamics-------------------------
\begin{align}
&\vi_t^l = \sigma(\mW_{xi}\vh_t^{l-1}+ \mW_{hi}\vh_{t-1}^l+ \vb_i)\\
&\vf_t^l = \sigma(\mW_{xf}\vh_t^{l-1}+ \mW_{hf}\vh_{t-1}^l+ \vb_f)\\
&\vo_t^l = \sigma(\mW_{xo}\vh_t^{l-1}+ \mW_{ho}\vh_{t-1}^l+ \vb_o)\\
&\vz_t^l = \tanh (\mW_{xz}\vh_t^{l-1}+ \mW_{hz}\vh_{t-1}^l+ \vb_z)\\
&\vc_t^l = \vf_t^l \circ \vc_{t-1}^l + \vi_t^l \circ \vz_t^l\\ 
&\vh_t^l = \vo_t^l \circ \tanh(\vc_t^l).
\end{align}
%
%Jacobians:
%\begin{equation}
%J^{l}_t = \mD_{\tanh(\vc_t^l)}\mD_{(\vo_{t}^{l})'}\mW_{xo} +  \mD_{\tanh(\vc_t^l)'}\mD_{\vo_t^l}(\mD_{\vc_{t-1}^l}\mD_{(\vf_t^l)'}\mW_{xf} + \mD_{\vz_t^l}\mD_{(\vi_t^l)'}\mW_{xi} + \mD_{\vi_t^l}\mD_{(\vz_t^l)'}\mW_{xz} )
%\label{eq:H_lstm}
%\end{equation}
%\begin{equation}
%H^{l}_t = \mD_{\tanh(\vc_t^l)}\mD_{(\vo_{t}^{l})'}\mW_{ho} +  \mD_{\tanh(\vc_t^l)'}\mD_{\vo_t^l}(\mD_{\vc_{t-1}^l}\mD_{(\vf_t^l)'}\mW_{hf} + \mD_{\vz_t^l}\mD_{(\vi_t^l)'}\mW_{xi} + \mD_{\vi_t^l}\mD_{(\vz_t^l)'}\mW_{hz} )
%\label{eq:J_lstm}
%\end{equation}
%
%
LSTM with only forget gate, update rule:
%LSTM with only forget gate update rule:
\begin{align}
&\vf_t^l = \sigma(\mW_{xf}\vh_t^{l-1}+ \mW_{hf}\vh_{t-1}^l+ \vb_f)\\
&\vz_t^l = \tanh (\mW_{xz}\vh_t^{l-1}+ \mW_{hz}\vh_{t-1}^l+ \vb_z)\\
&\vh_t^l = \tanh(\vf_t^l \circ \vh_{t-1}^l + (1 - \vf_t^l) \circ \vz_t^l ) \end{align}
GRU update rule:
%GRU update rule:
\begin{align}
\vz_t^l& = \sigma(\mW_{xz}\vh_t^{l-1}+ \mW_{hz}\vh_{t-1}^l+ \vb_z)\\
\vr_t^l& = \sigma (\mW_{xr}\vh_t^{l-1}+ \mW_{hr}\vh_{t-1}^l+ \vb_r)\\ 
\vh_t^l& = (1 - \vz_t^l) \circ \vh_{t-1}^l +\\ \nonumber
&+\vz_t^l \circ \tanh \big( \mW_{xh} \vh_{t}^{l-1} +\mW_{hh}(\vr_t^l \circ \vh_{t-1}^l) + \vb_h   \big)
\end{align}
%
%
%BN-STAR update rule:
%\begin{align}
%\vz_t^l &=  \tanh( BN( \mW_z \vh_{t}^{l-1}) + \vb_z) \label{eq:star_z}\\ 
%\vk_{t}^l &=  \sigma( BN(\mW_x\vh_{t}^{l-1}) + BN(\mW_h\vh_{t-1}^l) + \vb_k )  \label{eq:star_k}\\
%\vh_{t}^l &= \tanh \big( (1 - \vk_{t}^l)\circ \vh_{t-1}^l + \vk_{t}^l\circ \vz_t^l \big). \label{eq:star_h}
%\end{align}
%
%Jacobians:
STAR Jacobians:
\begin{align}
J_t^l =& \mD_{\tanh(\vh_{t-1}^l + \vk_{t}^l\circ(\vz_t^l-\vh_{t-1}^l))'}\\ \nonumber
&\cdot(\mD_{\vz_t^l-\vh_{t-1}^l}\mD_{(\vk_t^l)'}\mW_{x}  + \mD_{\vk_t^l}\mD_{(\vz_t^l)'}\mW_{z} )
\label{eq:J_star}
\\
H_t^l =& \mD_{\tanh(\vh_{t-1}^l + \vk_{t}^l\circ(\vz_t^l-\vh_{t-1}^l))'}\\ \nonumber
&\cdot(\mI + \mD_{\vz_t^l-\vh_{t-1}^l} \mD_{(\vk_t^l)'}\mW_h - \mD_{\vk_t^l} ) \label{eq:H_star}
\end{align}
%The expressions for the Jacobians of STAR cell are given in Eqs.~(\ref{eq:J_star},\ref{eq:H_star}).
%
%\begin{table*}
%\begin{align}
%&J_t^l = \mD_{\tanh(\vh_{t-1}^l + \vk_{t}^l\circ(\vz_t^l-\vh_{t-1}^l))'}(\mD_{\vz_t^l-\vh_{t-1}^l}\mD_{(\vk_t^l)'}\mW_{x}  + \mD_{\vk_t^l}\mD_{(\vz_t^l)'}\mW_{z} )
%\label{eq:J_star}
%\\
%&H_t^l = \mD_{\tanh(\vh_{t-1}^l + \vk_{t}^l\circ(\vz_t^l-\vh_{t-1}^l))'} (\mI + \mD_{\vz_t^l-\vh_{t-1}^l} \mD_{(\vk_t^l)'}\mW_h - \mD_{\vk_t^l} ) %\label{eq:H_star}
%\end{align}
%\end{table*}
%
%
%\begin{equation}
%J^{l}_t = \mD_{\tanh(\vf_t^l \circ \vh_{t-1}^l + (1 - \vf_t^l) \circ \vz_t^l)'}(\mD_{\vh_{t-1}^l-\vz_t^l}\mD_{(\vf_t^l)'}\mW_{xf} +  \mD_{1-\vf_t^l}\mD_{(\vz_t^l)'}\mW_{xz} )
%\label{}
%\end{equation}
%
%\begin{equation}
%H^{l}_t = \mD_{\tanh(\vf_t^l \circ \vh_{t-1}^l + (1 - \vf_t^l) \circ \vz_t^l)'}(\mD_{\vh_{t-1}^l-\vz_t^l}\mD_{(\vf_t^l)'}\mW_{hf} +  \mD_{1-\vf_t^l}\mD_{(\vz_t^l)'}\mW_{hz} )
%\label{}
%\end{equation}
%
%

\noindent
Convolutional STAR:
We briefly describe the convolutional version of our proposed cell. The main difference is matrix multiplications now become convolutional operations. The dynamics of the convSTAR cell is given in the following equations.
\begin{align}
&\tK_{t}^{l} = \sigma( \tW_x*\tH_t^{l-1} + \tW_h*\tH_{t-1}^{l} + \tB_K )\\
&\tZ_t^{l} =  \tanh(\tW_z*\tH_t^{l-1} + \tB_z)\\
&\tH_t^{l} = \tanh(\tH_{t-1}^{l} + \tK_{t}^{l}\circ\big(\tZ_t^{l}-\tH_{t-1}^{l})\big) 
\end{align}

\subsection{Further Numerical Gradient Propagation Analysis}
In this section, we extend the numerical simulations of the gradient propagation in the unfolded recurrent neural network to two further cell architectures, namely the GRU~\cite{gruX} and the LSTM with only forget gate for the synthetic dataset (Sec.~\ref{suppsynth}); and for the real dataset, MNIST (Sec.~\ref{suppmnist}).% (dynamics can be found in Supplementary material~\ref{app:dynamics}). 

\subsubsection{Synthetic Dataset}\label{suppsynth}

\begin{figure}[h]
    \centering
    \renewcommand\tabcolsep{0pt}
    \begin{tabular}{p{6mm}ccc}
\rotatebox{90}{$\quad\,\,$final output loss} &
\includegraphics[width=0.32\columnwidth]{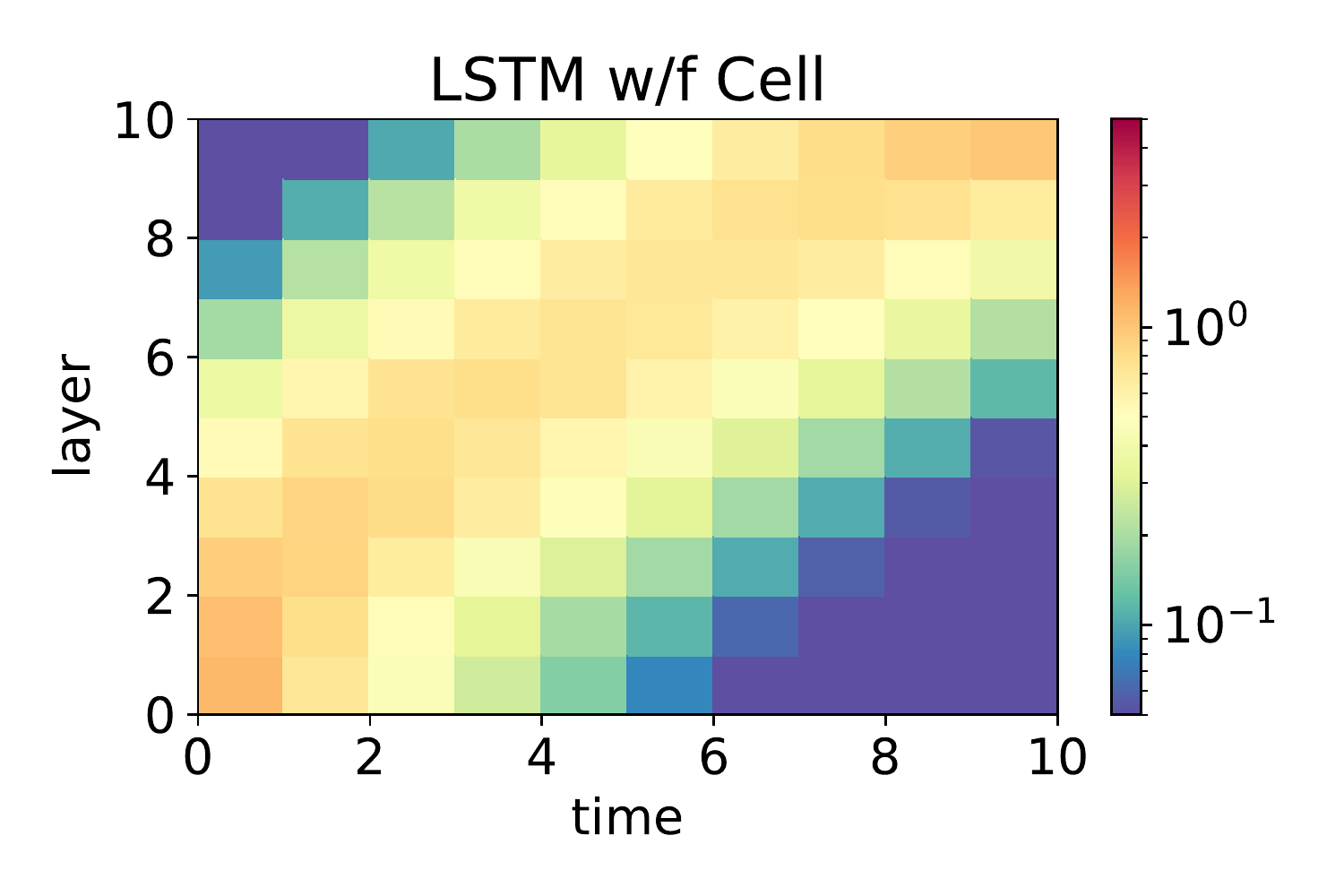} &
\includegraphics[width=0.32\columnwidth]{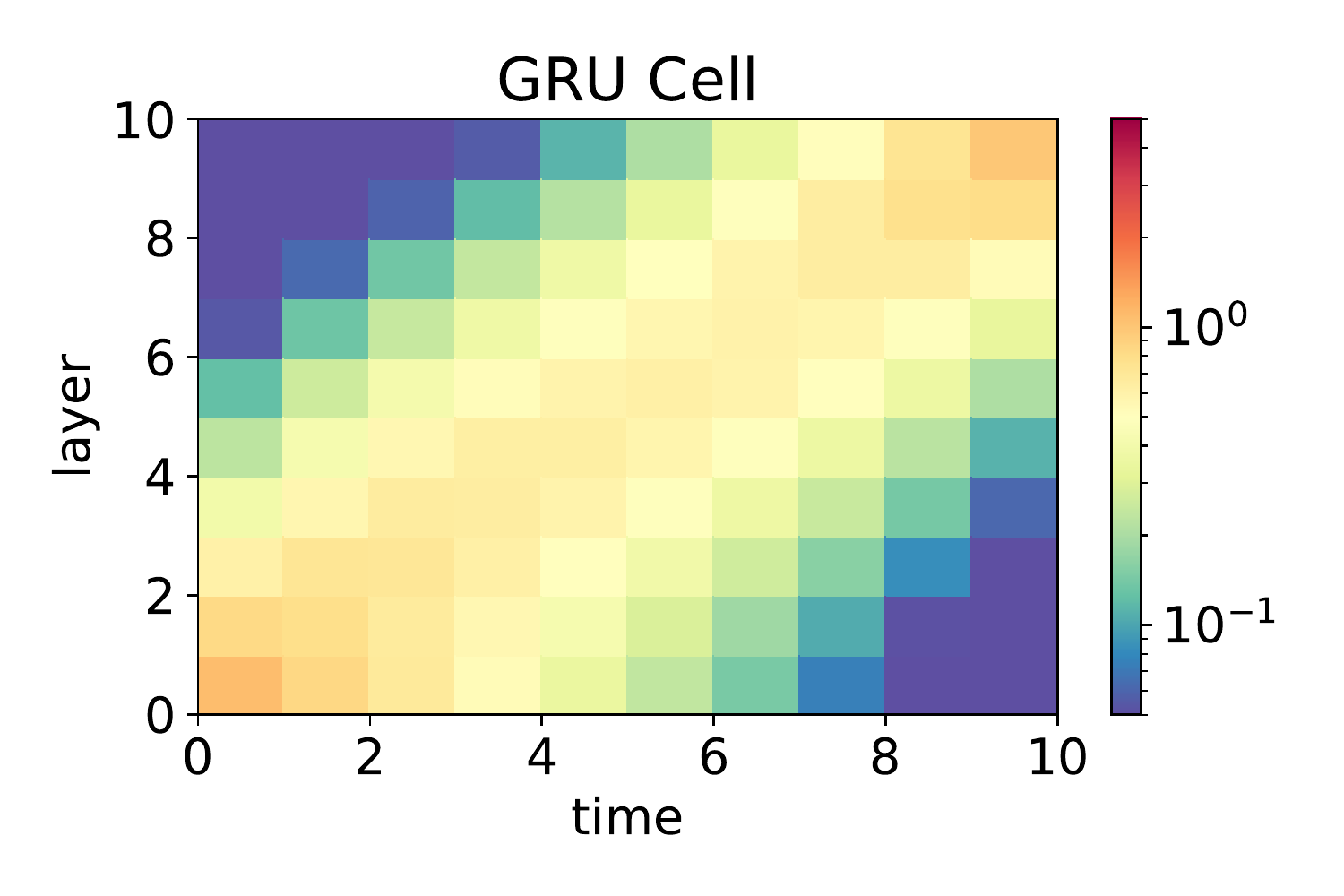} &
\includegraphics[width=0.32\columnwidth]{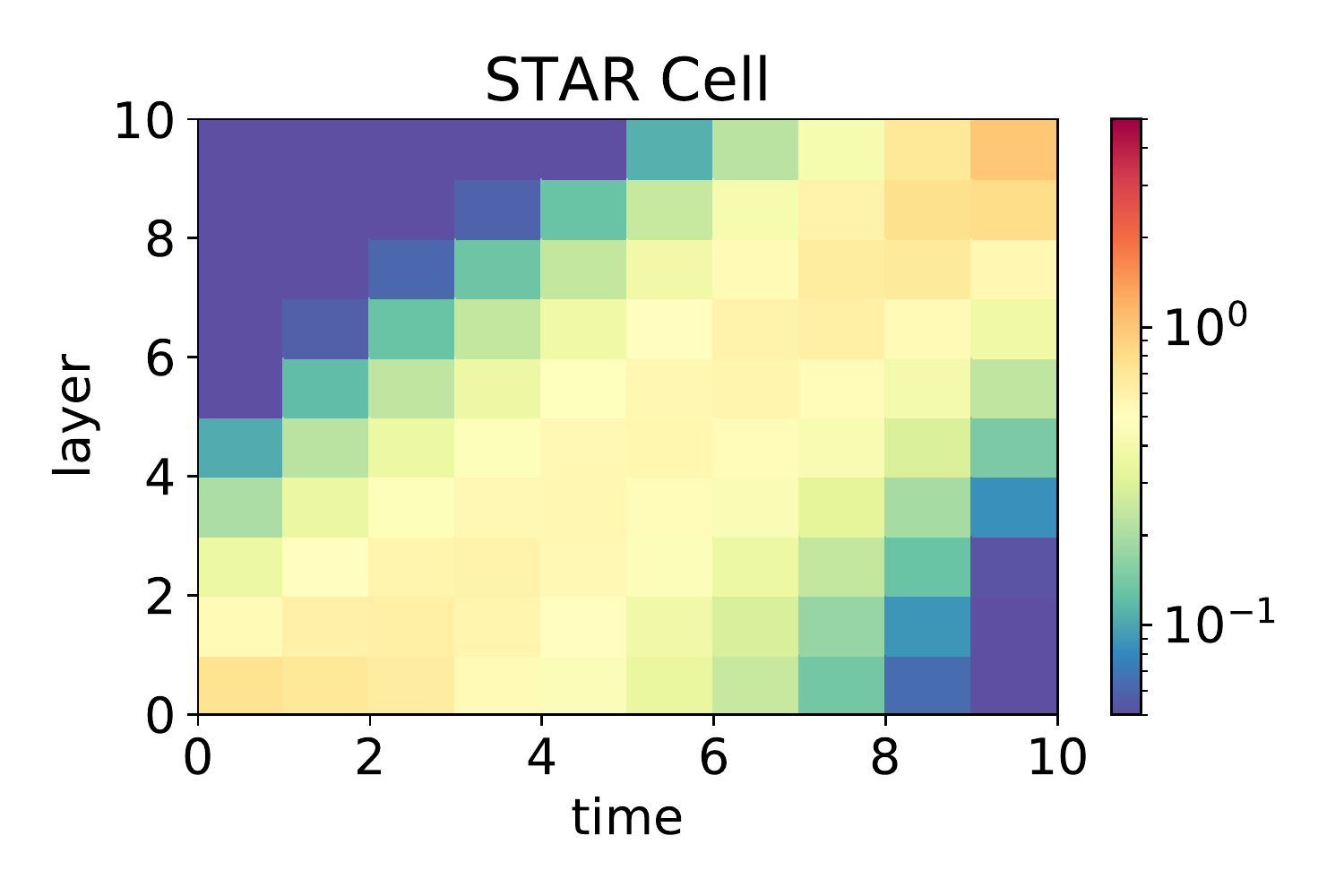} \\
\rotatebox{90}{$\qquad$sequence loss} &
\includegraphics[width=0.32\columnwidth]{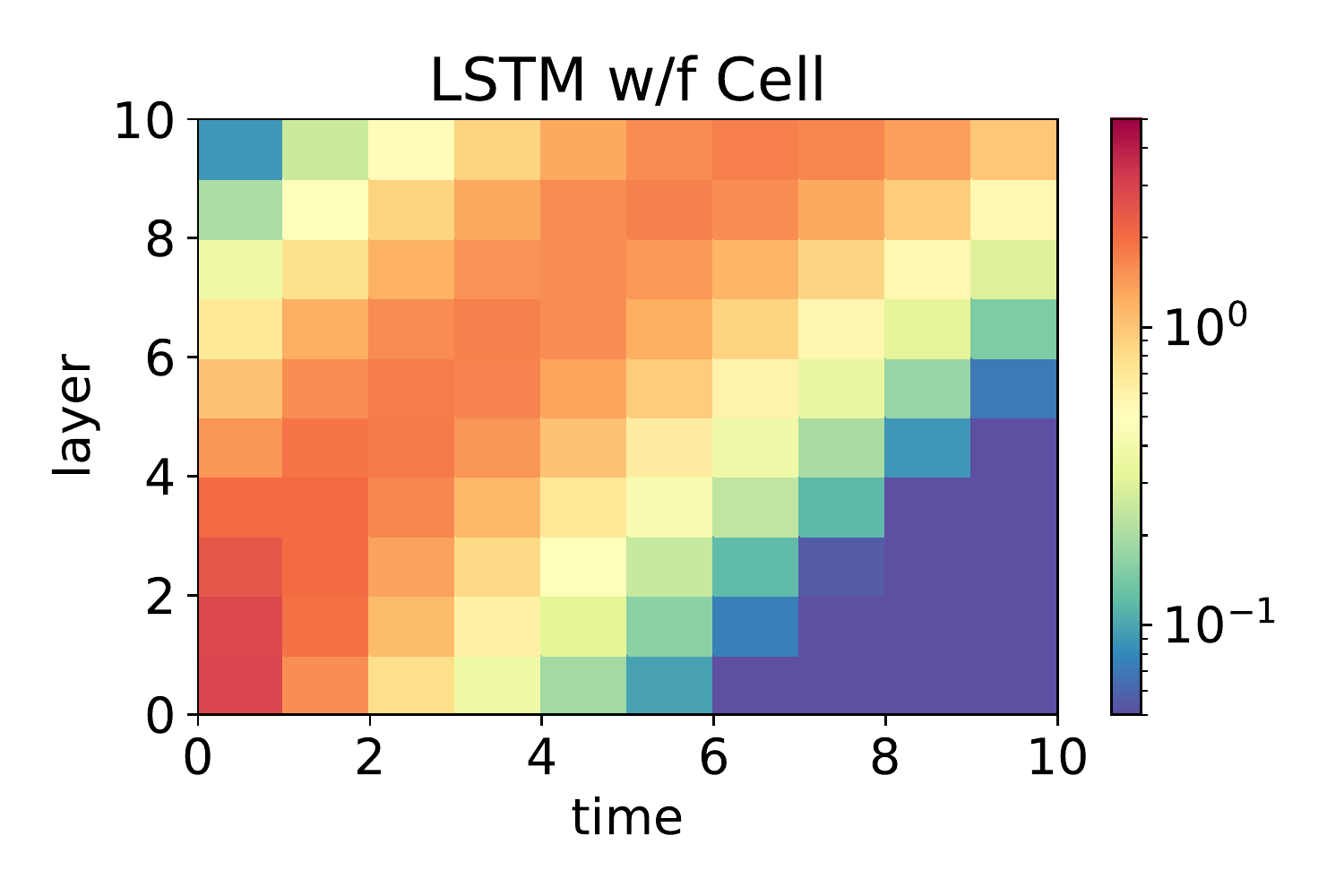} &
\includegraphics[width=0.32\columnwidth]{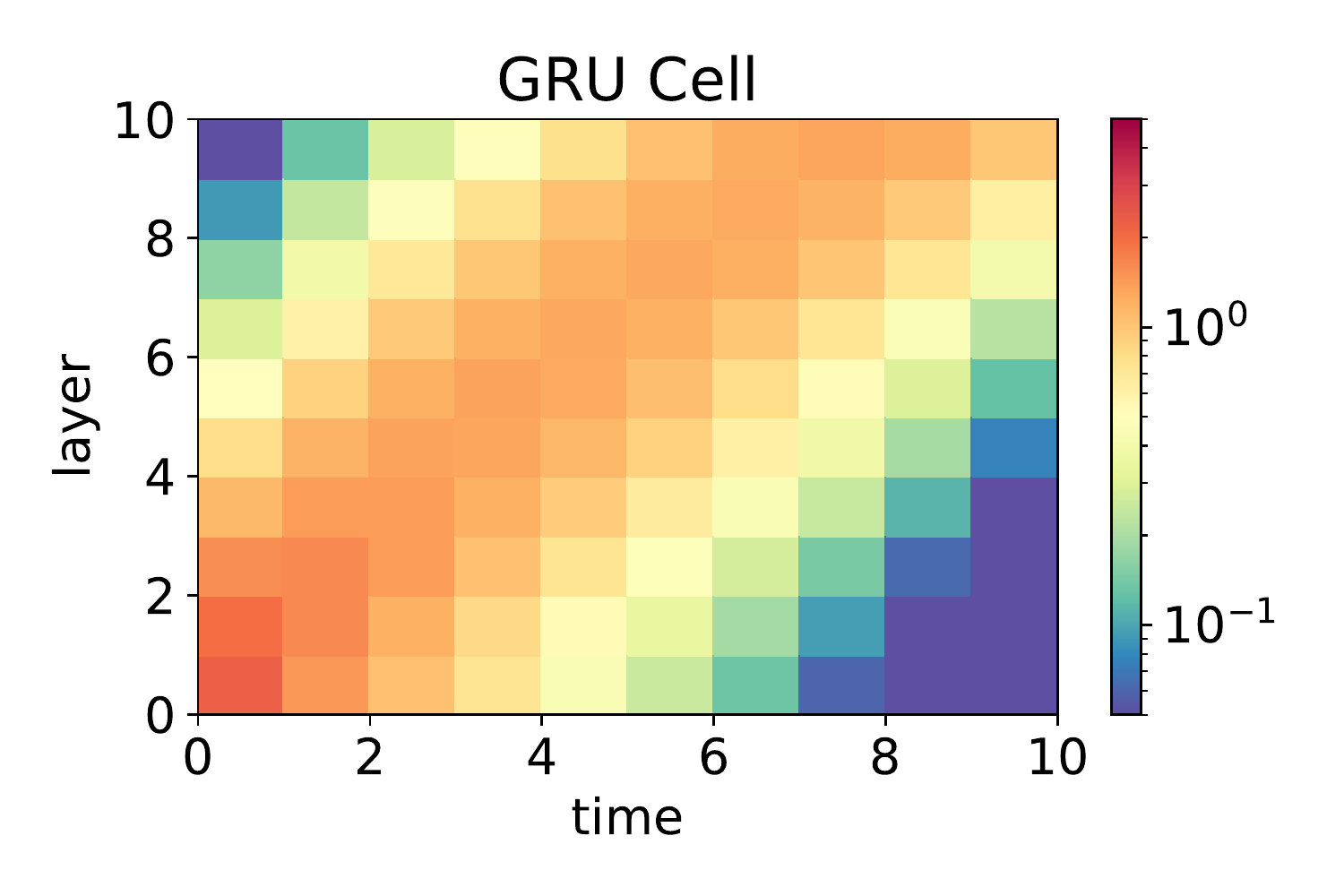} &
\includegraphics[width=0.32\columnwidth]{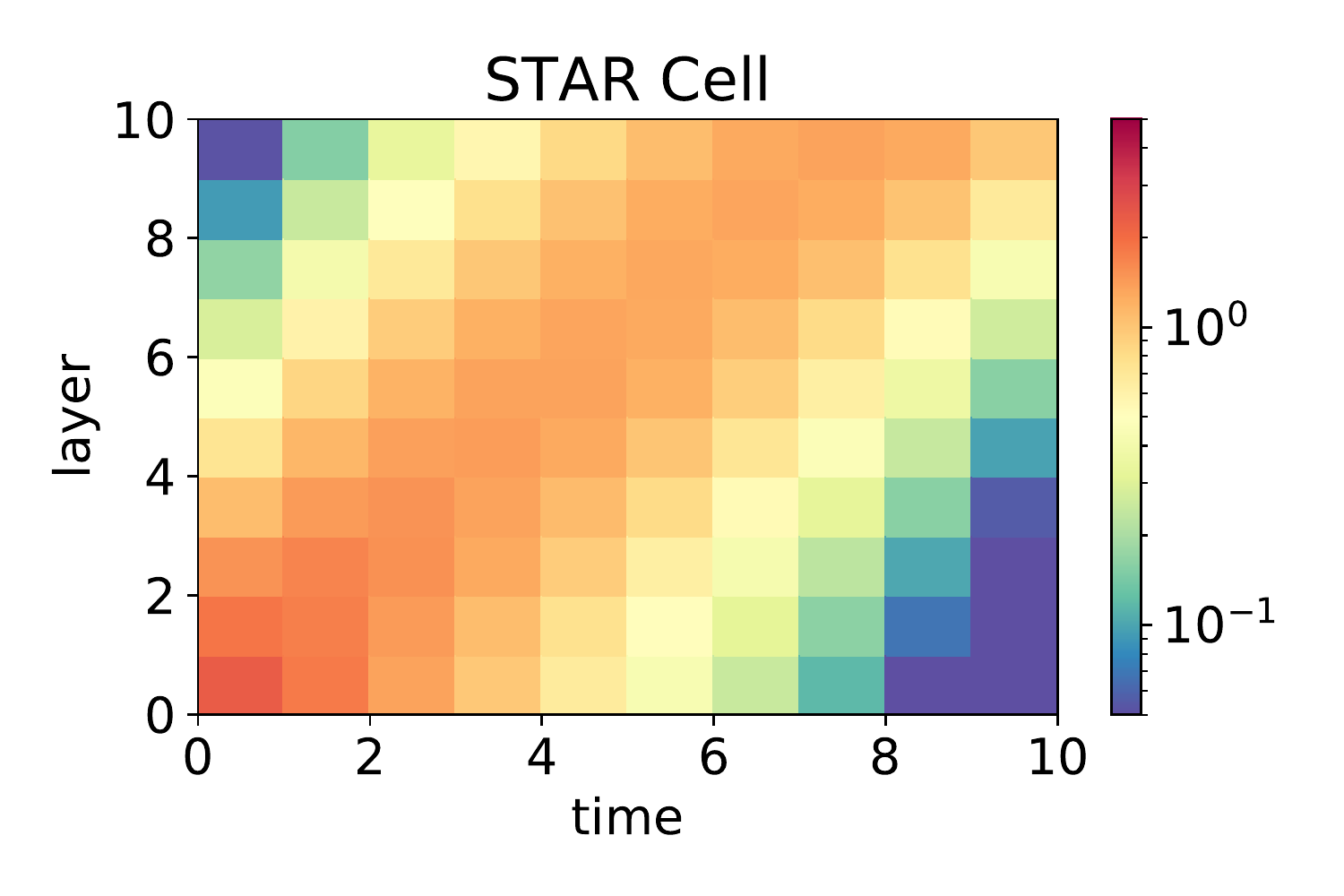} \\
%& LSTMwo/o & GRU & STAR \\
\end{tabular}
    \caption{Mean gradient magnitude w.r.t.\ the parameters for LSTM with only forget gate, GRU, and the proposed STAR cell. \emph{top row:} loss $\mathcal{L}(h^L_T)$ only on final prediction. \emph{bottom row:} loss $\mathcal{L}(h^L_1\hdots h^L_T)$ over all time steps.}
    \label{fig:simulation_app_mean}
\end{figure}

\begin{figure}[h]
    \centering
    \renewcommand\tabcolsep{0pt}
    \begin{tabular}{p{6mm}ccc}
\rotatebox{90}{$\quad\,\,$final output loss} &
\includegraphics[width=0.32\columnwidth]{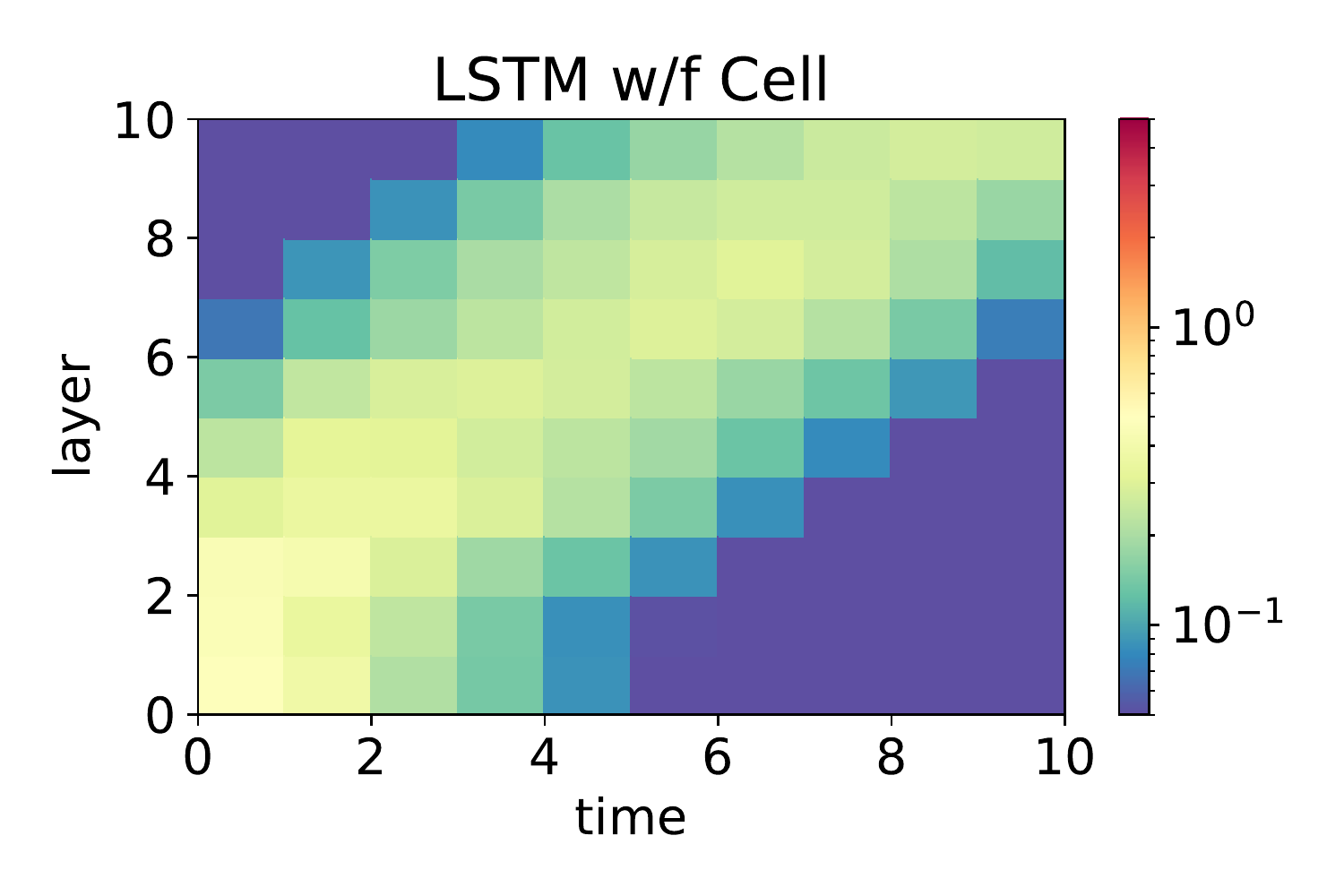} &
\includegraphics[width=0.32\columnwidth]{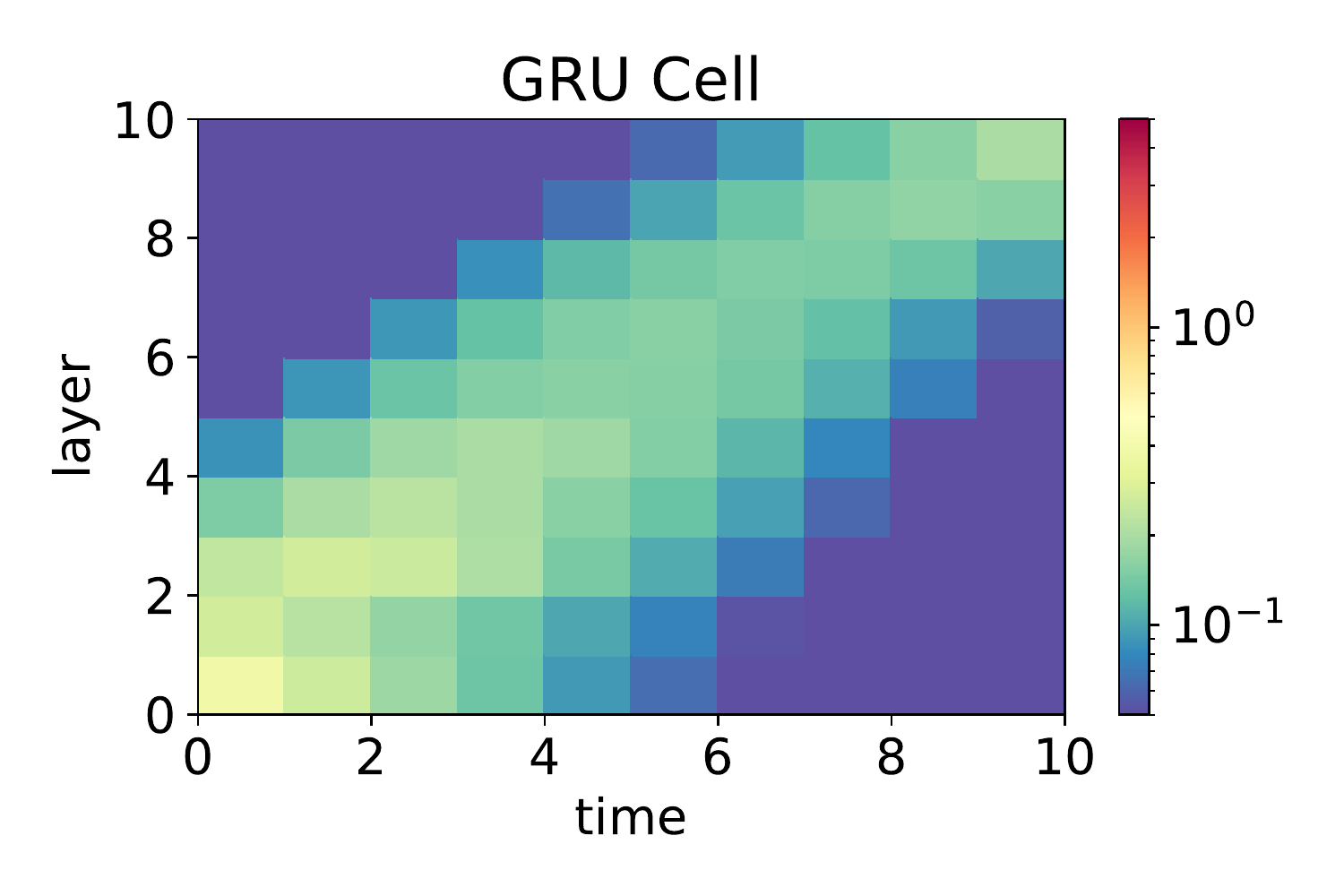} &
\includegraphics[width=0.32\columnwidth]{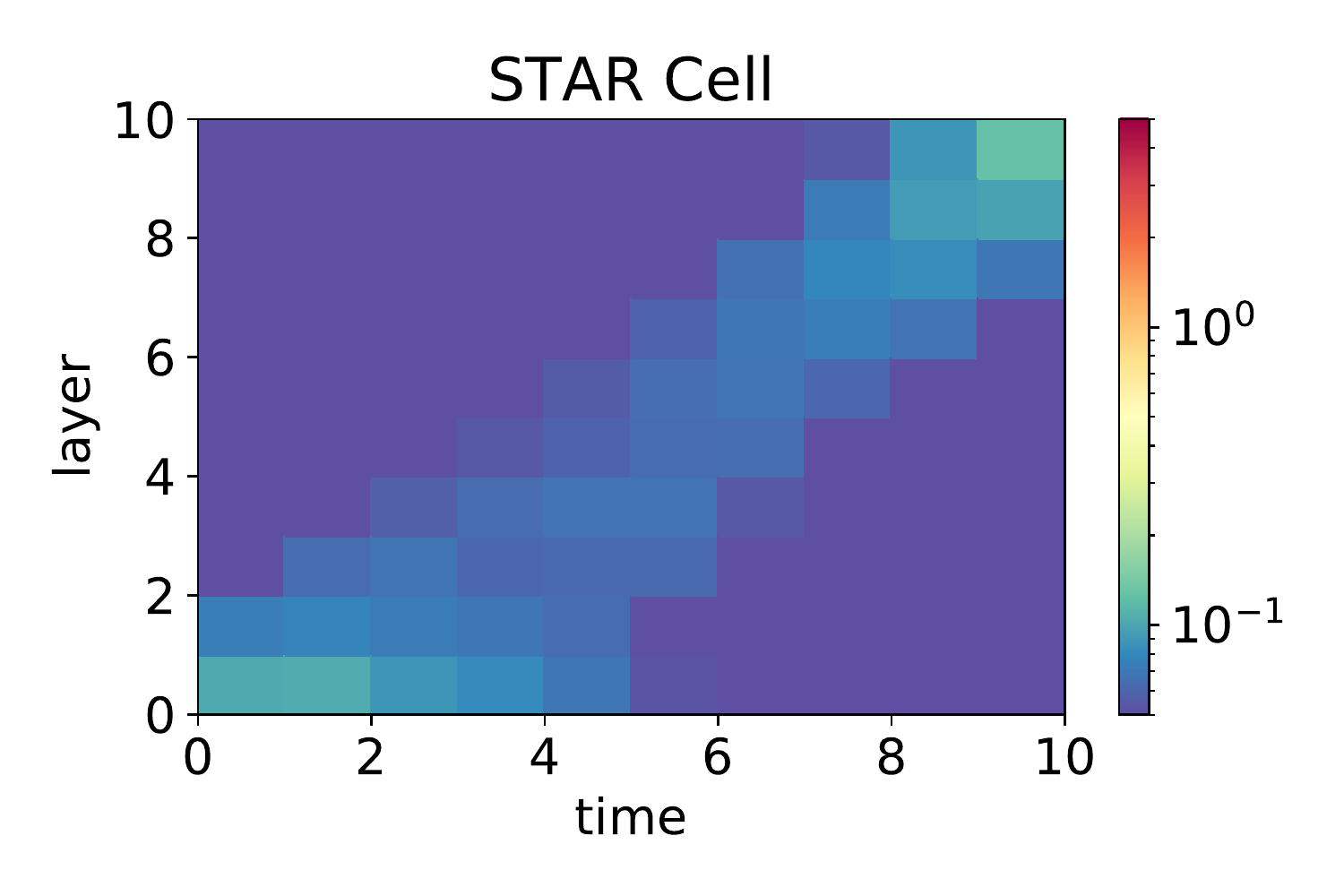} \\
\rotatebox{90}{$\qquad$sequence loss} &
\includegraphics[width=0.32\columnwidth]{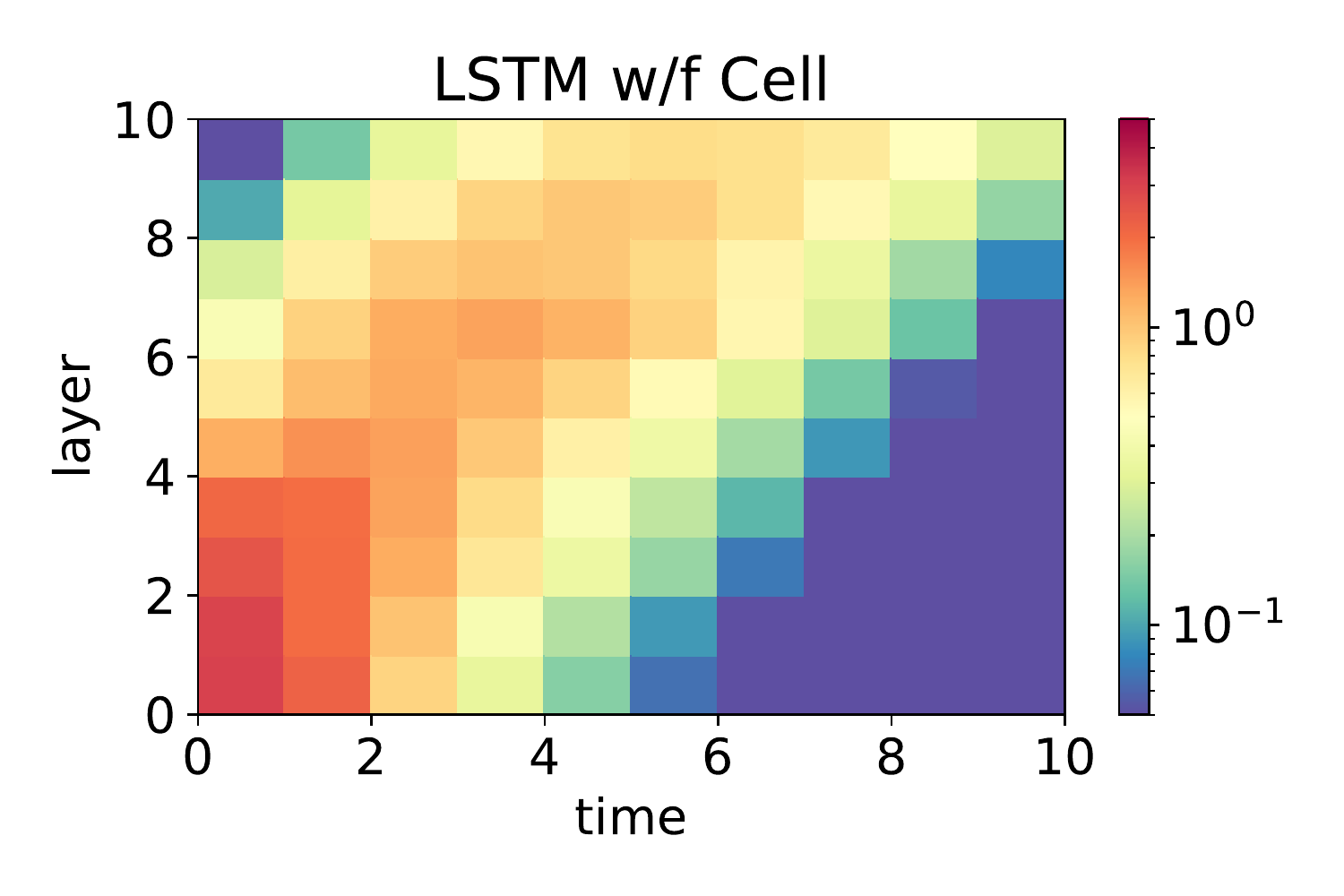} &
\includegraphics[width=0.32\columnwidth]{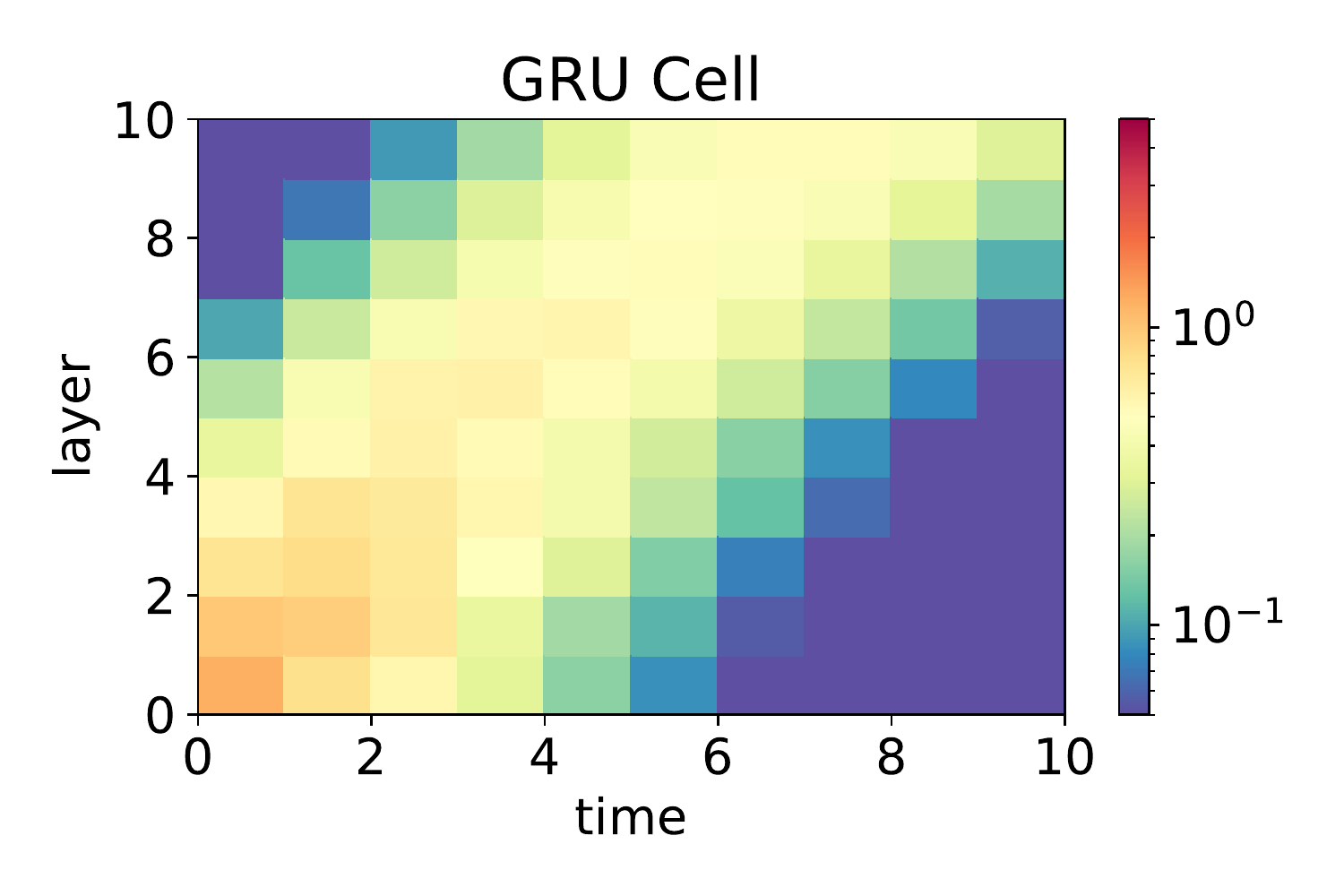} &
\includegraphics[width=0.32\columnwidth]{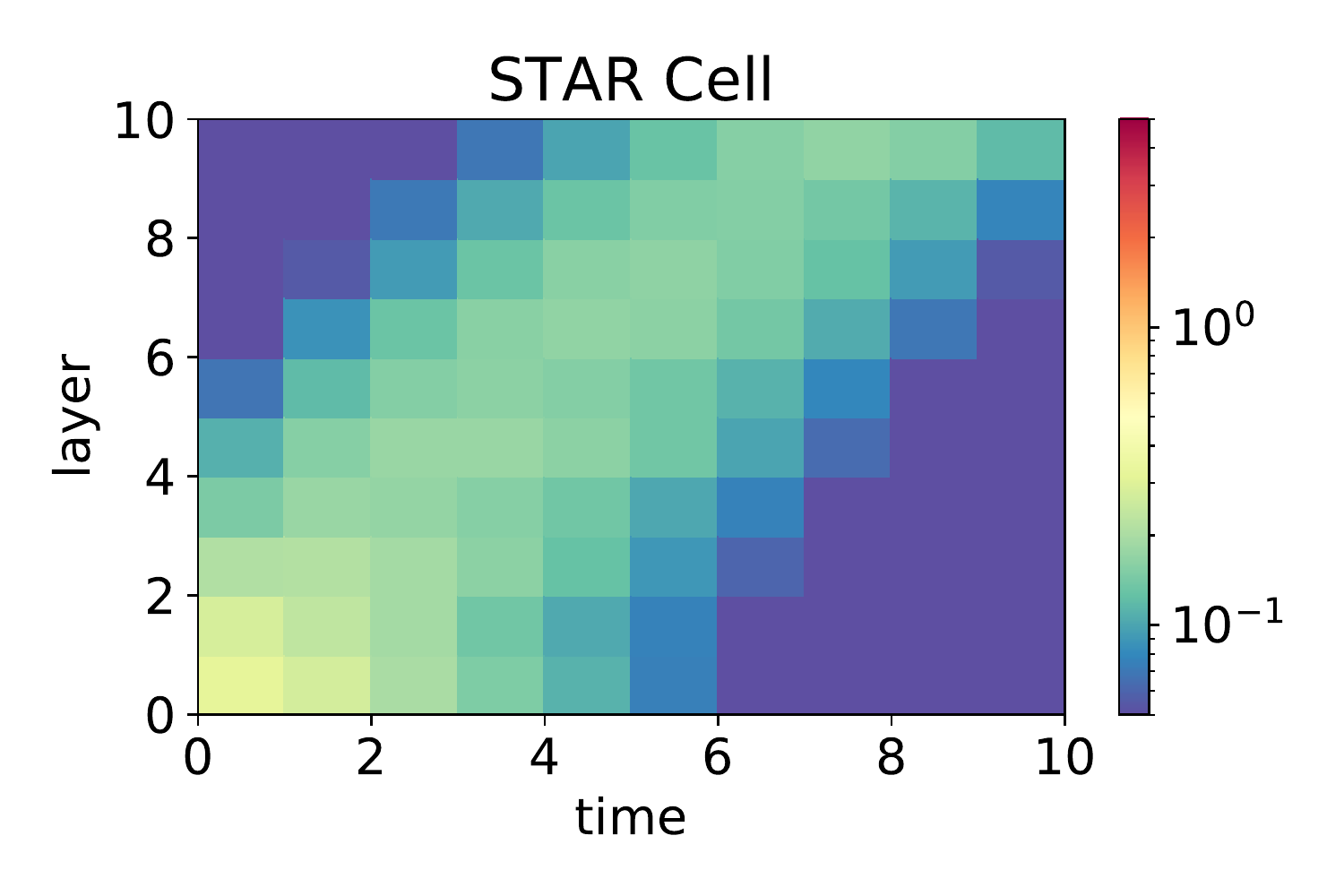} \\
%& LSTMwo/o & GRU & STAR \\
\end{tabular}
 \caption{Mean-normalised standard deviation of gradient magnitude for LSTM with only forget gate, GRU, and the proposed STAR cell.  \emph{top row:} loss $\mathcal{L}(h^L_T)$ only on final prediction. \emph{bottom row:} loss $\mathcal{L}(h^L_1\hdots h^L_T)$ over all time steps.}
    \label{fig:simulation_app_std}
\end{figure}

The setup of the numerical simulations is the same as the one described in Section 3.
As can be seen from Fig.~\ref{fig:simulation_app_mean} the GRU and the LSTM with only forget gate mitigate the attenuation of gradients to some degree. However, we observe that the corresponding standard deviations are much higher, i.e., the gradient norm greatly varies across different runs, see Fig.~\ref{fig:simulation_app_std}.
We found that the gradients within a single run oscillate a lot more, for both LSTMw/f and GRU, and make training unstable which is undesirable. Moreover, the gradient magnitudes evolve very differently for different initial values, meaning that the training is less robust against fluctuations of the random initialisation.

\subsubsection{MNIST Dataset}\label{suppmnist}
 
\begin{figure}
    \centering
    \begin{subfigure}[b]{0.5\textwidth}
        \includegraphics[width=\textwidth]{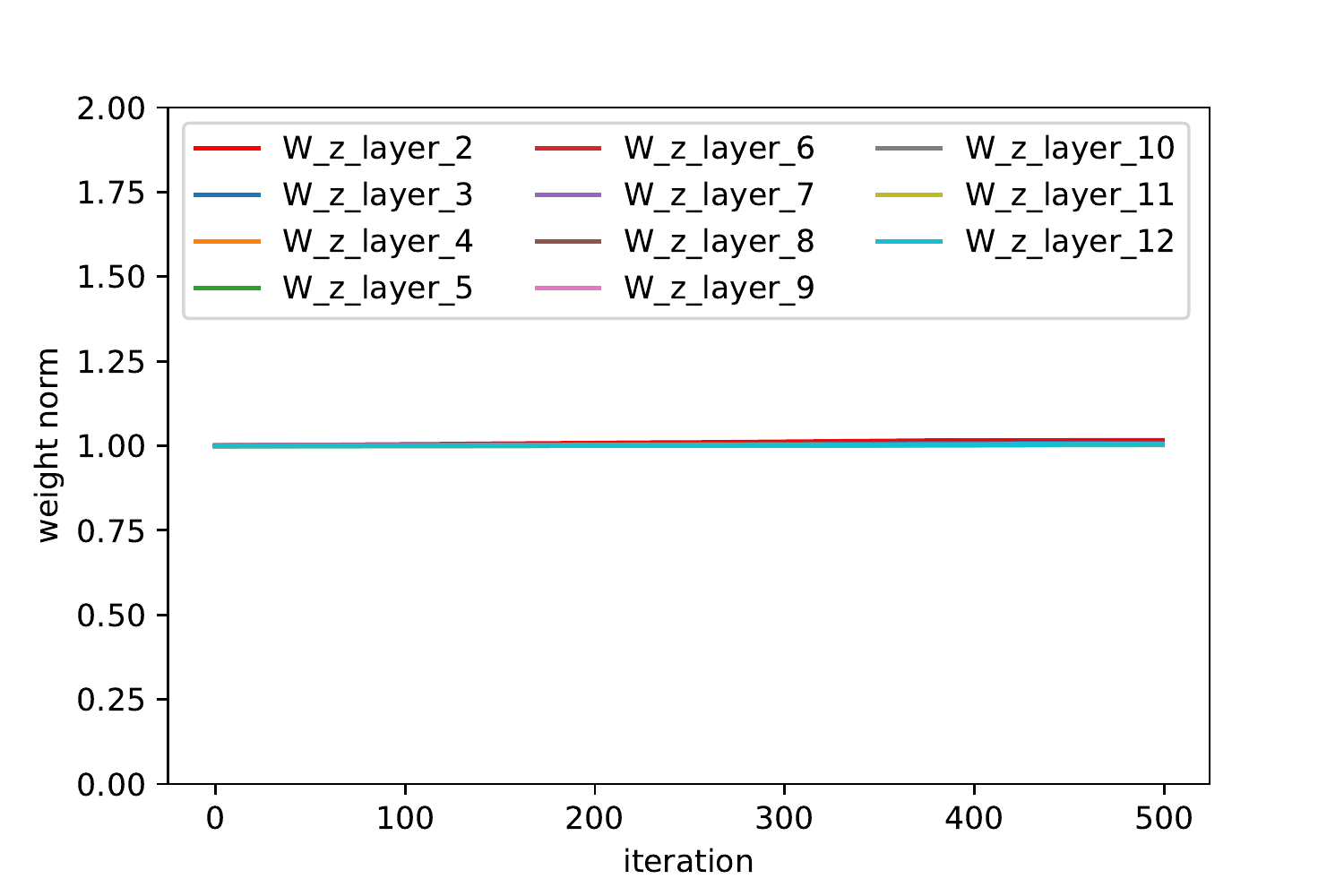}
        \caption{matrix norm, $||\mW_z||$ versus iteration, 1$^\text{st}$ epoch}
        \label{fig:wn_1}
    \end{subfigure}  
      \begin{subfigure}[b]{0.5\textwidth}
        \includegraphics[width=\textwidth]{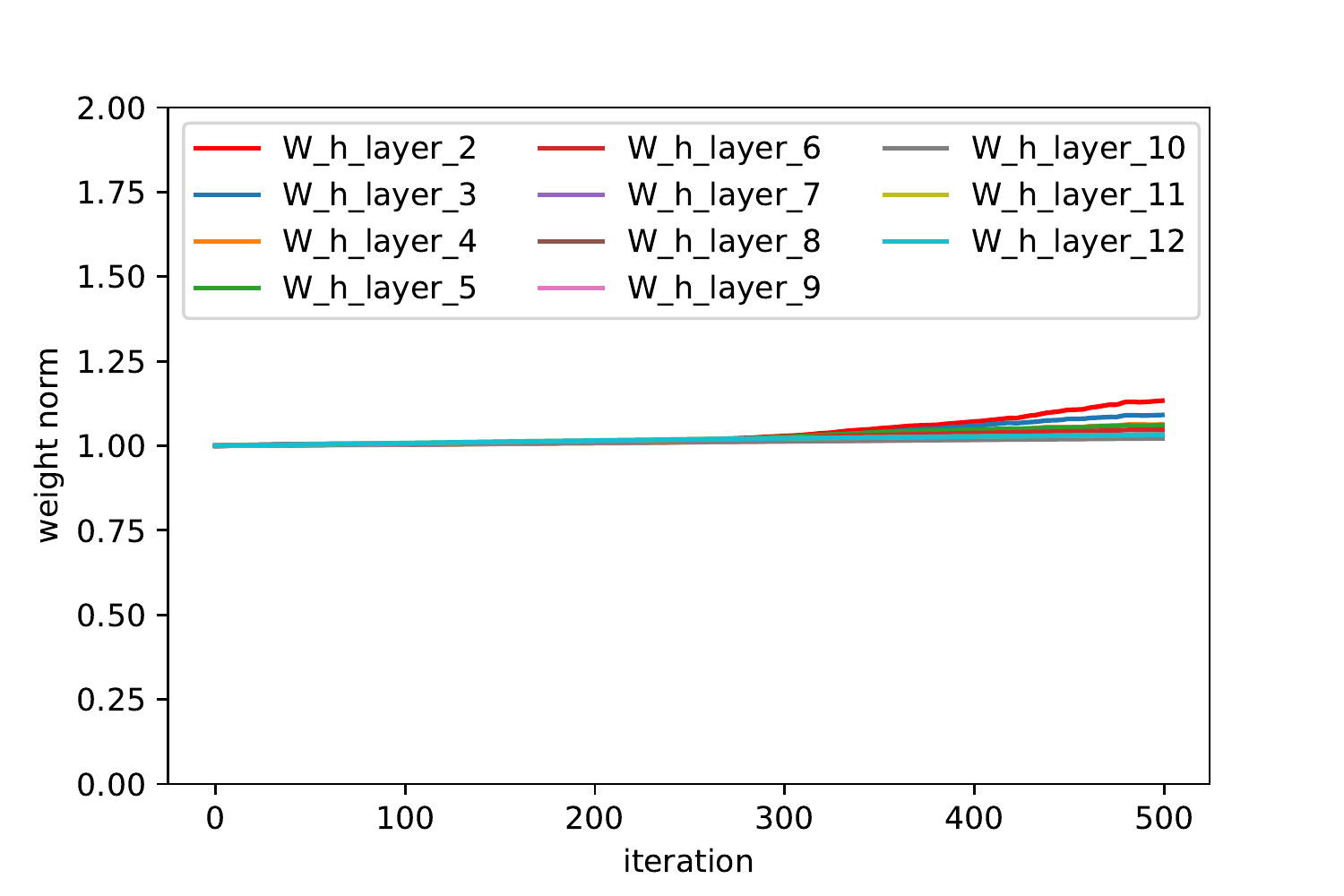}
        \caption{matrix norm, $||\mW_h||$ versus iteration, 1$^\text{st}$ epoch}
        \label{fig:wn_2}
    \end{subfigure}
    \begin{subfigure}[b]{0.5\textwidth}
        \includegraphics[width=\textwidth]{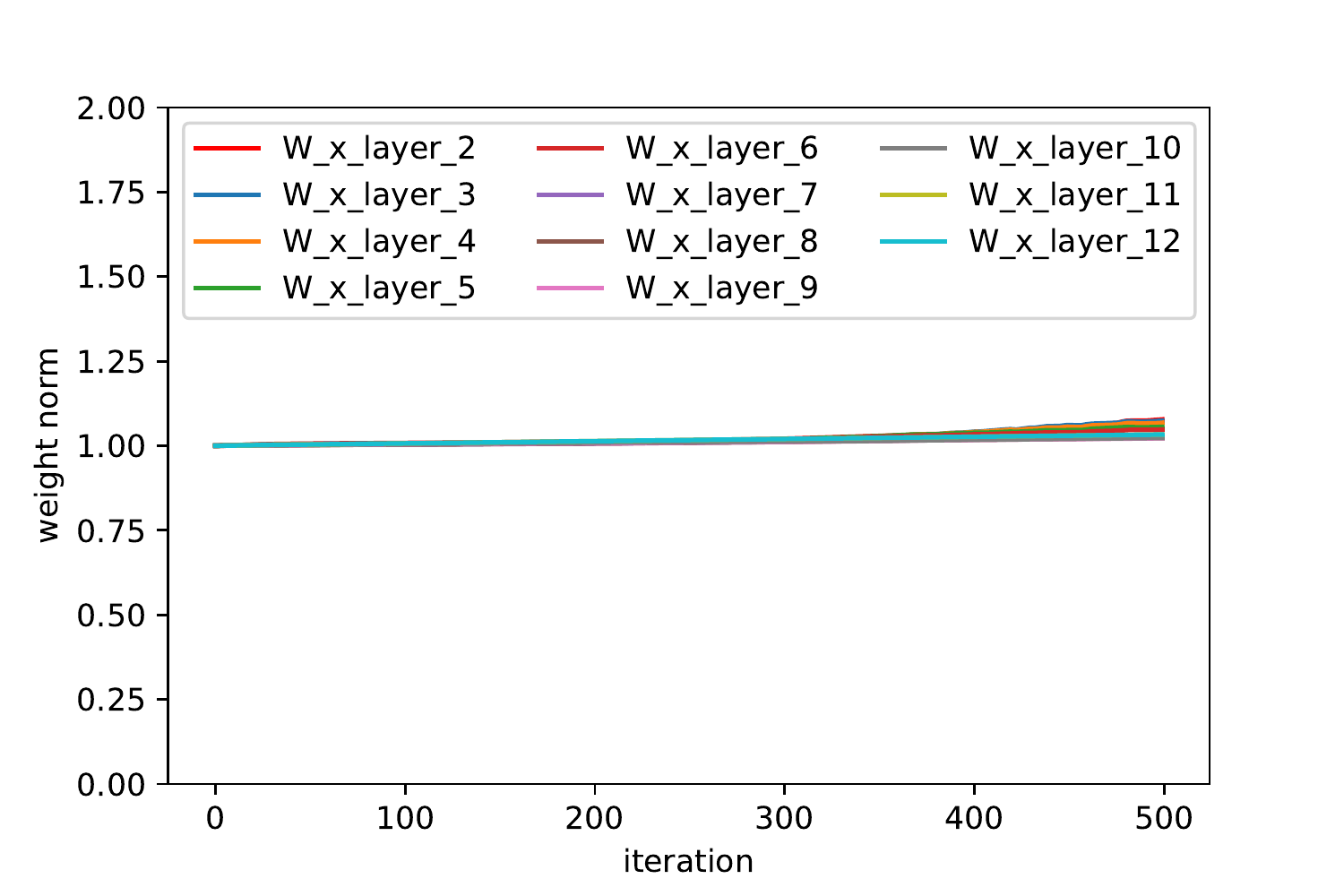}
        \caption{matrix norm, $||\mW_x||$ versus iteration, 1$^\text{st}$ epoch}
        \label{fig:wn_3}
    \end{subfigure}  
    \caption{Weight matrix norms of pix-by-pix MNIST during 1$^\text{st}$ epoch, the Hilbert–Schmidt norm, $||\mA_{mxm}||= \sqrt{Tr(\mA\mA^T)}$, divided by $\sqrt{m}$. Different curves correspond different layers.}\label{fig:weight_norms}
\end{figure}

\begin{figure}[t]
    \centering
        \includegraphics[width=\columnwidth]{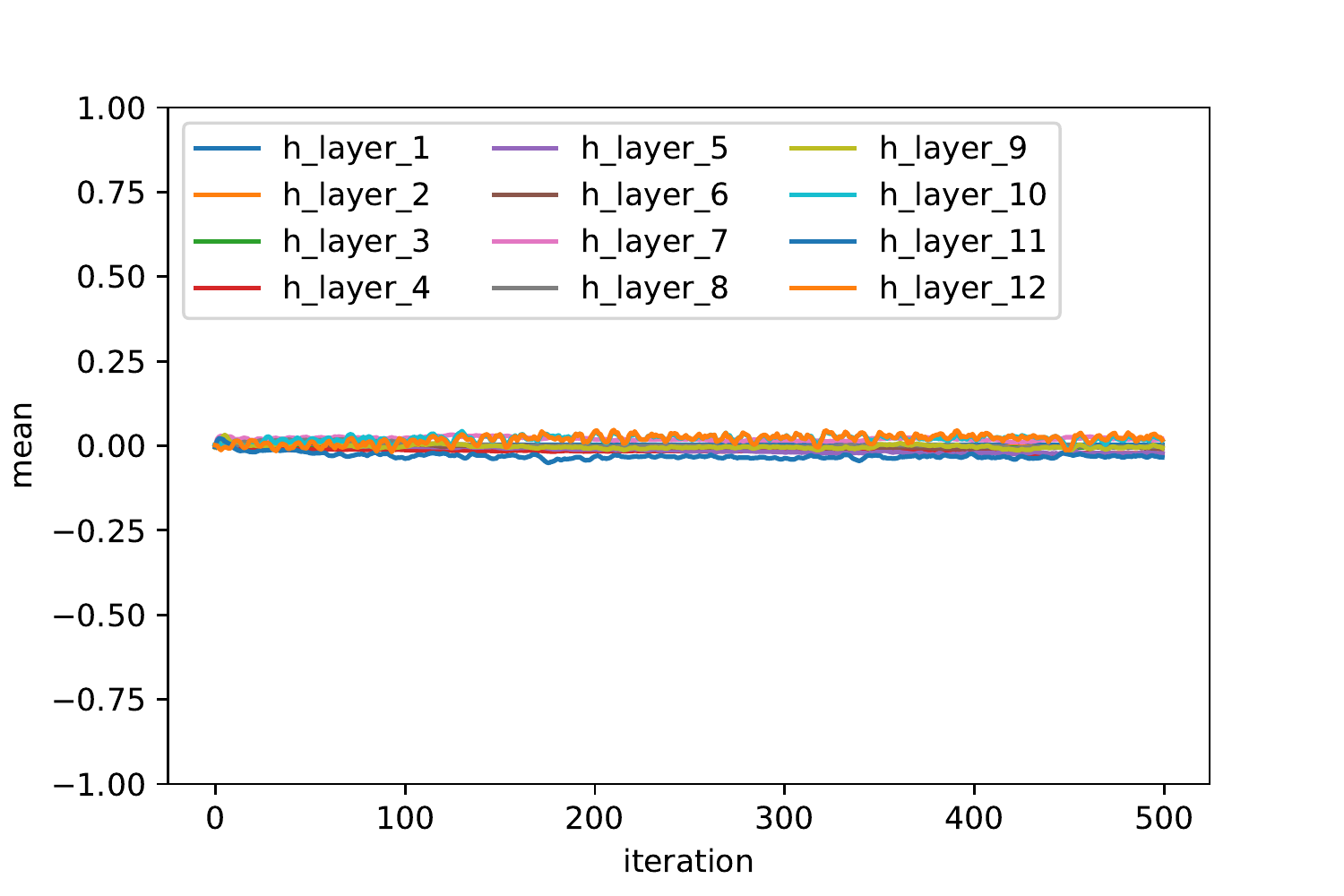}
    \caption{Mean hidden state vector, $\displaystyle  \E_{t,n} [\vh^l]$ of pix-by-pix MNIST during 1$^\text{st}$ epoch. Different curves correspond different layers.}
    \label{fig:mean_h}
\end{figure}

 In this section, we perform the same numerical analysis conducted before but using MNIST as input data. The goal is to verify whether during the first epoch the gradient propagation behaves in the same way as for the synthetic dataset. First, in Fig~\ref{fig:weight_norms} and Fig~\ref{fig:mean_h}, we plot the evolution of the Hilbert-Schmidt norm (also called Frobenius norm) normalized by the square root of the hidden state size and the average hidden state value, respectively. The experiments are conducted using the proposed STAR method with MNIST as input data, the figures show the evolution of the norms and hidden states for the different layers of the recurrent network. The plots show the validity of our assumptions, during the initial training phase and with orthogonal matrix initialization the norm of the matrices is close to one, which translates to singular values close to one. The mean values of the hidden states are instead close to zero, which is consistent with the analysis conducted in Section 3. We reiterate that the plot shows how the \emph{average value} of the hidden state remains close to $0$ during training. This does \emph{not} say that the hidden state is always $\equiv\!0$.
 
 Additionally, we show the gradient propagation in the two-dimensional lattice, as done in Fig.~\ref{fig:simulation_app_mean}, for different cell types with MNIST as input data. We create 12-by-784 latices with twelve layers RNNs. RNN weights are initialized the same way in the real experiments except the forget bias of the LSTM which is set to one (popular initialization scheme for the LSTM) due to numerical instability with the chrono method~\cite{chrono_init}. 
 
\begin{figure}
    \centering
    \begin{subfigure}[b]{0.22\textwidth}
        \includegraphics[width=\textwidth]{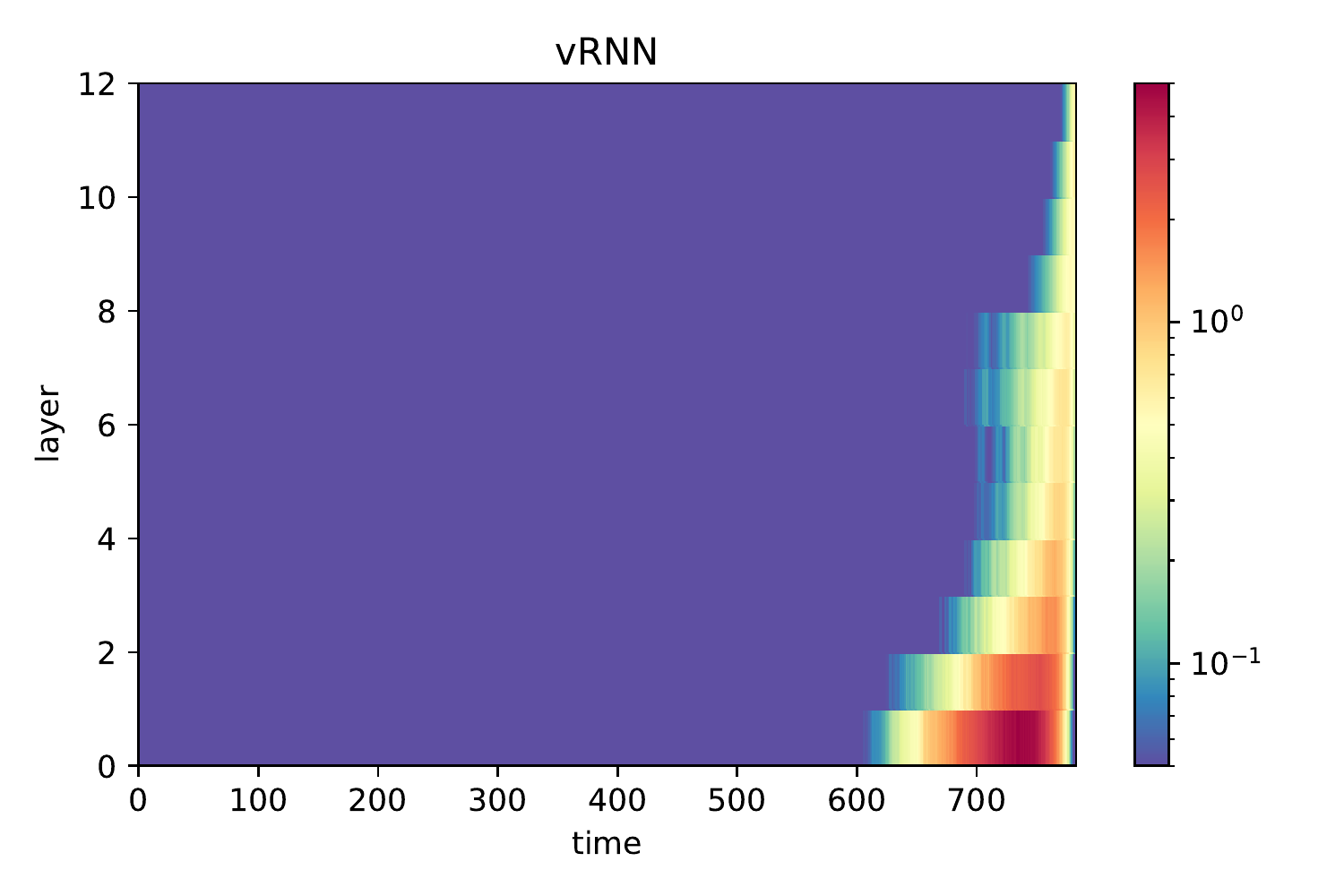}
        \caption{vRNN}
        \label{fig:sim_mnist_vrnn}
    \end{subfigure}  
      \begin{subfigure}[b]{0.22\textwidth}
        \includegraphics[width=\textwidth]{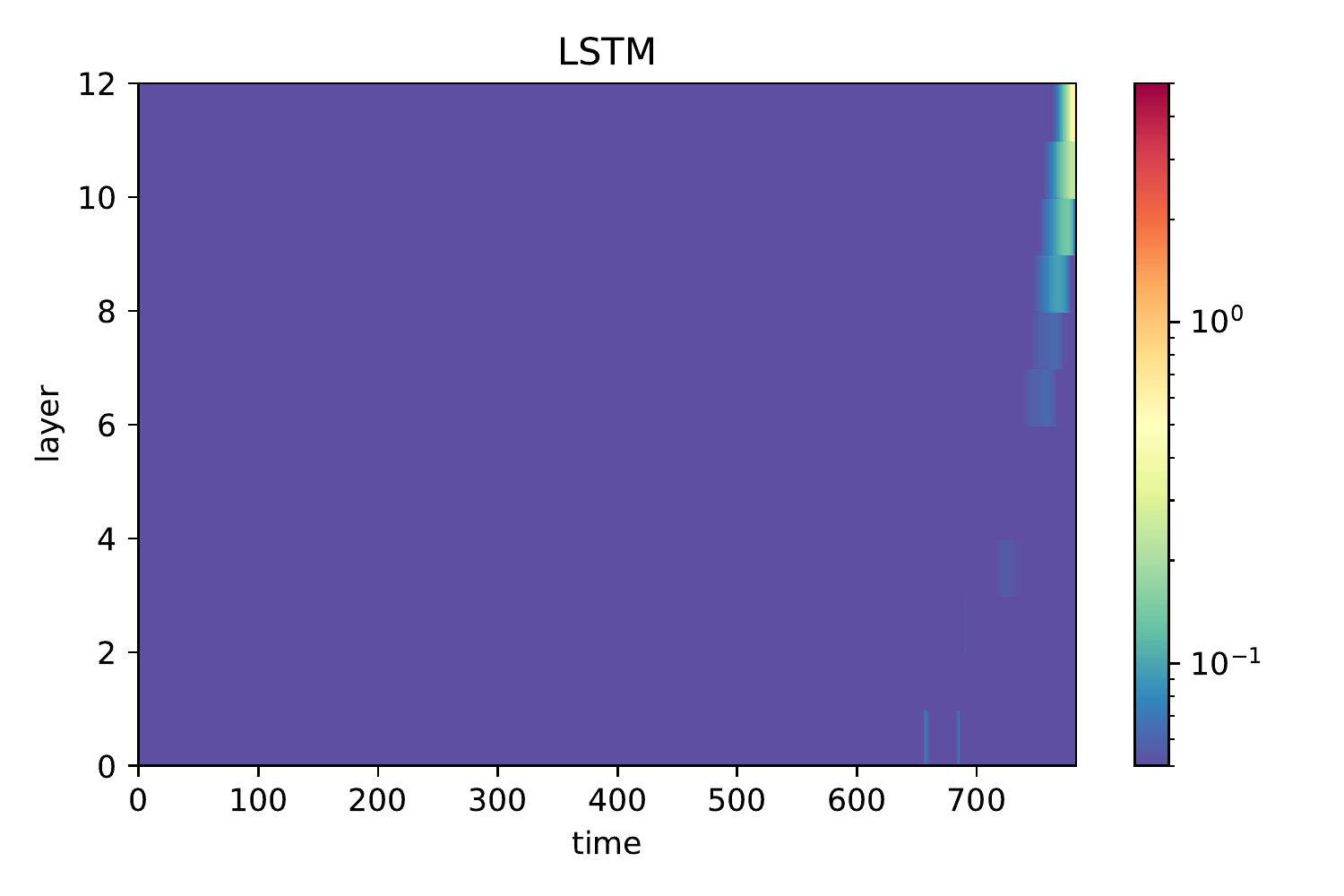}
        \caption{LSTM}
        \label{fig:sim_mnist_lstm}
    \end{subfigure}
    \begin{subfigure}[b]{0.22\textwidth}
        \includegraphics[width=\textwidth]{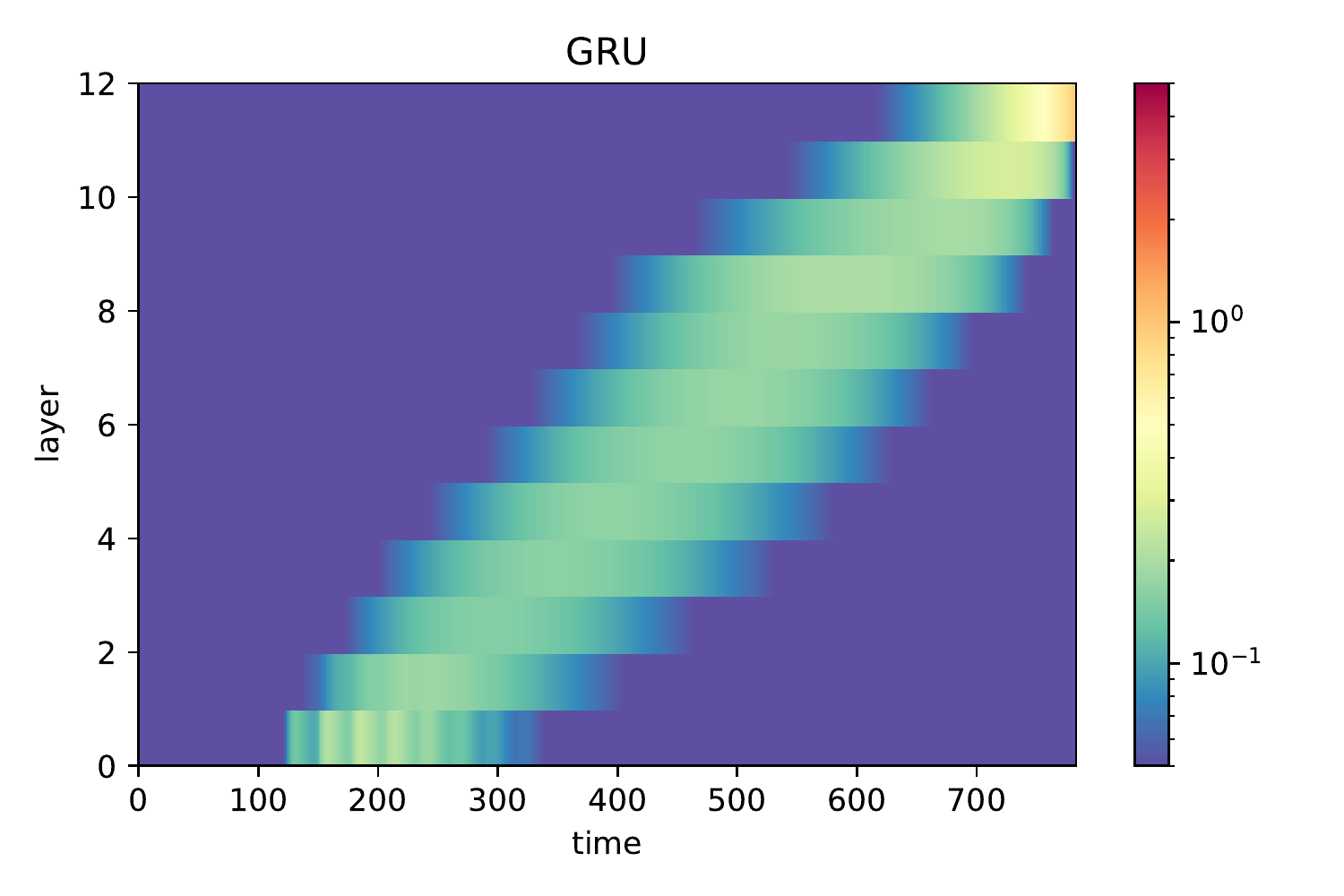}
        \caption{GRU}
        \label{fig:sim_mnist_gru}
    \end{subfigure}  
        \begin{subfigure}[b]{0.195\textwidth}
        \includegraphics[width=\textwidth]{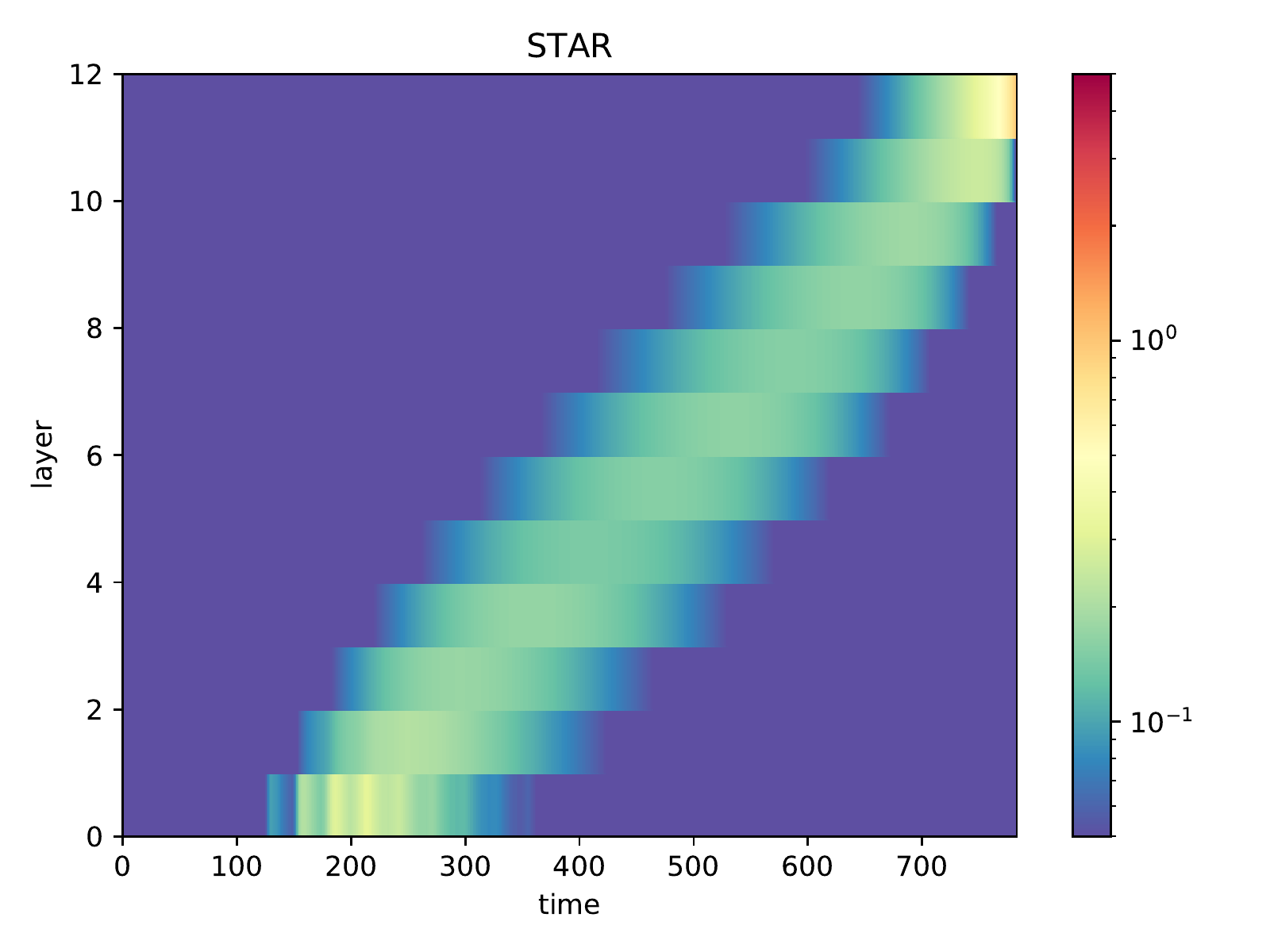}
        \caption{STAR}
        \label{fig:sim_mnist_star}
    \end{subfigure}  
    \caption{Mean gradient magnitude w.r.t.\ the parameters for vRNN, LSTM, GRU, and the proposed STAR cell for MNIST dataset. Loss $\mathcal{L}(h^L_T)$ only on final prediction.}\label{fig:sim_mnist}
\end{figure}

\begin{figure}
    \centering
    \begin{subfigure}[b]{0.22\textwidth}
        \includegraphics[width=\textwidth]{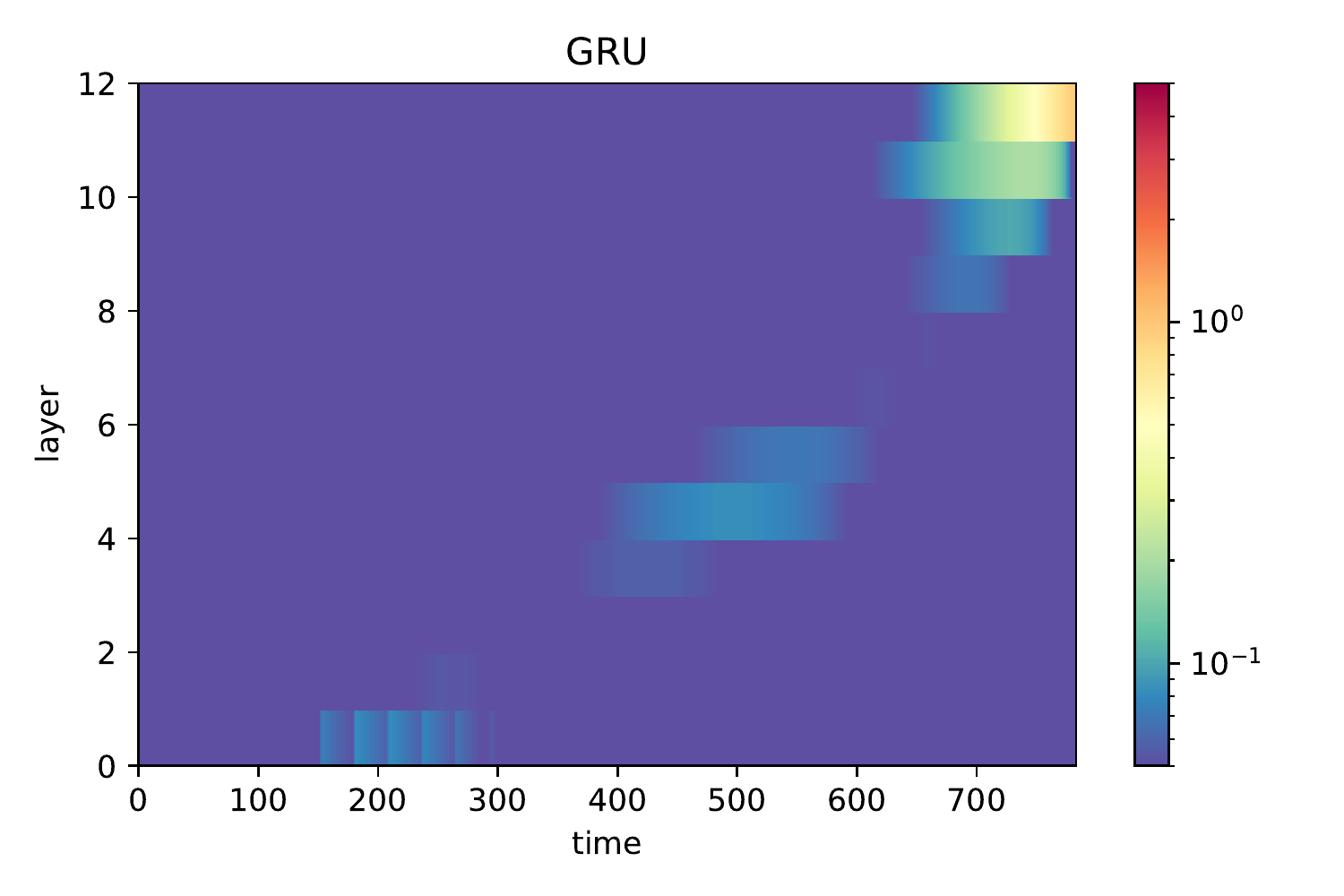}
    \end{subfigure}  
    \begin{subfigure}[b]{0.22\textwidth}
        \includegraphics[width=\textwidth]{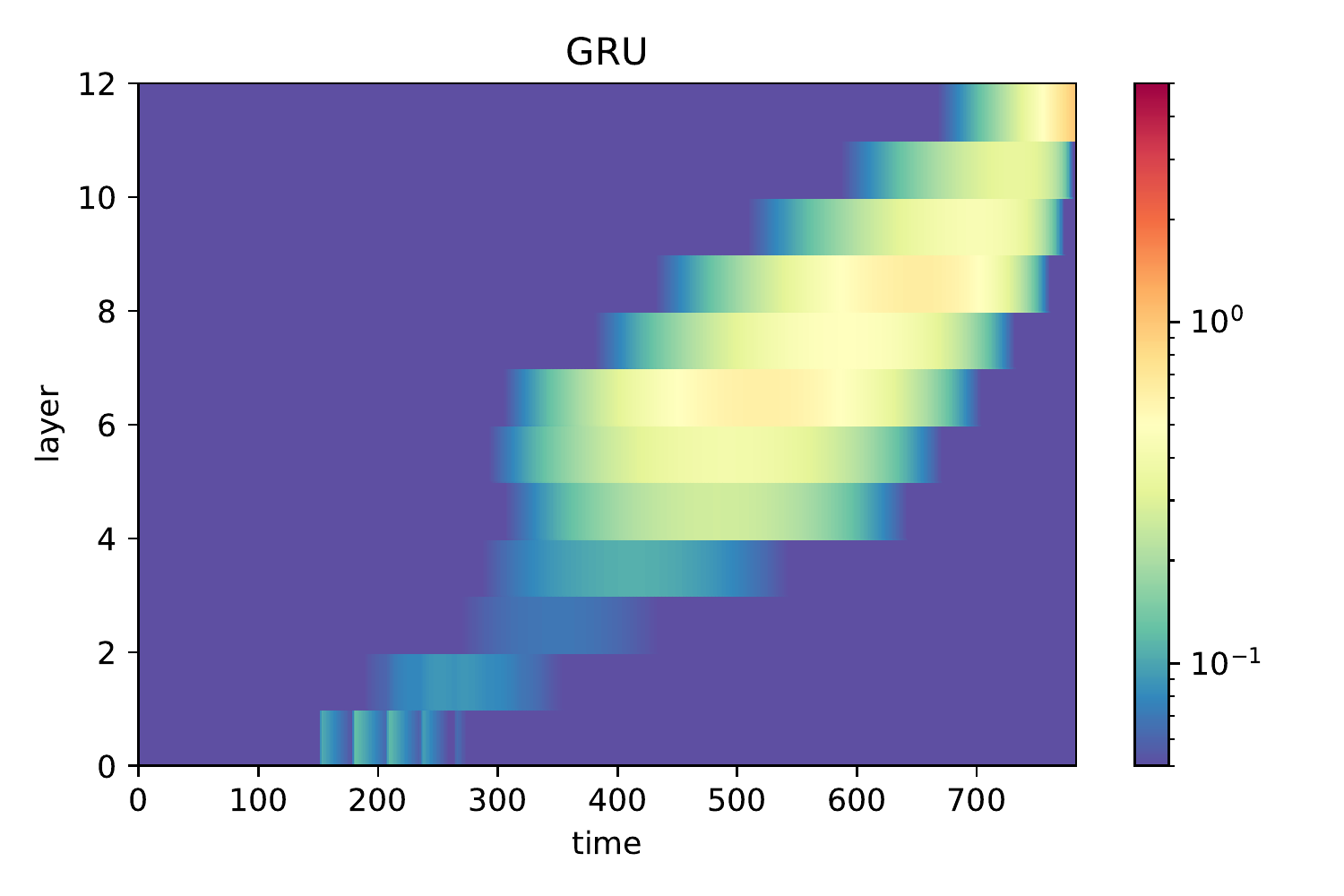}
    \end{subfigure}  
      \begin{subfigure}[b]{0.22\textwidth}
        \includegraphics[width=\textwidth]{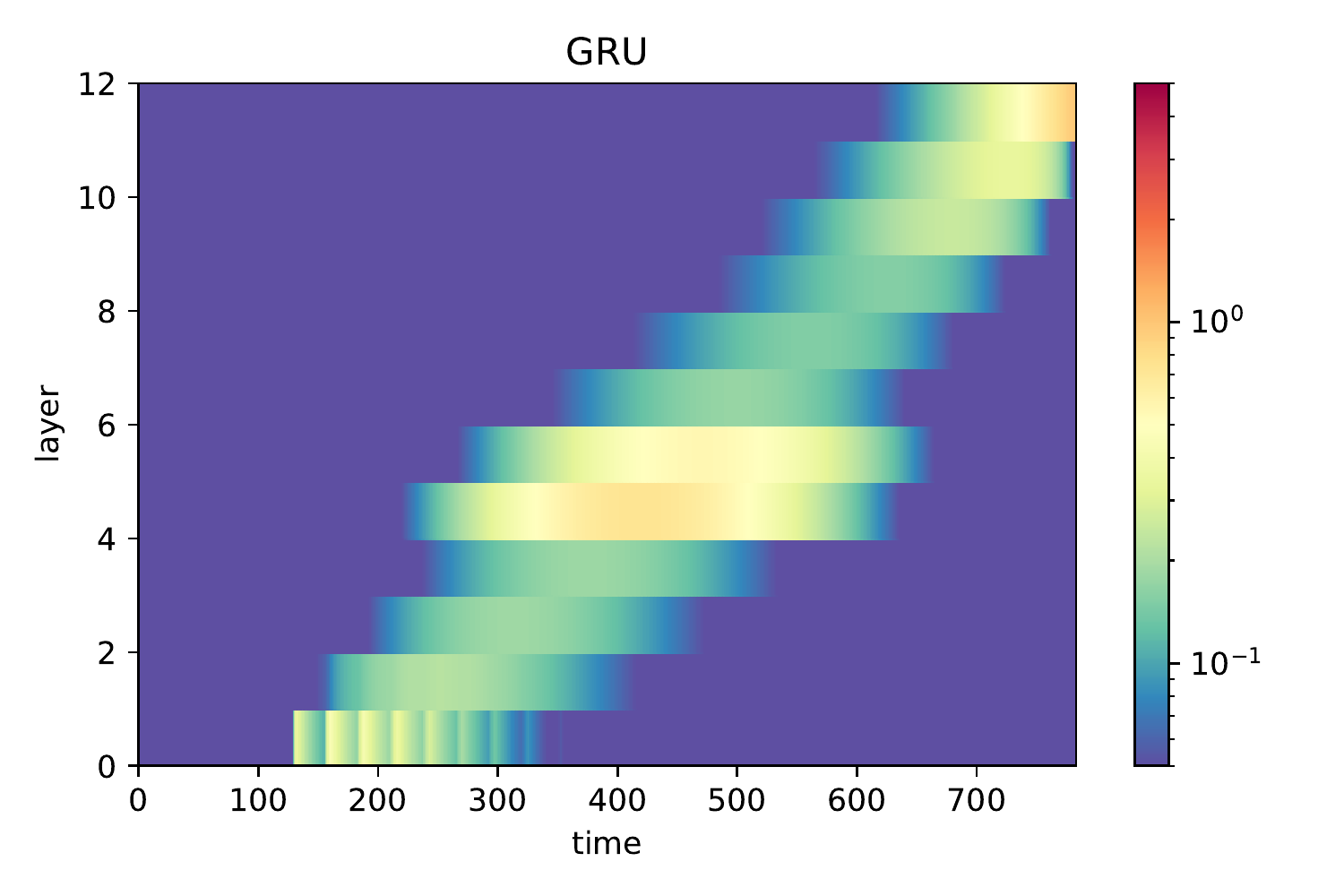}
    \end{subfigure}
        \begin{subfigure}[b]{0.22\textwidth}
        \includegraphics[width=\textwidth]{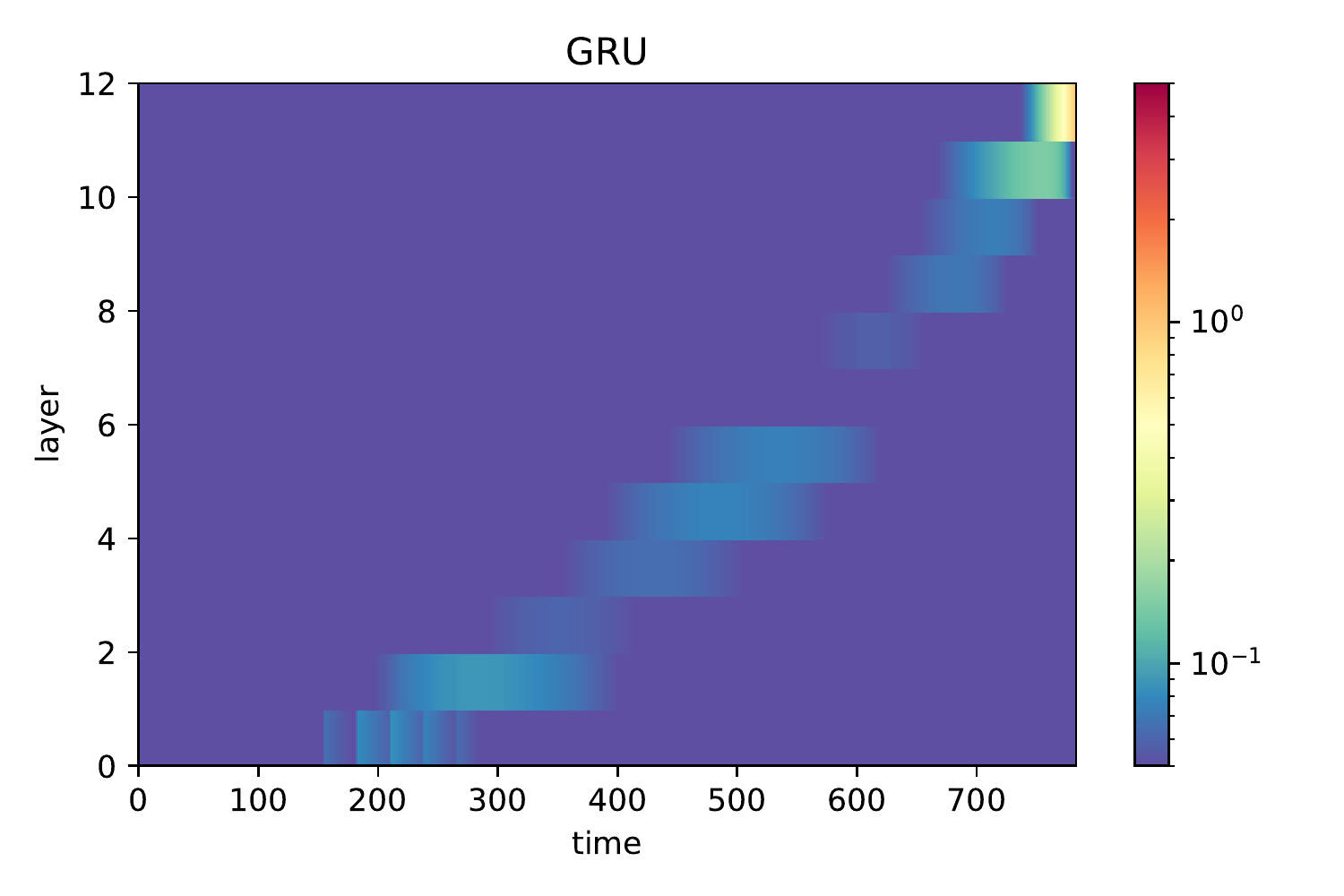}
    \end{subfigure}  
    \begin{subfigure}[b]{0.22\textwidth}
        \includegraphics[width=\textwidth]{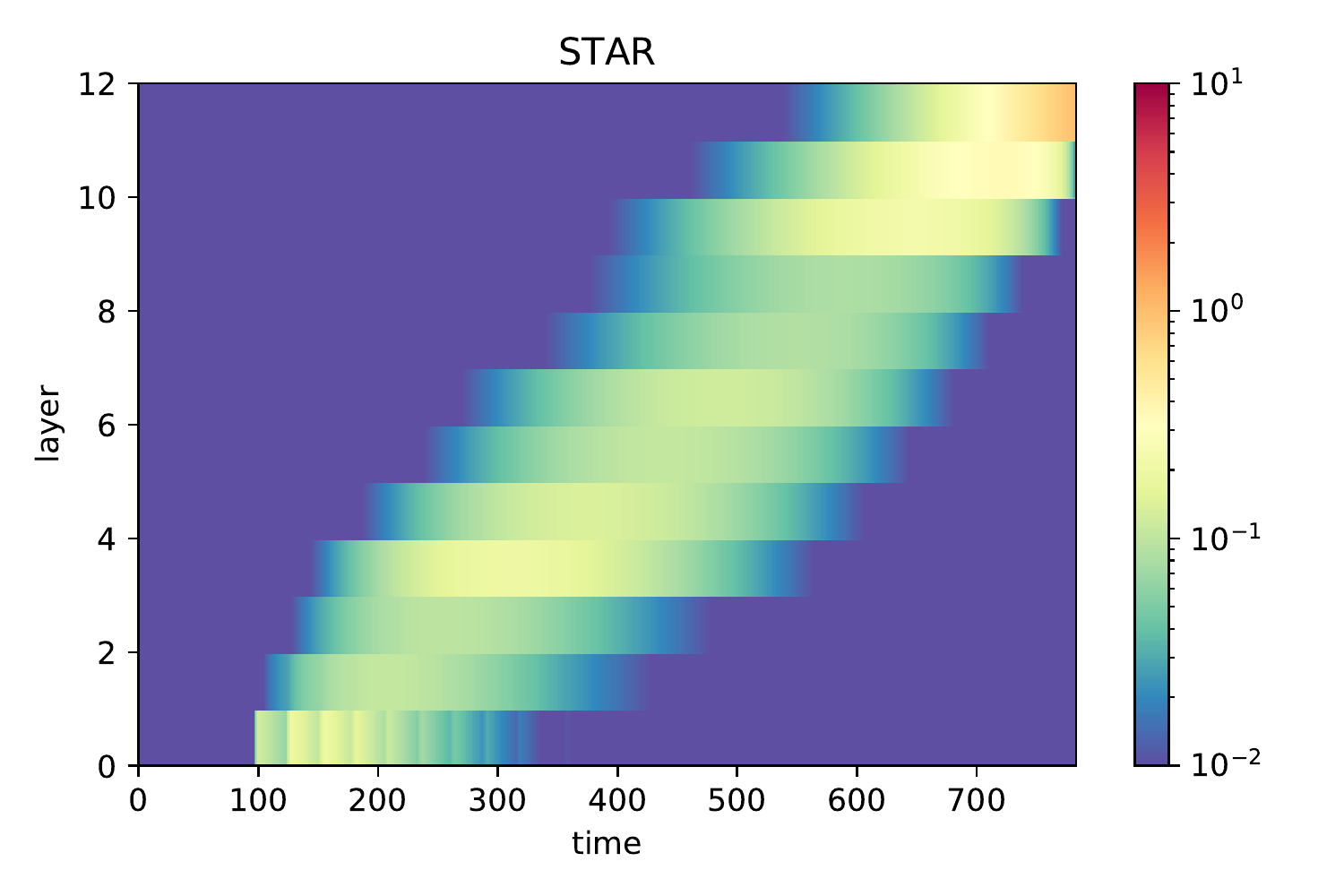}
    \end{subfigure}  
      \begin{subfigure}[b]{0.22\textwidth}
        \includegraphics[width=\textwidth]{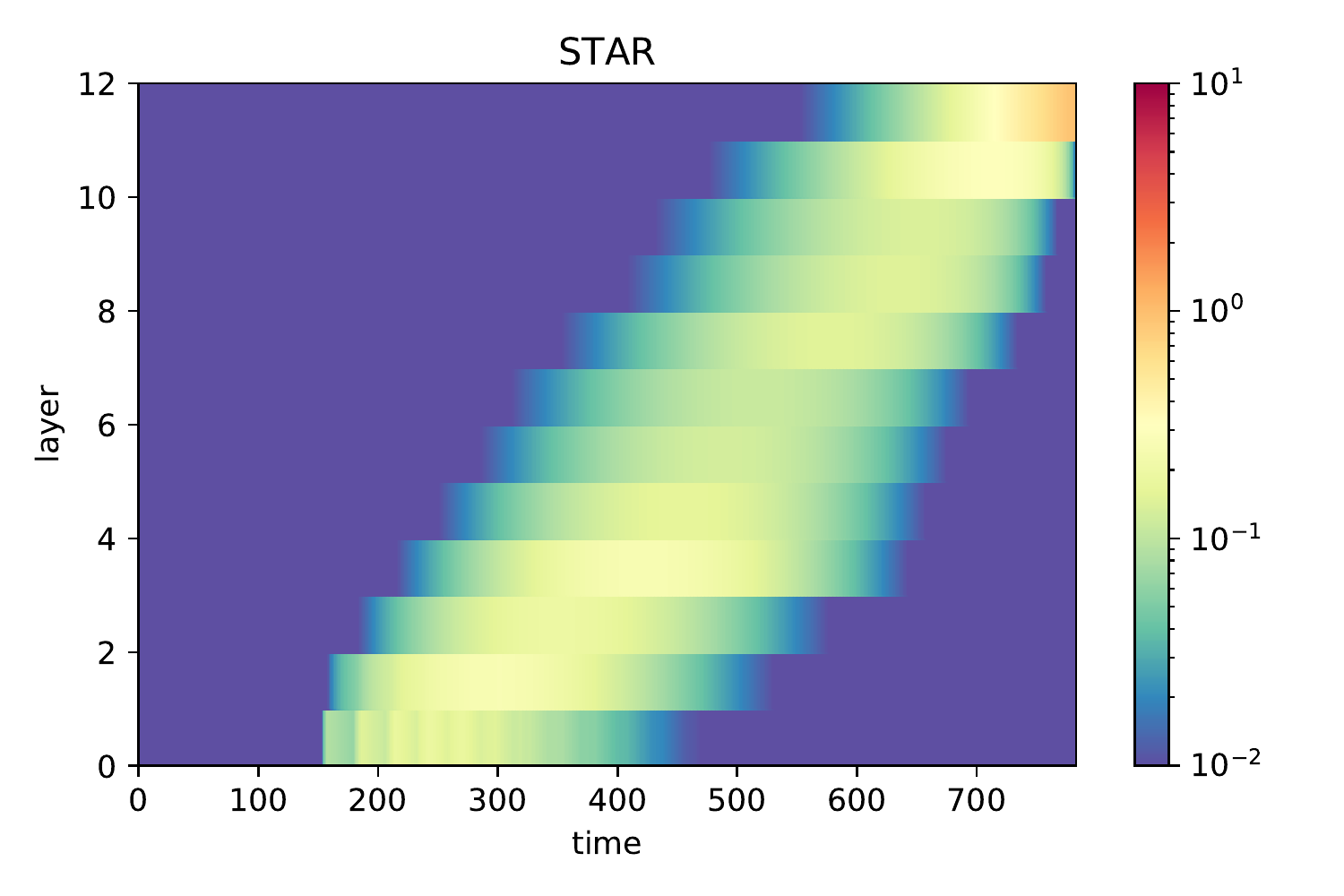}
    \end{subfigure}
    \begin{subfigure}[b]{0.22\textwidth}
        \includegraphics[width=\textwidth]{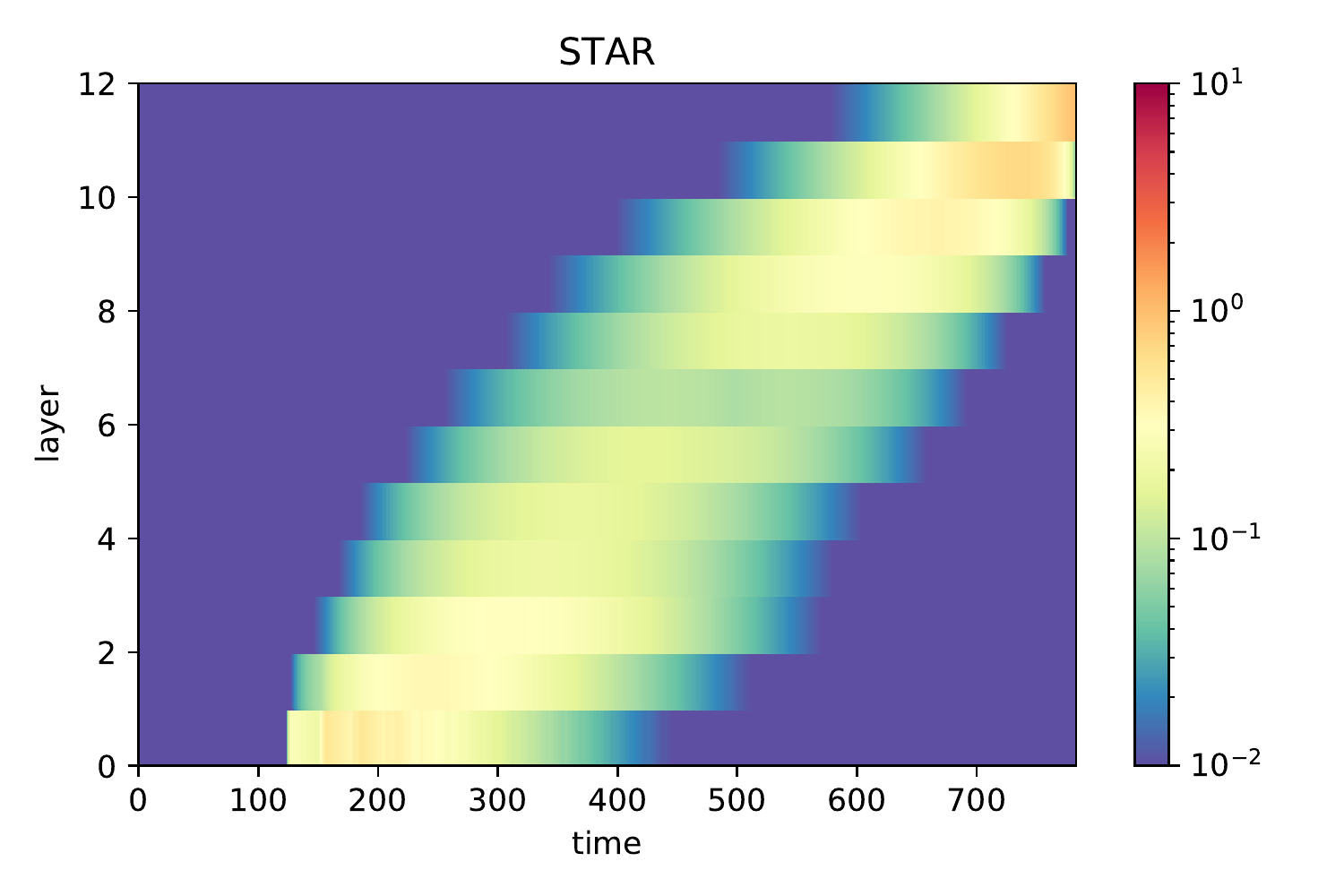}
    \end{subfigure}  
        \begin{subfigure}[b]{0.22\textwidth}
        \includegraphics[width=\textwidth]{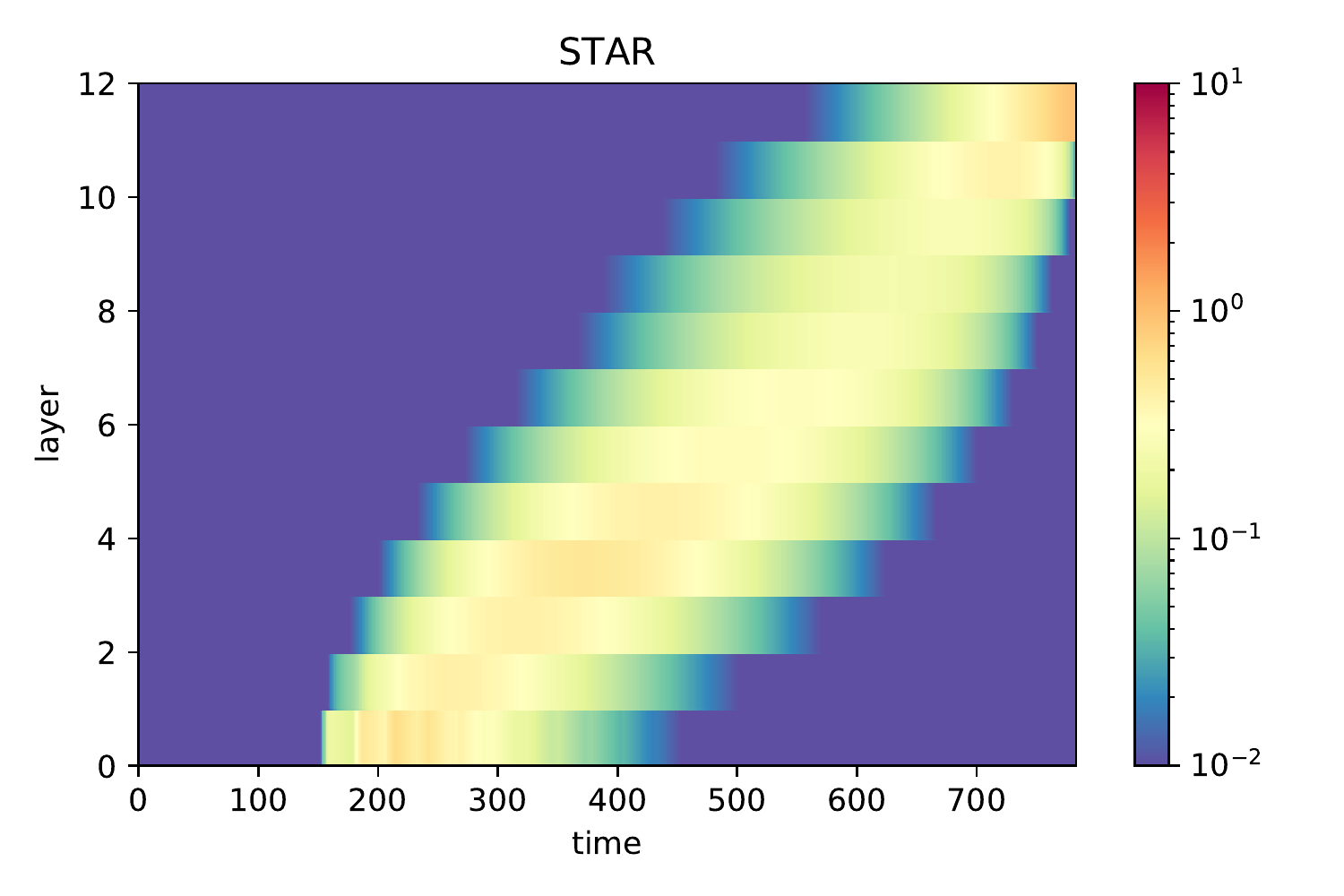}
    \end{subfigure}      
    
    \caption{Gradient magnitude comparison within a single run for MNIST dataset. \emph{top two rows:} GRU samples. \emph{bottom two rows:} STAR samples. Samples are randomly picked.}\label{fig:gru_vs_star}
\end{figure}

In Fig.~\ref{fig:sim_mnist} we can see that cells show similar behavior for the MNIST dataset. Even though on average STAR and GRU signal propagation looks fine, 
gradients within a single run oscillate a lot more for GRU (see Fig.~\ref{fig:gru_vs_star}) as seen in the previous numerical simulation (see Fig.~\ref{fig:simulation_app_std}).  

\subsection{Training details}
We provide more details about training procedures for the experimental analysis in the main paper in this section.

\subsubsection{Pixel-by-pixel MNIST}
Following \cite{chrono_init}, chrono initialisation is applied for the bias term of $\vk$, $\vb_k$. The basic idea is that $\vk$ should not be too large; such that the memory $\vh$ can be retained over longer time intervals. The same initialisation is used for the input and forget bias of the LSTM and the RHN and for the forget bias of the LSTMw/f and the GRU. For the final prediction, a feedforward layer with softmax activation converts the hidden state to a class label. The numbers of hidden units in the RNN layers are set to 128. All networks are trained for 100 epochs with batch size 100, using the Adam optimizer \cite{adam} with learning rate $0.001$, $\beta_1=0.9$ and $\beta_2=0.999$.

\subsubsection{TUM time series classification}
We use the same training procedure as described in the previous section for pixel-by-pixel MNIST. Again, a feedforward layer is appended to the RNN output to obtain a prediction. The numbers of hidden units in the RNN layers is set to $128$. All networks are trained for 30 epochs with batch size 500, using Adam~\cite{adam} with learning rate $0.001$, $\beta_1=0.9$ and $\beta_2=0.999$.

\subsubsection{BreizhCrop time series classification}
A feedforward layer is appended to the RNN output to obtain a prediction. The numbers of hidden units in the RNN layers is set to $128$. All networks are trained for 30 epochs with batch size 1024, using Adam~\cite{adam} with learning rate $0.001$ and $\beta_1=0.9$ and $\beta_2=0.999$. The learning rate scheduler of~\cite{transformer} is used with 10 warm-up steps.

\subsubsection{Adding problem / Copy memory}
Following \cite{chrono_init}, chrono initialisation is applied for the bias term of $\vk$, $\vb_k$. The same initialisation is used for the input and forget bias of the LSTM and for the forget bias of the GRU. The number of hidden units is set to $128$ for STAR and LSTM, $150$ for GRU and $256$ for vRNN. 2-layer STAR is used to have same number of parameters. Networks are trained using Adam~\cite{adam} with learning rate $0.001$, $\beta_1=0.9$ and $\beta_2=0.999$. 

\subsubsection{Music modeling}
We follow the exact same experimental setup described in~\cite{diagRNN}. Baseline results are taken from~\cite{diagRNN}. The input sequence length is set to $200$. STAR is trained for 500 iterations with batch size 1, using RMSProp. Dropout with keep probability $0.8$ is applied. Other hyper-parameters (number of layer, momentum etc.) are searched as described in~\cite{diagRNN}. 

\subsubsection{Character-level language modeling}
We follow the exact same experimental setup described in~\cite{tcn}. Results for vRNN, LSTM, GRU and TCN are directly taken from~\cite{tcn}. The input sequence length is set to $400$. The number of hidden units is set to $410$ for STAR; therefore, the total number of parameters for 6-layers STAR makes 3M. STAR is trained for 50 epochs with batch size 32, using Adam~\cite{adam} with learning rate $0.001$, $\beta_1=0.9$ and $\beta_2=0.999$. The learning rate is decayed when the validation performance is no longer improved. Gradient clipping with $1$ is applied. For IndRNN, the input sequence length is set to $50$ because it performs poorly if set to $400$. We set the number of hidden units to $660$; therefore, the total number of parameters for 6-layers IndRNN is 3M and we train it for $100$ epochs. Note that we took the IndRNN implementation from \href{https://github.com/Sunnydreamrain/IndRNN_pytorch}{https://github.com/Sunnydreamrain/IndRNN\_pytorch}.

\subsubsection{Hand-gesture recognition from video}
%Model details
All convolutional kernels are of size 3$\times$3. Each convolutional RNN layer has $64$ filters. A shallow CNN is used to convert the hidden state to a label, with 4 layers that have filter depths 128, 128, 256 and 256, respectively.
All models are trained with stochastic gradient descent (SGD) with momentum ($\beta=0.9$). The batch size is set to $8$, the learning rate starts at $0.001$ and decays polynomially to $0.000001$ over a total of $30$ epochs. $L2$-regularisation with weight $0.00005$ is applied to all parameters.

\subsubsection{TUM image series pixel-wise classification}
%Model details
All convolutional kernels are of size 3$\times$3. Each convolutional RNN layer has $32$ filters. A shallow CNN is used to convert the hidden state to a label, with 2 layers that have filter depths 64.
All models are fitted with Adam~\cite{adam}. The batch size is set to $1$, the learning rate starts at $0.001$ and decays polynomially to $0.000001$ over a total of $25$ epochs.

\clearpage

% Weight norms and average h

%comparison with indRNN

\begin{figure}[th]
    \centering
        \includegraphics[width=\columnwidth]{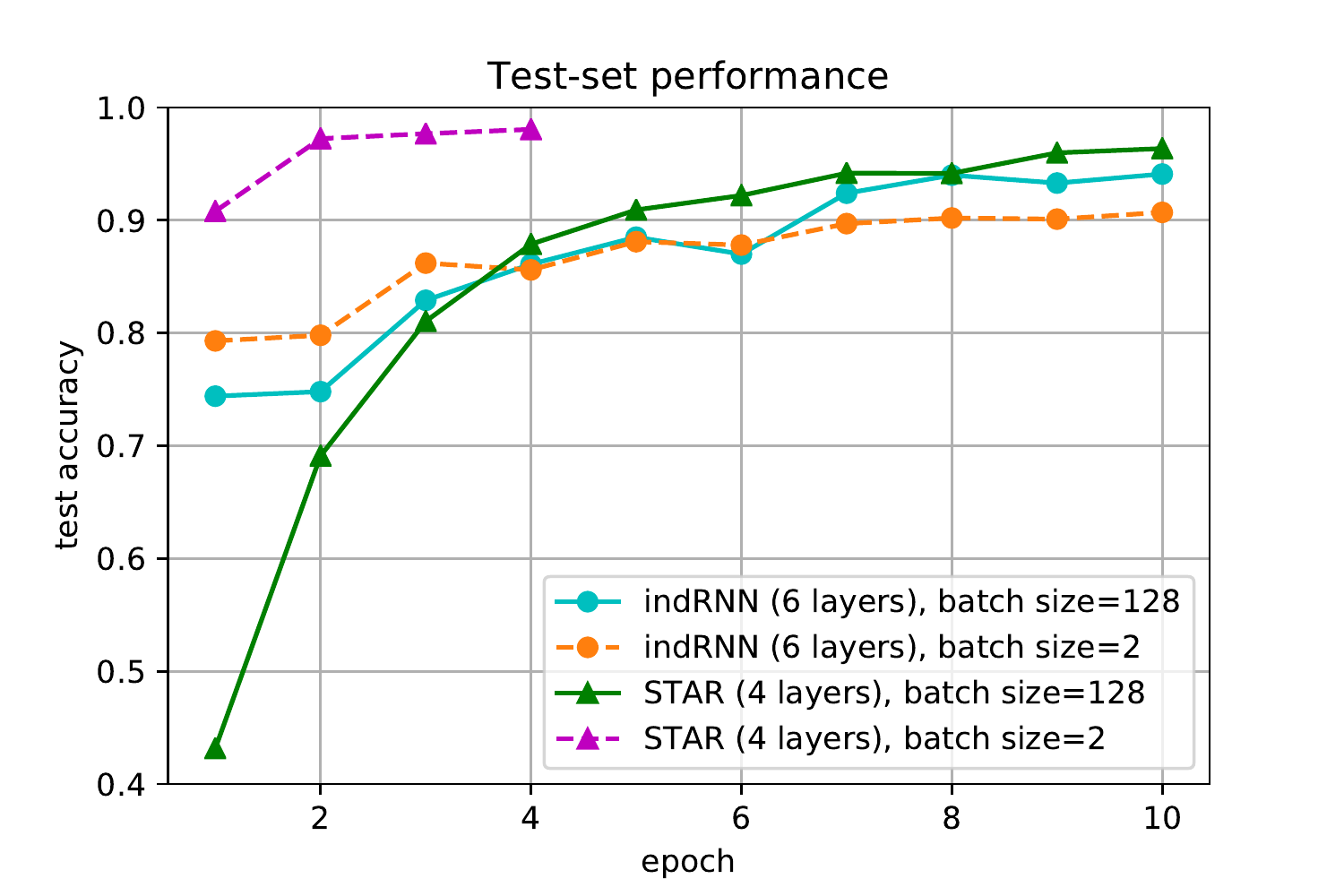}
    \caption{Performance comparison for different batch sizes on the sequential MNIST task. If using a batch size of $128$, both STAR and IndRNN converge to a solution; IndRNN is faster at the beginning but STAR eventually achieves better performance. IndRNN becomes very slow to train for a batch size of $2$ ($64x$ more steps per epoch) and it cannot achieve the same test performance as with the standard batch size ($128$). In contrast, STAR does not encounter these problems and clearly performs superior.}
   \label{fig:bn_effect}
\end{figure}

%\begin{figure}[th]
%    \centering
%        \includegraphics[width=\columnwidth]{reviews/rebuttal_images/phase_portrait_star.pdf}
%    \caption{\textcolor{blue}{Mapping of STAR cell for 1-D hidden state case is depicted. Arrows show in which direction the hidden state is mapped by the cell function. It shows $h_t^{l-1} = h_{t-1}^l=0$ is stable and attracting fixed point.} }
%    \label{fig:phase}
%\end{figure}

\end{document}